\documentclass[12pt,singlespace,twoside]{mitthesis}
\usepackage{lgrind}
\usepackage{cmap}
\usepackage[T1]{fontenc}

\usepackage{lineno,hyperref}
\usepackage{tikz,pgffor}
\usepackage{times}
\usepackage{wrapfig}
\usepackage{graphicx}
\usepackage{amsmath}
\usepackage{amssymb}
\usepackage{verbatim}
\usepackage{booktabs}
\usepackage{slgloss}
\usepackage{array,multirow}
\usepackage{booktabs}

\newcommand{\caja}[4][1]{{%
    \renewcommand{\arraystretch}{#1}%
    \begin{tabular}[#2]{@{}#3@{}}%
      #4%
    \end{tabular}%
    }}

\usetikzlibrary{calc}
\usetikzlibrary{decorations.pathreplacing}

\graphicspath{{grf/}}

\renewcommand{\sp}{\overline{p}}

\newcommand{\karenedit}[1]{{#1}}

\renewcommand{\eqref}[1]{Eq.~(\ref{#1})}

\begin{document}

\title{American Sign Language fingerspelling recognition \\
from video: Methods for unrestricted recognition \\ 
and signer-independence}

\author{Taehwan Kim}
\department{TOYOTA TECHNOLOGICAL INSTITUTE AT CHICAGO}

\degree{Doctor of Philosophy in Computer Science}

\degreemonth{August}
\degreeyear{2016}
\thesisdate{August 25, 2016}

\copyrightnoticetext{\copyright Taehwan Kim}

\supervisor{Karen Livescu}{Assistant Professor}

\chairman{Arthur C. Smith}{Chairman, Department Committee on Graduate Theses}

\maketitle



\cleardoublepage
\setcounter{savepage}{\thepage}
\begin{abstractpage}
%
%
%
In this thesis, we study the problem of recognizing video sequences of fingerspelled
letters in American Sign Language (ASL).  Fingerspelling comprises a
significant but relatively understudied part of ASL, and recognizing it is challenging for a number of reasons:  It involves quick, small motions
that are often highly coarticulated; it exhibits significant variation
between signers; and there has been a dearth of
continuous fingerspelling data collected.  In this work, we propose several types of
recognition approaches, and explore the signer variation problem.
 Our best-performing models are segmental (semi-Markov) conditional
 random fields using deep neural network-based features.  In the
 signer-dependent setting, our recognizers achieve up to about 8\%
 letter error rates.  The signer-independent setting is much more
 challenging, but with neural network adaptation we achieve up to 17\%
 letter error rates.

\end{abstractpage}


\cleardoublepage
%
%
%

\begin{titlepage}
\large
\Large{\bf American Sign Language fingerspelling recognition \\
from video: Methods for unrestricted recognition \\ 
and signer-independence} \\

 \hfill\break
  \hfill\break
\normalsize
A thesis presented \\
by \\
\Large{Taehwan Kim} \\ 
 \hfill\break
  \hfill\break

\normalsize
in partial fulfillment of the requirements for the degree of \\
 \hfill\break
Doctor of Philosophy in Computer Science \\
  \hfill\break
Toyota Technological Institute at Chicago \\
  \hfill\break
Chicago, Illinois \\
  \hfill\break
August 2016

  \hfill\break
  \hfill\break
  \large
  $-$ Thesis Committee $-$ \\
 \hfill\break
\resizebox{\linewidth}{!}{
\begin{tabular}{lclcc}
&  & & & \\
Vassilis Athitsos &  & & &  \\\hline
Committee member (print/type) & ~~~~~~~~~ & Signature & ~~~~~~ & Date \\
&  & & & \\
Karen Livescu &  & & &  \\\hline
Thesis/Research Advisor (print/type) & ~~~~~~~~~ & Signature & ~~~~~~ & Date \\
&  & & & \\
Greg Shakhnarovich &  & & &  \\\hline
Committee member (print/type) & ~~~~~~~~~ & Signature & ~~~~~~ & Date \\
&  & & & \\
Yisong Yue &  & & &  \\\hline
Committee member (print/type) & ~~~~~~~~~ & Signature & ~~~~~~ & Date \\
&  & & & \\
David McAllester &  & & & \\\hline
Chief Academic Officer (print/type) & ~~~~~~~~~ & Signature & ~~~~~~ & Date \\

\end{tabular}
}

\end{titlepage}

\begin{titlepage}

To my parents
\end{titlepage}

\section*{Acknowledgments}

First and foremost, I would like to thank my PhD advisor, Karen Livescu, for insightful guidance, endless patience, and kindness throughout my PhD. This thesis would not have been possible without her support and encouragement. I also thank my thesis committee members, Vassilis Athitsos, Greg Shakhnarovich, and Yisong Yue for guiding and shaping this work through their invaluable feedback and help.

Many others have contributed, both directly and indirectly, to the work described in this thesis. I was fortunate to collaborate with and thank my co-authors: Iain Matthews, Hao Tang, Sarah Taylor, Raquel Urtasun, and Weiran Wang. I am grateful to Adam Bohlander and Chrissy Novak for their kindness and help for all things administrative. I thank David McAllester for his help and leading this great school. I thank my fellow graduate students, Avleen Bijral, Heejin Choi, Andrew Cotter, Jonathan Keane, Jian Peng, Karthik Sridharan, Zhiyong Wang, Payman Yadollahpour, Jian Yao, and Feng Zhao for their inspiration and encourgement.

I would like to express my deepest gratitude to my family, without whose support I would not have been able to complete my PhD. I thank Taemi for her endless encouragement. I thank my parents, Kwang-kuk and Jung-soon, for their unconditional love and support. I am grateful to always be in their prayer. I dedicate this thesis to my parents.

At the end, I would like to give my thanks to the living God, who has guided and will guide my life, and whose steadfast love for me endures forever.


\pagestyle{plain}
\tableofcontents
\newpage
\listoftables
\newpage
\listoffigures

\chapter{Introduction}
\label{sec:intro}

Sign languages are the primary means of communication for millions of Deaf people in the world\footnote{This thesis includes material previously published in several papers~\cite{kim2012,kim2013,kim2016}.}.  
In the US, there are about 350,000--500,000 people for whom American Sign Language (ASL) is the primary language~\cite{Mitchell:2006}.  
Automatic sign language recognition is a nascent technology that has the potential to improve the ability of Deaf and hearing individuals to communicate, as well as Deaf individuals' ability to take full advantage of modern information technology.  For example, online sign language video blogs and news\footnote{E.g., \tt{http://ideafnews.com, http://aslized.org}.} are currently almost completely unindexed and unsearchable as they include little accompanying annotation.  
While there has been extensive research over several decades on automatic recognition and analysis of spoken language, much less progress has been made for sign languages.
Both signers and non-signers would benefit from technology that improves communication between these populations and facilitates search and retrieval in sign language video.

Research on this problem has included both speech-inspired approaches and computer vision-based techniques, using either/both video and depth sensor input~\cite{dreuw,Zaki-Shaheen-11,Bowden-et-al-04,Liwicki-Everingham-09,The2010,VoglerM99,kim2012,kim2013,forster2013improving}.
We focus on recognition from video for applicability to existing recordings.  Before the technology can be applied ``in the wild'', it must overcome challenges posed by visual nuisance parameters (e.g., lighting, occlusions) and signer variation.  Annotated data sets for this problem are scarce, in part due to the need to recruit signers and skilled annotators.  



In this thesis we consider American Sign Language (ASL), and focus in particular on
recognition of fingerspelled letter sequences.  In fingerspelling,
signers spell out a word as a sequence of handshapes or hand
trajectories corresponding to individual letters.
\karenedit{The handshapes used in fingerspelling are also used throughout ASL.  In fact, the fingerspelling handshapes account for about 72\% of ASL handshapes~\cite{Brentari:2001}, making research on fingerspelling applicable to ASL in general.}

\begin{figure}
\centering
\includegraphics[width=0.5\textwidth]{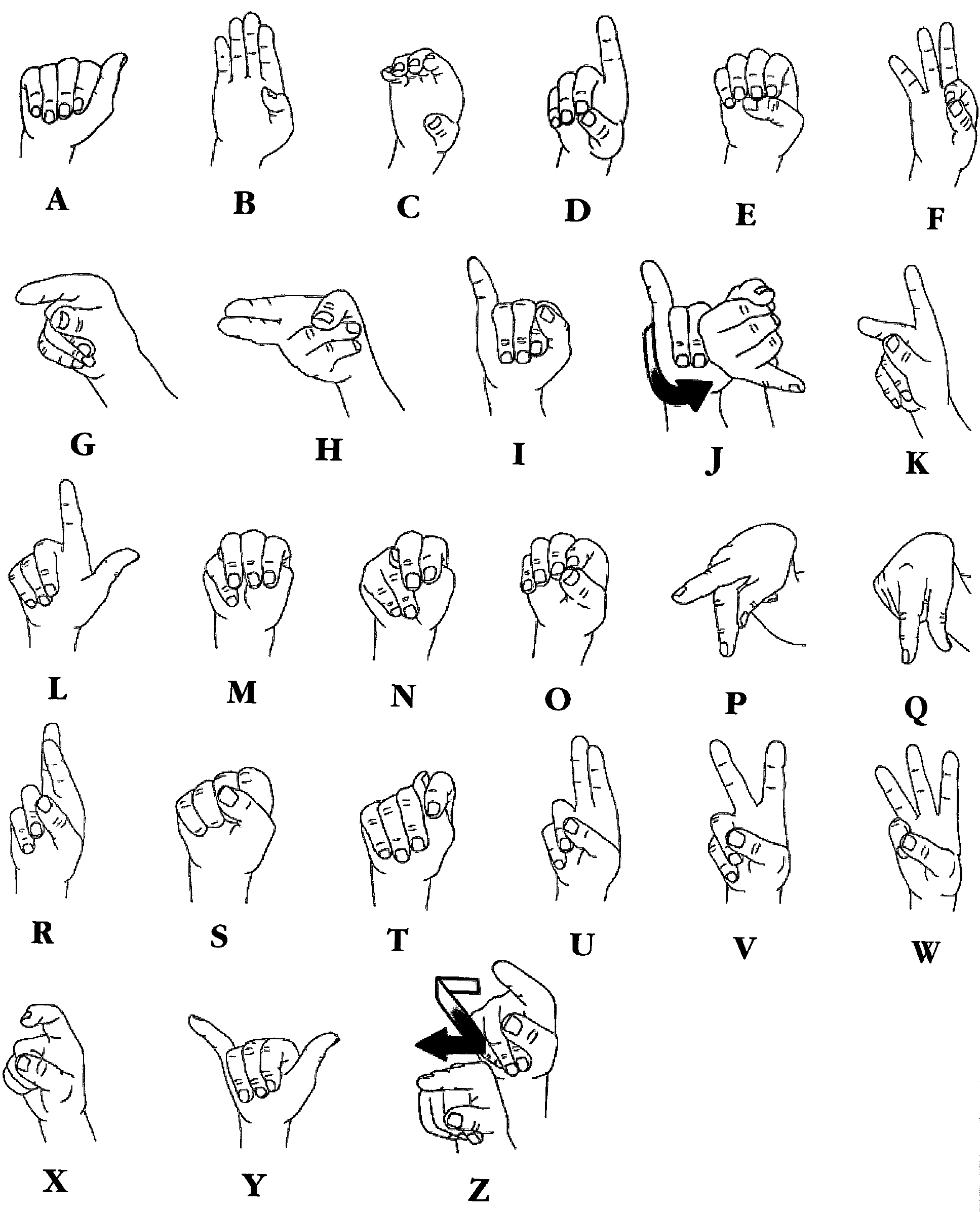}
\caption[The ASL fingerspelled alphabet]{The ASL fingerspelled alphabet.  Reproduced from~\cite{HumPad2004}.}
\label{f:alphabet}
\end{figure}

Figure~\ref{f:alphabet} shows the ASL fingerspelling alphabet.
Fingerspelling is a constrained but important part of ASL, accounting
for up to 35\% of ASL~\cite{Padden}.  Fingerspelling is typically used
for names, borrowings from English or other spoken languages, or new coinages.  ASL fingerspelling uses a single hand and involves relatively small and quick motions of the hand and fingers, as opposed to the typically larger arm motions involved in other signs.  Therefore, fingerspelling can be difficult to analyze with standard approaches for pose estimation and tracking from video.

\section{Challenges}
In many ways the problem of fingerspelling recognition is analogous to that of word sequence or phone sequence recognition.  However, there are some special challenges that fingerspelling introduces.  We address some of these challenges here, while others are left to future work.

\begin{enumerate}
\item ASL fingerspelling uses a
single hand and involves relatively small and quick motions
of the hand and fingers, as opposed to the typically larger
arm motions involved in other signs. Therefore, fingerspelling
can be difficult to analyze with standard approaches
for pose estimation and tracking from video.

\item Existing work on fingerspelling recognition has mostly focused on static images extracted from videos, or on letter sequences from a small vocabulary.

\item Another challenge is the high variability among signers, with the limited available training data; main sources of signer variation are
speed, hand appearance, and non-signing motion variation before/after
signing as shown in Figure \ref{fig:ex}. These make signer-independent recognition quite difficult.

\item There are specific linguistic characteristics of sign language and fingerspelling. For example, the concept of a ``silence'' unit is quite different from that of acoustic silence.  By ``silence'', we mean any video segment that does not correspond to fingerspelling, including any hand motion that is not linguistically meaningful.
This may be thought of as a ``garbage'' unit, but its appearance is highly dependent on the context.  
In our data, for example, ``silence'' typically corresponds to the time at the beginning and end of each letter sequence, when the signer's hand is rising to/falling from the signing position.  Also, double letters are usually not signed as two copies of the same letter, but rather as either a single longer articulation or a special sign for the doubled letter.  For example, 'ZZ' is often signed identically to a 'Z' but using two extended fingers rather than the usual one. 


\item The language model over fingerspelled letter sequences is difficult to estimate.  There is no large database of natural fingerspelled sequences in running sign.
In our data set, the distribution of words has been chosen to maximize coverage of letter n-grams and word types rather than to follow some natural distribution.  Fingerspelling does not follow the distribution of, say, English words, since fingerspelling is most often used for words lacking ASL signs, such as names.  
For this work, we estimate language models from large English dictionaries that include names.  However, such language models are not a perfect fit, and this issue requires more attention.

\end{enumerate}

\begin{figure}[h!]
	\hspace{-.5em}\includegraphics[width=1\linewidth]{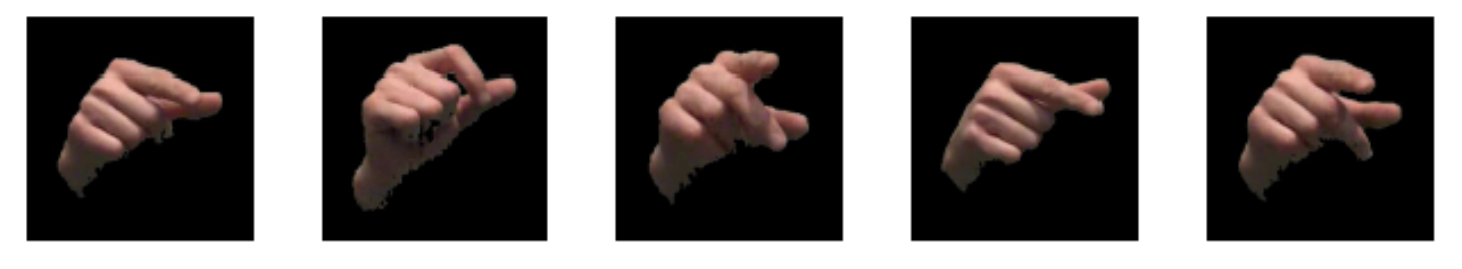}
\vspace{-.2in}
\caption{Example images of Q recognized as G.}
\label{f:Q_as_G}
\end{figure}
Therefore, these make fingerspelling recognition quite challenging. For instance, Figure \ref{f:Q_as_G} gives example images of Q misrecognized as G by one of our recognition systems. It shows various handshape variations of Q that are similar to G, confusing to recognizers. 

We are interested in addressing both the technological and the linguistic issues, focusing mainly on {\it handshape modeling}.  
Sign language handshape has its own phonology that has been studied but does not yet enjoy a broadly agreed-upon understanding \cite{bren}.
Recent linguistic work on sign language phonology has developed approaches based on articulatory features, which are closely related to motions of parts of the hand \cite{DingM09}.
Sign language recognition research has focused more on the larger motions of sign and on interactions between multiple body parts, and less so on the details of handshape \cite{Dreuw-et-al-07, Yang-Sarkar-06}.
At the same time, computer vision research has studied pose estimation and tracking of hands \cite{tang2013real}, but usually not in the context of a grammar that constrains the motion.  
There is therefore a great need to better understand and model the handshape properties of sign language. Therefore, we propose to use linguistic features suggested by linguistics research on ASL \cite{bren,Keane2014diss}.

\section{Contributions}


This thesis advances fingerspelling recognition from video in several ways.

\begin{enumerate}

\item Most previous work on fingerspelling and handshape has focused on restricted conditions such as careful articulation, isolated signs, restricted (20-100 word) vocabularies~\cite{goh,Liwicki-Everingham-09,Ricco-Tomasi-09}. In this project, we consider {\it unconstrained} fingerspelling sequences. This is a more natural setting, since fingerspelling is often used for names and other ``new'' terms, which may not appear in any closed vocabulary. The work presented here used the largest video data set of which we are aware
containing unconstrained, connected fingerspelling, consisting
of four signers each signing 600 word tokens for a total of
$\sim350$k image frames.

\item We develop discriminative segmental models, which
allow us to introduce more flexible features of fingerspelling
segments. The use of such segmental
feature functions is useful for gesture modeling, where it
is natural to consider the trajectory of some measurement
or the statistics of an entire segment. In this work we define feature functions based on scores of deep neural network classifiers of letter and handshape linguistic
features. In the category of segmental models, we compare models for
rescoring lattices and a first-pass segmental model.
We find that our segmental models outperform previous approaches including hidden Markov model / Gaussian mixture approaches and tandem hidden Markov models with deep neural network classifier input.


\item Signer-dependent applications~\cite{Liwicki-Everingham-09,kim2013} have achieved letter error rates (Levenshtein distances between hypothesized and true letter sequences, as a proportion of the number of true letters) of 10\% or less. However, the signer-independent setting
presents challenges due to significant variations among signers,
coupled with the dearth of available training data. We
investigate this problem with approaches inspired by automatic
speech recognition. We find that, using signer adaptation techniques, we are able to bridge a large part of the gap between signer-dependent and signer-independent recognition performance.

\item Ground-truth frame-level letter labels can be difficult and time-consuming to acquire.  In this thesis we study the effect of quality of labeling on recognition performance.  We develop recognizers that use either manually labeled and aligned data or unaligned data labeled only at the letter sequence level.  While we are able to use both with significant success, we find that there is still a large degradation in performance between manually labeled frames and automatically aligned data.  This remains one of the major challenges for future work.

\end{enumerate}


\section{Related work}
\label{sec:related_work}
There has been significant work on sign language
recognition from video. Even though some prior work has used additional input modalities, such as specialized gloves and depth sensors~\cite{Kadous-96,Grobel-Hienz-96,Oz-Leu-05,Wang-Popovic-09,Feris-et-al-04,pugeault2011spelling,keskin2012hand,zhang2015histogram,li2015feature}, 
video is more practical in many settings, and for online or archival recordings is the only choice; we restrict the remaining discussion to video.

Much prior work has used hidden Markov model (HMM)-based approaches~\cite{Starner-Pentland-98,Vogler-Metaxas-99,Grobel-Assan-97,Dreuw-et-al-07}, and this is the starting point for our work as well.
In addition, there have been some efforts involving conditional
models~\cite{Yang-Sarkar-06} and more complex (non-linear-chain)
graphical models~\cite{ThaEtal2011,VoglerM03}.  


Video corpora have been constructed for the task of recognition of sign language from video \cite{Martinez-et-al-02, Dreuw-et-al-07, Dreuw-et-al-2010, Athitsos-et-al-10}. But there are, to date, no corpora designed for ASL fingerspelling (and linguistic features of handshape in general) to our knowledge.

Motion capture systems \cite{Tyrone:1999, Wilcox:1992, Cheek:2001aa} and, more recently, Microsoft's Kinect\textsuperscript{TM} \cite{OikEtal2011, pugeault2011spelling} have also been used to collect sign language data. Most of this prior work has focused on the larger gestures of sign language rather than finger movement. One exception is \cite{LuHuenerfauth:2010}, which includes motion-capture glove data; however, this database is aimed at learning models for ASL generation and does not include the type of naturalistic, coarticulated data that we would like to study. Also \cite{lu2012cuny} collected motion capture ASL data consisting of both sign and fingerspelling, but small scale for fingerspelling.

Sign language recognition from video begins with front-end features.  Prior work has used a variety of visual 
features, including ones based on estimated position, shape and movement of the hands and
head~\cite{Bowden-et-al-04,Yang-Sarkar-06,Zahedi-et-al-06,Zaki-Shaheen-11}, sometimes combined with appearance
descriptors~\cite{Farhadi-et-al-07,Dreuw-et-al-07,Nayak-et-al-09,DingM09}
and color models~\cite{Buehler11,Pfister12}.  In this work, we are
aiming at relatively small motions that are difficult to track a
priori, and therefore begin with general image appearance features
based on histograms of oriented gradients (HOG
features)~\cite{HOG}.

Linguistically motivated representations of handshape and motion have been used in some prior work, e.g.,~\cite{Bowden-et-al-04,DingM09,ThaEtal2011,VoglerM03,VoglerM01,VoglerM99,Vogler-Metaxas-99,Pit2011,The2010}.
However, for connected fingerspelling recognition, a much finer level of detail is needed to represent the sub-articulators of the hand.  This motivates our study of linguistic handshape features.

A subset of ASL recognition work has focused specifically on fingerspelling and/or handshape classification~\cite{Athitsos-et-al-04,Rousos-et-al-10,pugeault2011spelling} and fingerspelling sequence recognition~\cite{goh,
Liwicki-Everingham-09,Ricco-Tomasi-09}.  Letter error rates of 10\% or less have been achieved when the recognition is constrained to a small (up to 100-word) lexicon of allowed sequences.
Our work is the first of which we are aware to address the task of {\it unrestricted} fingerspelling sequence recognition, and to explicitly study the contrast between signer-dependent and signer-independent recognition.

The segmental models we propose are related to prior work in both vision and speech.  For speech recognition, segmental conditional random fields (SCRFs) and their variants have been applied fairly widely~\cite{zweig,tang2014comparison,tang2015, zweig2011speech,zweig2012classification,he2012efficient}.
In natural language processing, semi-Markov CRFs have been used for named entity recognition~\cite{sarawagisemi}, where the labeling is binary.
Finally, segmental models have been applied to vision
tasks such as classification and segmentation of action sequences~\cite{shi2011human,duong2005activity} with a small set of possible activities to choose from, including work on spotting of specific (non-fingerspelled) signs in sign language video~\cite{Choetal2009}.
One aspect that our work shares with the speech recognition work is that we have a relatively large set of labels (26 letters plus non-letter ``N/A'' labels), and widely varying lengths of segments corresponding to each label (our data includes segment durations anywhere from 2 to 40 frames), which makes the search space larger and the recognition task more difficult than in the vision and text tasks to which such models have been applied.
In prior speech recognition work, this computational difficulty has
often been addressed by adopting a lattice rescoring approach, where a
frame-based model such as an HMM system generates first-pass lattices
and a segmental model rescores them~\cite{zweig,zweig2011speech}.  We
compare this approach to an efficient first-pass segmental model~\cite{tang2015}.

For the challenging signer-independent setting in which the signers in the training and test data are different, we may consider domain adaptation \cite{ben2010theory}. Due to differences in data distributions of source and target domain, the performance of the trained model can suffer and domain adaptation approaches have been popular in this setting. Applications include natural language processing \cite{daume2009frustratingly, blitzer2006domain, jiang2007instance}, computer vision \cite{saenko2010adapting,Donahue_2013_CVPR,gopalan2011domain} and speech recognition \cite{acero2000hmm,gales1998maximum,leggetter1995maximum}. We consider adapting our trained models toward a new test signer. Since our application and models are most similar to those in speech recognition, we base our signer adaptation approaches on those used for speaker adaptation in speech recognition.  The classic techniques include maximum likelihood linear regression~\cite{leggetter1995maximum} and maximum a posteriori~\cite{lee1993speaker}.  In recent years, however, speech recognition systems are typically based on deep neural networks (DNNs), and speaker adaptation approaches specific to DNNs have been developed~\cite{liao2013speaker,abdel2013fast,swietojanski2014learning,doddipatla2014speaker}.  We therefore base our adaptation techniques for our DNN-based models on these approaches.  More details are given in Section \ref{sec:dnn}.

\chapter{Recognition methods}
\label{sec:method}

Our task is to take as input a video (a sequence of images)
corresponding to a fingerspelled word, as in Figure~\ref{fig:ex}, and
predict the signed letters.  This is a sequence prediction task
analogous to connected phone or word recognition, but there are some
interesting sign language-specific properties to the data domain.  For
example, one striking aspect of fingerspelling sequences, such as
those in Figure~\ref{fig:ex}, is the large amount of motion and lack
of any prolonged ``steady state'' for each letter.  Typically, each
letter is represented by a brief ``peak of articulation'' of one or a
few frames, during which the hand's motion is at a minimum and the
handshape is the closest to the target handshape for the letter.  This
peak is surrounded
by longer period of motion between the current letter and the
previous/next letters.

We consider
signer-dependent, signer-independent, and signer-adapted recognition.
We next describe
the recognizers we compare, as well as the techniques we explore for
signer adaptation.  All of the recognizers use deep neural network
(DNN) classifiers of letters or handshape features.

\input{ex_tulip}
\input{ex_art}

\section{Recognizers}
In designing recognizers, we keep several considerations in mind.
First, the data set, while large by sign language research standards,
is still quite small compared to typical speech data sets.  This means
that large models with many context-dependent units are infeasible to
train on our data (as confirmed by our initial experiments).  We
therefore restrict attention here to ``mono-letter'' models, that is
models in which each unit is a context-independent letter.  We also
consider the use of articulatory (phonological and phonetic) feature
units, as there is evidence from speech recognition research that
these may be useful in low-data
settings~\cite{livescu2007articulatory,stuker2003integrating,cetin}.  Second, we would
like our models to be able to capture rich sign language-specific
information, such as the dynamic aspects of fingerspelled letters as
discussed above; this suggests the segmental models that we consider
below.  Finally, we would like our models to be easy to adapt to new
signers.  In order to enable this, all of our recognizers use independently trained deep
neural network (DNN) classifiers, which can be adapted and plugged
into different sequence models.  Our DNNs are trained using an
L2-regularized cross-entropy loss.  The inputs are the image features
concatenated over a multi-frame window centered at the current frame, 
which are fed through several fully connected layers followed by a
softmax output layer with as many units as labels.  

\subsection{Tandem model}
\label{sec:tandem}
The first recognizer we consider is based on the popular tandem approach to speech recognition~\cite{ellis}.  In tandem-based speech recognition, Neural Networks (NN) are trained to classify phones, and their outputs (phone posteriors) are post-processed and used as observations in a standard HMM-based recognizer with Gaussian mixture observation distributions.  The post-processing may include taking the logs of the posteriors (or simply taking the linear outputs of the NNs rather than posteriors), applying principal components analysis, and/or appending acoustic features to the NN outputs.

In this work, we begin with a basic adaptation of the tandem approach, where instead of phone posteriors estimated from acoustic frames, we use letter posteriors estimated from image features.  We also propose to use classifiers of phonological features of fingerspelling.  The motivation is that, since features have fewer values, it may be possible to learn them more robustly than letters from small training sets, and certain features may be more or less difficult to classify.  This is similar to the work in speech recognition of \c{C}etin et al.~\cite{cetin}, who used articulatory feature NN classifiers rather than phone classifiers.

We use a phonological feature set developed by Brentari~\cite{bren}, who proposed seven features for ASL handshape.  Of these, we use the six that are contrastive in fingerspelling.  The features and their values are given in Table~\ref{t:features}.  
 Example frames for values of the ``SF thumb'' feature are shown in Figure~\ref{f:feature_ex}, and entire phonological feature vectors for several letters are shown in Appendix, Figure~\ref{f:tandem_example}.

\begin{table}[t]
\begin{center}
\begin{tabular}{|l|l|}
\hline
{\bf Feature} & {\bf Definition/Values}  \\ \hline \hline
SF point of & side of the hand where \\ 
reference & SFs are located  \\ \cline{2-2} 
(POR) & {\it SIL, radial, ulnar, radial/ulnar} \\ \hline\hline
SF joints & degree of flexion or \\
& extension of SFs  \\  \cline{2-2}
& {\it SIL, flexed:base, flexed:nonbase,} \\ & {\it flexed:base \& nonbase,} \\
& {\it stacked, crossed, spread} \\ \hline\hline
SF quantity &combinations of SFs \\ \cline{2-2}
& {\it N/A, all, one,} \\ & {\it one $>$ all, all $>$ one} \\ \hline\hline
SF thumb & thumb position  \\ \cline{2-2}
& {\it N/A, unopposed, opposed} \\ \hline\hline
SF handpart & internal parts of the hand  \\ \cline{2-2}
& {\it SIL, base, palm, ulnar} \\ \hline\hline
UF & open/closed \\ \cline{2-2}
& {\it SIL, open, closed} \\ \hline
\end{tabular}
\caption[Definition and possible values for phonological features]{Definition and possible values for phonological features based on~\cite{bren}.  The first five features are properties of the active fingers ({\it selected fingers}, SF); the last feature is the state of the inactive or {\it unselected fingers} (UF).  In addition to Brentari's feature values, we add a SIL (``silence'') value to the features that do not have an N/A value.  For detailed descriptions, see~\cite{bren}.}
\label{t:features}
\end{center}
\end{table}

\begin{figure}[h!]
\begin{center}
        \hspace{-.5em}\includegraphics[width=0.65\linewidth]{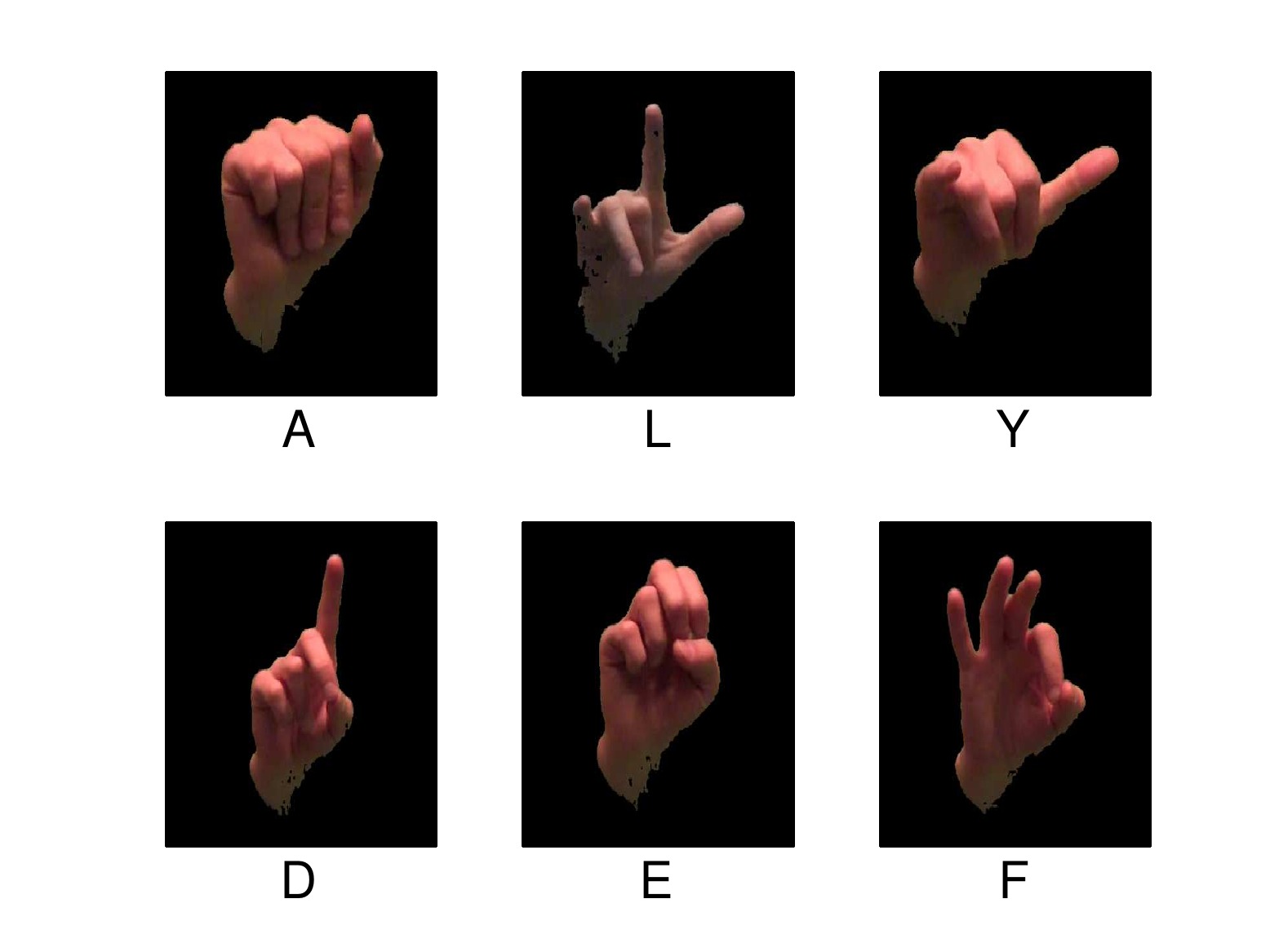}
\end{center}
\caption[Example images corresponding to SF thumb]{Example images corresponding to SF thumb = 'unopposed' (upper row) and SF thumb = 'opposed' (bottom row). More examples of phonological features are given in Appendix, Figure \ref{f:ex_ph_features}.}
\label{f:feature_ex}
\end{figure}


Frame-level image features are fed to seven DNN classifiers, one of which predicts
the frame's letter label and six others which predict handshape
phonological features.  The classifier outputs and image features are applied a dimensionality reduction via PCA, respectively, and concatenated with each other.  The concatenated features form the observations in an HMM-based recognizer with Gaussian mixture
observation densities.  We use a 3-state HMM for each
(context-independent) letter, plus one HMM each for initial and final
``silence'' (non-signing segments). Figure ~\ref{f:tandem_example} shows an example of the operation of the tandem model recognizers.

In addition, we use a bigram letter language model.  In general the language
model of fingerspelled letter sequences is difficult to define or
estimate, since fingerspelling does not follow the same distribution
as English words and there is no large database
of natural fingerspelled sequences on which to train. In addition, in our
data set, the words were selected so as to maximize
coverage of letter n-grams and word types rather than
following a natural distribution.  For this work, 
the language model is trained using ARPA CSR-III text, which
includes English words and names~\cite{csr-III}.  
The issue of language modeling for fingerspelling deserves more
attention in future work.

\begin{figure*}[h!]
\begin{center}
\begin{tabular}{c}
\hspace{-.35in}\includegraphics[width =1\linewidth]{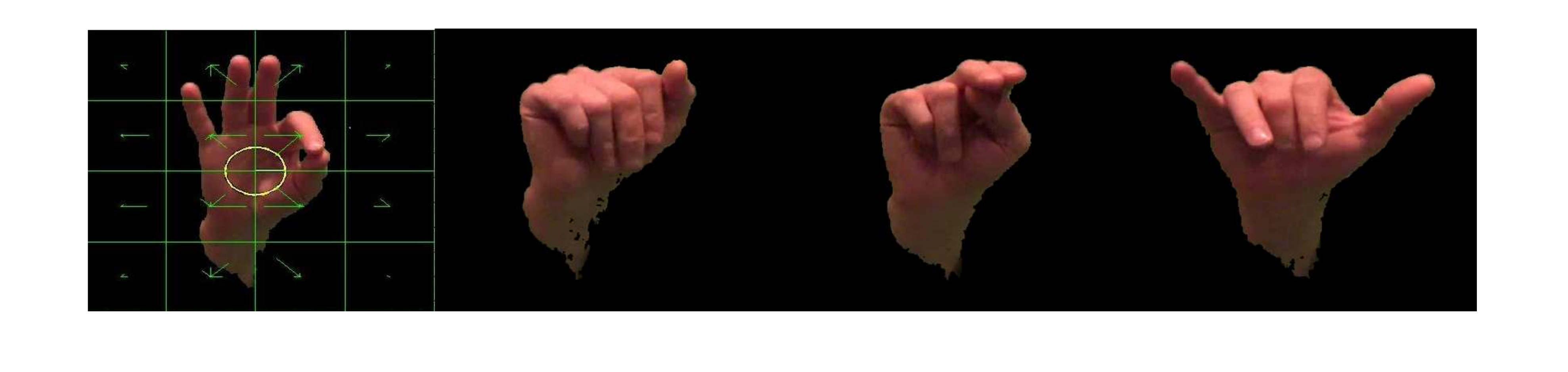} \\ [-3ex]
\hspace{-.1in}\includegraphics[width =1\linewidth]{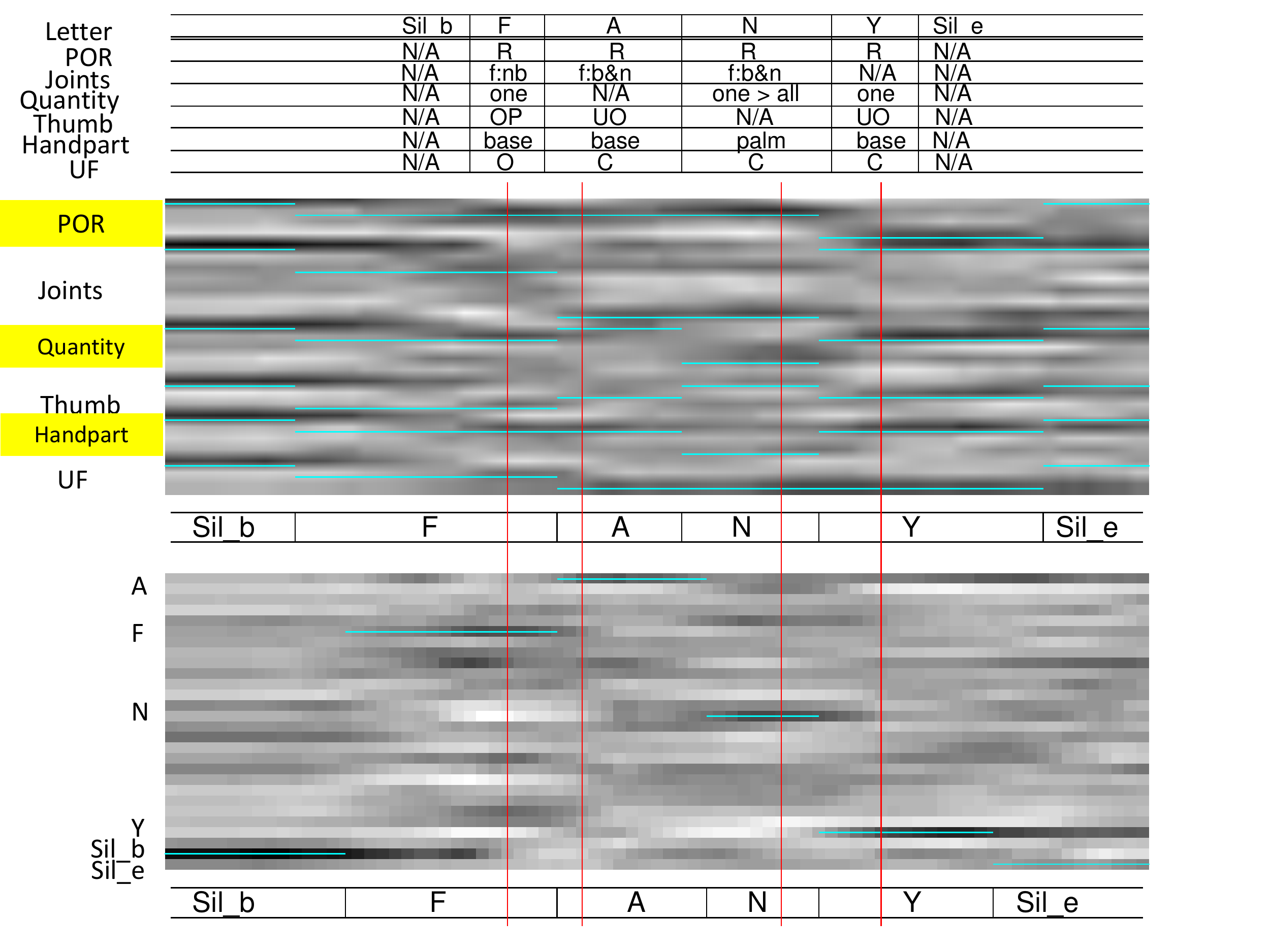}
\end{tabular}
\end{center}
\vspace{-.25in}
\caption[Example of the operation of the tandem model recognizers with phonological feature vectors]{Several components of fingerspelling recognition with feature- and letter-based tandem models, for an example of the sequence 'F-A-N-Y'.  Top: apogee image frames of the four letters (we use all frames, but only the apogees are shown).  The first image shows single-depth SIFT features (lengths of arrows show strengths of image gradients in the corresponding directions).  Middle: ``ground-truth'' letter and feature alignments derived from the manually labeled apogees.  POR = point of reference.  UF = unselected fingers.  Red vertical lines:  manually labeled apogee times.  Bottom:  NN posteriors (darker = higher) for features (upper matrix) and letters (lower matrix), and output hypotheses from feature-based and letter-based tandem models, respectively.  Horizontal cyan bars: decoded hypothesis (in the feature case, the letter hypothesis is mapped to features).}
\label{f:tandem_example}
\end{figure*}

\subsection{Rescoring segmental CRF}
\label{sec:re_scrf}

The second recognizer is a segmental CRF (SCRF).  SCRFs~\cite{sarawagisemi,zweig} are conditional log-linear models with feature functions that can be based on variable-length segments of input frames, allowing for great flexibility in defining feature functions.  
We use an SCRF to rescore lattices produced by a baseline frame-based
recognizer (in this case, the tandem model above).


We begin by defining the problem and our notation.  Let the sequence of visual observations for a given video (corresponding to a single word) be $O = o_1, \ldots, o_T$, \karenedit{where each $o_t$ is a multidimensional image descriptor for frame $t$}.  Our goal is to predict the label (letter) sequence.  \karenedit{Ideally we would like to predict the best label sequence, marginalizing out different possible label start and end times, but in practice we use the typical approach of predicting the best sequence of frame labels $S = s_1, \ldots, s_T$}.  We predict $S$ by maximizing its conditional probability under our model, $\hat{S} = \text{argmax}_S P(S|O)$.
In generative models like HMMs, we have a joint model $P(S,O)$ and we make a prediction using Bayes' rule.
In conditional models we directly represent the conditional distribution $P(S|O)$.  For example, in a typical linear-chain CRF, we have:
\[
p(S|O) = \frac{1}{Z(O)}\exp\left(\sum_{v,k}\lambda_k f_k(S_v,O_v) + \sum_{e,k}\mu_k g_k(S_e)\right)
\]
where $Z(O)$ is the partition function, $f_k$ are the ``node'' feature
functions that typically correspond to the state in a single frame
$S_v$ and its corresponding observation $O_v$, $g_k$ are ``edge''
feature functions corresponding to inter-state edges, \karenedit{$e$
  ranges over pairs of frames and $S_e$ is the pair of states
  corresponding to $e$}, and $\lambda_k$ and $\mu_k$ are the 
weights.  

It may be more natural to consider feature functions that span entire
segments corresponding to the same label.  Semi-Markov
CRFs~\cite{sarawagisemi}, also referred to as segmental
CRFs~\cite{zweig} or SCRFs, provide this ability.  

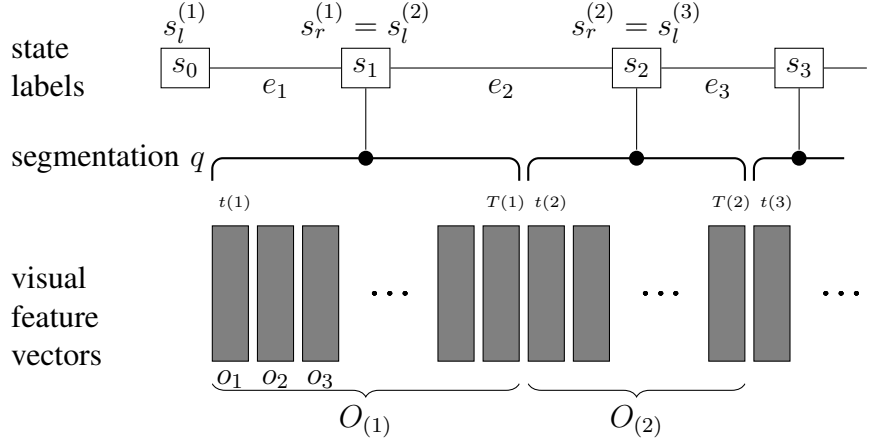
\begin{figure*}
  \centering
  \begin{tikzpicture}[scale=.6]
    \foreach \t in {1,2,3,6,7,8,9,12,13}{
      \draw[fill=gray] (\t,0) rectangle ++(.8,-3);
      \node at (5,-1.5) {\large \ldots};
      \node at (11,-1.5) {\large \ldots};
      \node at (15,-1.5) {\large \ldots};
    }
    \node at (1.4,-3.4) {$o_1$};
    \node at (2.4,-3.4) {$o_2$};
    \node at (3.4,-3.4) {$o_3$};
    \node[text width=7em] (labelvis) at (-1,-2) {visual\\feature\\vectors};
    \coordinate (s1) at (1,1.5);
    \coordinate (e1) at (7.8,1.5);
    \coordinate (s2) at (8,1.5);
    \coordinate (e2) at (12.8,1.5);
    \coordinate (s3) at (13,1.5);
    \draw[thick,rounded corners] ($(s1) - (0,.5)$) -- ++(0,.5) --
    (e1)
    node[midway,circle,fill=black,inner sep=2pt,name=seg1] {}  -- ++(0,-.5);
    \draw[thick,rounded corners] ($(s2) - (0,.5)$) -- ++(0,.5) -- (e2)
    node[midway,circle,fill=black,inner sep=2pt,name=seg2] {} -- ++(0,-.5);
    \draw[thick,rounded corners] ($(s3) - (0,.5)$) -- ++(0,.5) -- ++(2,0)
    node[midway,circle,fill=black,inner sep=2pt,name=seg3] {};
    \draw[decorate,decoration={brace,amplitude=5pt}] ($(e1) - (0,5)$)
    -- ($(s1) - (0,5)$) node [midway,below=4pt] {$O_{(1)}$};
    \draw[decorate,decoration={brace,amplitude=5pt}] ($(e2) - (0,5)$)
    -- ($(s2) - (0,5)$) node [midway,below=4pt] {$O_{(2)}$};
    \node[draw] (state1) at ($(seg1) + (0,2)$) {$s_1$};
    \node[draw] (state0) at ($(state1) - (4,0)$) {$s_0$};
    \node[draw] (state2) at ($(seg2) + (0,2)$) {$s_2$};
    \node[draw] (state3) at ($(seg3) + (0,2)$) {$s_3$};
    \path let \p1=(labelvis), \p2=(state0) in node[text width=7em]  at(\x1,\y2)
 {state\\labels};
    \draw (state1) -- (seg1);
    \draw (state2) -- (seg2);
    \draw (state3) -- (seg3);
    \path let \p1=(labelvis), \p2=(s1) in node[text width=7em]
    at(\x1,\y2)  {segmentation $q$};
    \draw (state0) -- (state1) node[midway,below] {$e_1$};
    \draw (state1)--(state2) node[midway,below] {$e_2$};
    \draw (state2)--(state3) node[midway,below] {$e_3$};
    \draw (state3) -- ++(1.5,0);
    \path (state0) -- ++(0,1) node {$s_l^{(1)}$};
    \path (state1) -- ++(0,1) node {$s_r^{(1)}=s_l^{(2)}$};
    \path (state2) -- ++(0,1) node {$s_r^{(2)}=s_l^{(3)}$};
    {\tiny
    \node at (1.5,.5) {$t(1)$};
    \node at (7.5,.5) {$T(1)$};
    \node at (8.5,.5) {$t(2)$};
    \node at (12.5,.5) {$T(2)$};
    \node at (13.5,.5) {$t(3)$};
  }
  \end{tikzpicture}

    \caption[Illustration of SCRF notation]{Illustration of SCRF notation. For example, edge $e_2$ is
associated with \karenedit{the ``left state'' $s_l^{(2)}$, the ``right state'' $s_r^{(2)}$, and the 
    segment of observations $O_{(2)}$ spanning frames $t(2)$ through
    $T(2)$.} 
}
  \label{fig:s-crf}
\end{figure*}

\karenedit{Figure~\ref{fig:s-crf} illustrates the SCRF notation, which we now describe.}
In a SCRF, we consider the segmentation to be a latent variable and sum over all possible segmentations of the observations corresponding to a given label sequence to get the conditional probability of the label sequence \karenedit{$S = s_1, \ldots, s_L$, where the length of $S$ is now the (unknown) number of distinct labels $L$}:
\[
p(S|O) = \frac{\sum_{q ~\text{s.t.}~ |q|=|S|} \exp\left(\sum_{e \in q,k} \lambda_k f_k (s_l^e,s_r^e,O_e)\right)}{\sum_{S^\prime} \sum_{q~\text{s.t.} |q|=|S^\prime|} \exp\left(\sum_{e \in q,k} \lambda_k f_k (s_l^e,s_r^e,O_e)\right)}
\]
Here, $S'$ ranges over all possible state (label) sequences, $q$ is a segmentation of the observation sequence whose length (number of segments) must be the same as the number of states in $S$ \karenedit{(or $S^\prime$)}, $e$ ranges over all state pairs in $S$, $s_l^e$ is the state which is on the left of an edge, $s_r^e$ is the state on the right of an edge, and $O_e$ is the \karenedit{multi-frame} observation segment associated with \karenedit{$s_r^e$}.
In our work, we use a tadem HMM frame-based recognizer to generate a set of candidate segmentations of $O$, and sum only over those candidate segmentations.  In principle the inference over all possible segmentations can be done, but typically this is only feasible for much smaller search spaces than ours.

We define several types of feature functions, some of which are quite general to \karenedit{sequence recognition tasks and some of which are tailored to fingerspelling recognition}:

\subsubsection{Language model \karenedit{feature}} 
The language model \karenedit{feature is a smoothed bigram probability} of the letter pair corresponding to an edge:
\[
f_{lm}(s_l^e,s_r^e,O_e)=p_{LM}(s_l^e,s_r^e).
\]

\subsubsection{Baseline consistency feature}
To take advantage of the existence of a high-quality baseline, we use a baseline feature like the one introduced by~\cite{zweig}.  This feature is constructed using the 1-best output hypothesis from an HMM-based baseline recognizer.  The feature value is 1 when a segment spans exactly one letter label hypothesized by the baseline and the label matches it:
\[
f_{b}(s_l^e,s_r^e,O_e)=\left\{
\begin{array}{rcl}
    +1 & \text{if}~C(t(e),T(e))=1, \\
    & \phantom{\text{if}} \text{and}~B(t(e),T(e))=w(s_r^e) \\
    -1 & \text{otherwise} \\
\end{array}
\right.
\]
where $t(e)$ and $T(e)$ are the start and end times corresponding to edge $e$, $C(t,T)$ is the number of distinct baseline labels in the time span from $t$ to $T$, $B(t,T)$ is the label corresponding to time span $(t,T)$ when $C(t,T) = 1$, and $w(s)$ is the letter label of state $s$.

\subsubsection{Handshape classifier-based feature functions} 
The next set of feature functions measure the degree of match between the intended segment label and the appearance of the frames within the segment.  For this purpose we use a set of frame classifiers, each of which classifies either letters or linguistic handshape features.
As in Section \ref{sec:tandem}, we use the linguistic handshape feature set developed by Brentari~\cite{bren}, who proposed seven features to describe handshape in ASL.  Each such linguistic feature (not to be confused with feature functions) has 2-7 possible values.  Of these, we use the six that are contrastive in fingerspelling (See Table~\ref{t:features} for the details.).
For each linguistic feature or letter, we train a classifier that produces a score for each feature value for each video frame.  We also train a separate letter classifier.  Specifically, we use deep neural network (DNN) classifiers.

\paragraph{\bf Feature functions}
Let $y$ be a letter and $v$ be the value of a linguistic feature or 
letter, \karenedit{$N_{e} = |O_e| = T(e)+1-t(e)$} the length of \karenedit{an observation segment corresponding to edge $e$}, and
\karenedit{$g(v|o_i)$ the output of a DNN classifier at frame $i$ corresponding
to class $v$}. We define  
\begin{itemize}
  \item $\textrm{mean}$: $f_{yv}(s_l^e,s_r^e,O_e)=  \delta(w(s_r^e)=y)\cdot\frac{1}{\karenedit{N_{e}}}\sum_{i=t(e)}^{T(e)} g(v|o_i)$
  \item $\textrm{max}$: $f_{yv}(s_l^e,s_r^e,O_e)=  \delta(w(s_r^e)=y)\cdot \max_{i \in (t(e),T(e))} g(v|o_i)$
  \item $\textrm{div$_s$}$: a concatenation of three $\textrm{mean}$ feature functions, each computed over a third of the segment
  \item $\textrm{div$_m$}$: a concatenation of three $\textrm{max}$ feature functions, each computed over a third of the segment
\end{itemize}


\subsubsection{Peak detection features} \label{sec:peaks}
Fingerspelling a sequence of letters yields a corresponding sequence of 
``peaks'' of
articulation. Intuitively, these are frames in which the hand reaches the
target handshape for a
particular letter.
The peak frame and the frames around it for each letter tend to be
characterized by very little motion as the transition to the current
letter has ended while the transition to the next letter has not yet begun,
whereas the transitional frames between letter peaks have more motion. To use this information and encourage each \karenedit{predicted} letter segment to have a single peak, we define letter-specific ``peak detection features'' as follows.
We first compute approximate derivatives of the visual descriptors, consisting of the $l_2$ norm of the difference between descriptors in every pair of consecutive frames, smoothed by averaging over 5-frame windows.  We 
\karenedit{expect there to be} a single local minimum in this approximate derivative function over the span of the segment.  Then we define the feature function corresponding to each letter $y$ as

\[f_{y}^{\textrm{peak}}(s_l^e,s_r^e,O_e)=\delta(w(s_r^e)=y)\cdot\delta_{\textrm{peak}}(O_e)\]

\noindent where $\delta_{\textrm{peak}}(O_e)$ is $1$ if there is only one local minimum in the segment $O_e$ and 0 otherwise.





\subsection{First-pass segmental CRF}
One of the drawbacks of a rescoring approach is that the quality of the
final outputs depends both on the quality of the baseline lattices and
the fit between the segmentations in the baseline lattices and those
preferred by the second-pass model.  We therefore also consider a
first-pass segmental model, using similar features to the rescoring model.
In particular, we also use a first-pass SCRF inspired by the phonetic
recognizer of Tang {\it et al.}~\cite{tang2015}.
We use the same feature functions as
in~\cite{tang2015}, namely average DNN outputs over each segment,
samples of DNN outputs within the segment, boundaries of DNN outputs in each segment, duration and bias, all
lexicalized. 

\section{DNN adaptation}
\label{sec:dnn}

\begin{figure*}[th]
\centering
\includegraphics[width=1.0\linewidth]{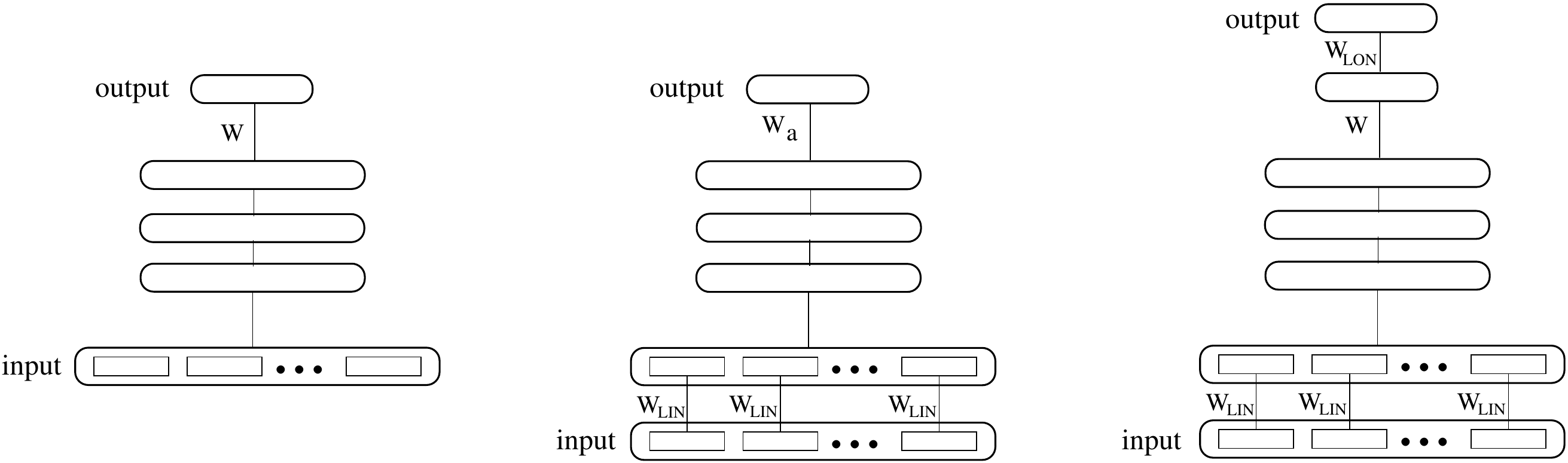}
\vspace{-.1in}
\caption[DNN adaptation approaches]{Left:  Unadapted DNN classifier; middle:  adaptation via linear input network and output layer updating (LIN+UP); right: adaptation via linear input network and linear output network (LIN+LON).}
\label{fig:dnn-adapt}
\end{figure*}

In experiments, we will consider both signer-dependent and
signer-independent recognition.  In the latter case, the test signer
is never seen in training.  As we will see, there is a very large gap
between signer-dependent and signer-independent recognition on our
data.  Inspection of data such as Fig.~\ref{fig:ex} and \ref{fig:ex2} reveals some sources of
signer variation, including differences in speed, hand appearance, and
non-signing motion before and after signing.  The speed variation is
large, with a factor of $1.8$ between the fastest and slowest signers.
In the absence of adaptation data, we consider a simple speed
normalization:  We augment the training data with resampled image
features, at 0.8x and 1.2x the original frame rate. 

If we have access to some labeled data from the test signer, but not a
sufficient amount for training full signer-specific models, we can
consider adapting a signer-independent model toward the test signer.  The most straightforward form of
adaptation for our models is to only adapt the DNN classifiers, and
then use the adapted ones in pre-trained signer-independent models.
The adaptation can use either ground-truth frame-level labels (given
by human annotation) or, if only word labels are available, we can
obtain frame labels via forced alignment using the signer-independent model.

A number of DNN adaptation approaches have been developed for speech
recognition
(e.g.,~\cite{liao2013speaker,abdel2013fast,swietojanski2014learning,doddipatla2014speaker}).
We consider several approaches, shown in Figure~\ref{fig:dnn-adapt}.
Two of the approaches are based on linear input networks (LIN) and
linear output networks (LON)~\cite{Neto_95a,Yao_12a,li2010}.  In these
techniques most of the network parameters are fixed; only a limited
set of weights at the input and/or output layers are learned. 

In the approach we refer to as LIN+UP in Figure~\ref{fig:dnn-adapt}, we apply a single affine transformation $W_{\text{LIN}}$ to the 
static features at each frame (before concatenation)
and use the transformed features as input to the trained signer-independent DNNs. We jointly learn $W_{\text{LIN}}$ and adapt the last (softmax) layer weights by minimizing the same cross-entropy loss
on the adaptation data.
The softmax layer adaptation is achieved by ``warm-starting'' with the learned signer-independent weights.

The second approach, referred to as LIN+LON in
Figure~\ref{fig:dnn-adapt}, uses the same adaptation layer at the
input.  However, instead of adapting the softmax weights at the top
layer, it removes the softmax output activation and adds a new softmax
output layer $W_{\text{LON}}$ trained for the test signer.
The new input and output layers are trained jointly with the same
cross-entropy loss. 

Finally, we also consider adaptation by fine-tuning all of the DNN
weights on adaptation data, starting from the signer-independent DNN weights.

\chapter{Experimental Results}


We report on experiments using the fingerspelling data from four
native ASL signers.  We begin by describing data and annotation, some of the
front-end details of hand segmentation and feature extraction,
followed by experiments with the frame-level DNN classifiers
(Sec.~\ref{sec:dnn-exp}) and letter sequence recognizers (Sec.~\ref{sec:LER-exp}).\\

\section{Data and annotation}
\label{sec:data}

\begin{figure}[!th]
  \centering
  \begin{tabular}{cc}
    \includegraphics[width=.48\linewidth]{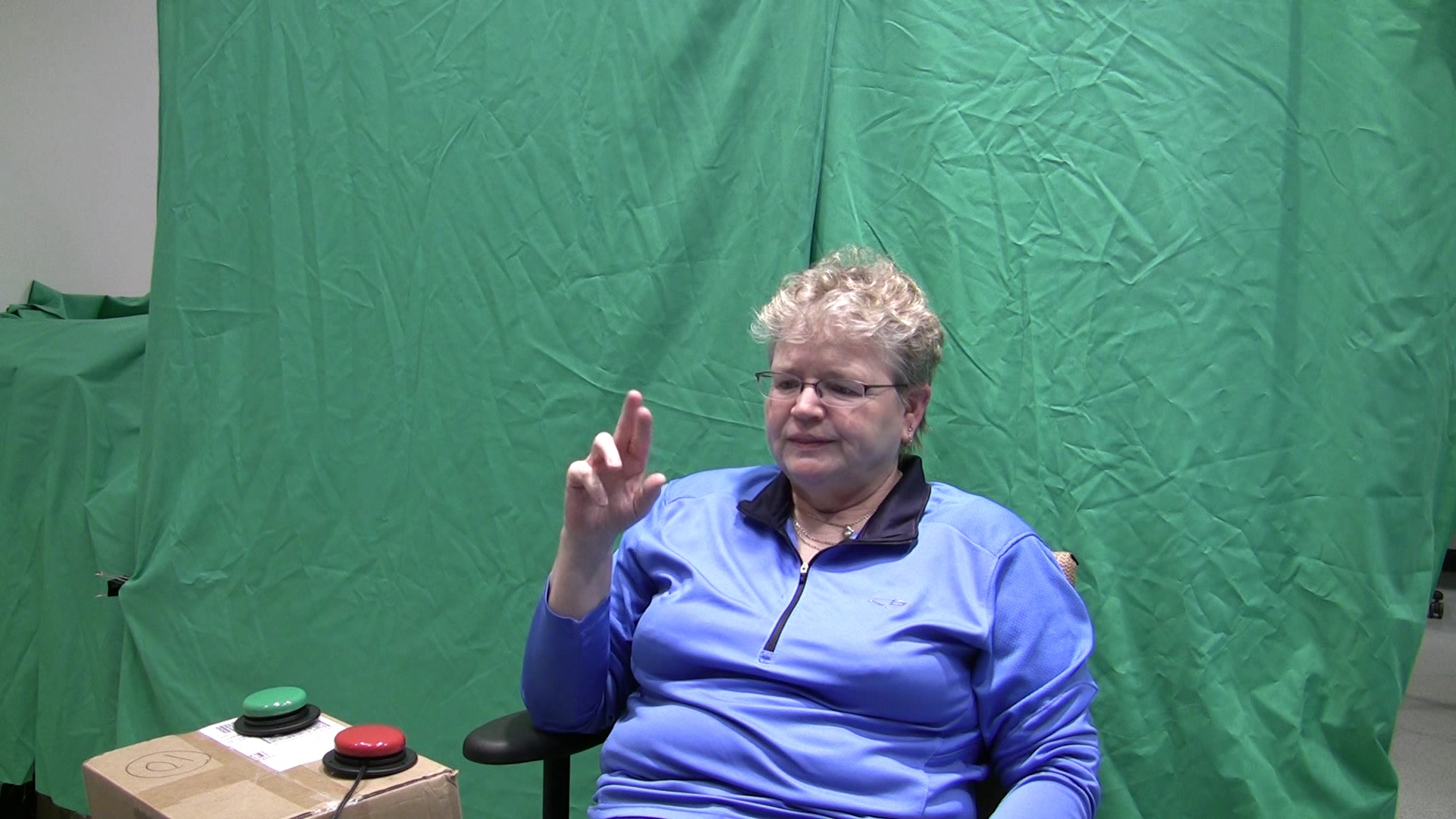}&
    \includegraphics[width=.48\linewidth]{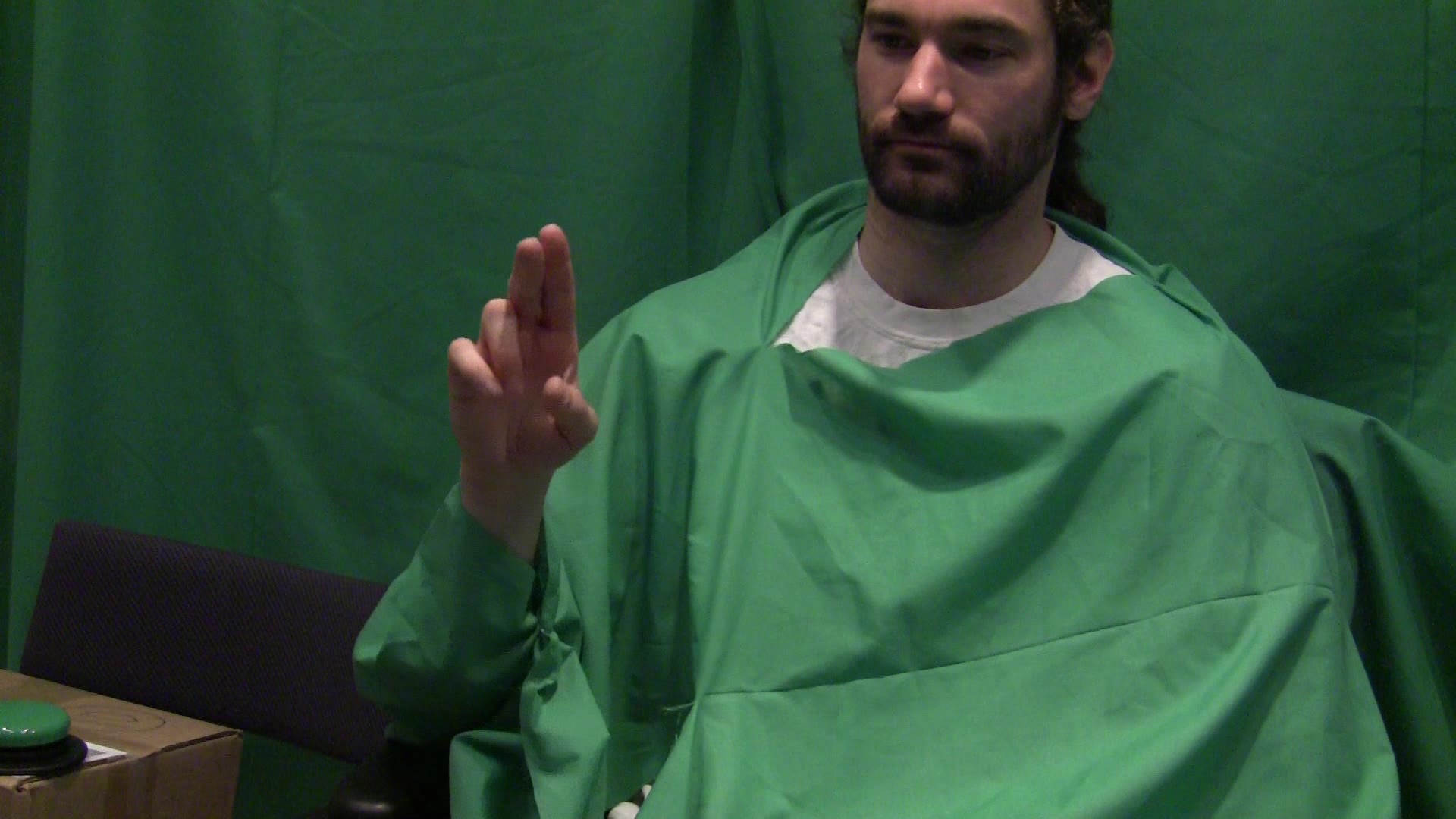}\\
    \includegraphics[width=.48\linewidth]{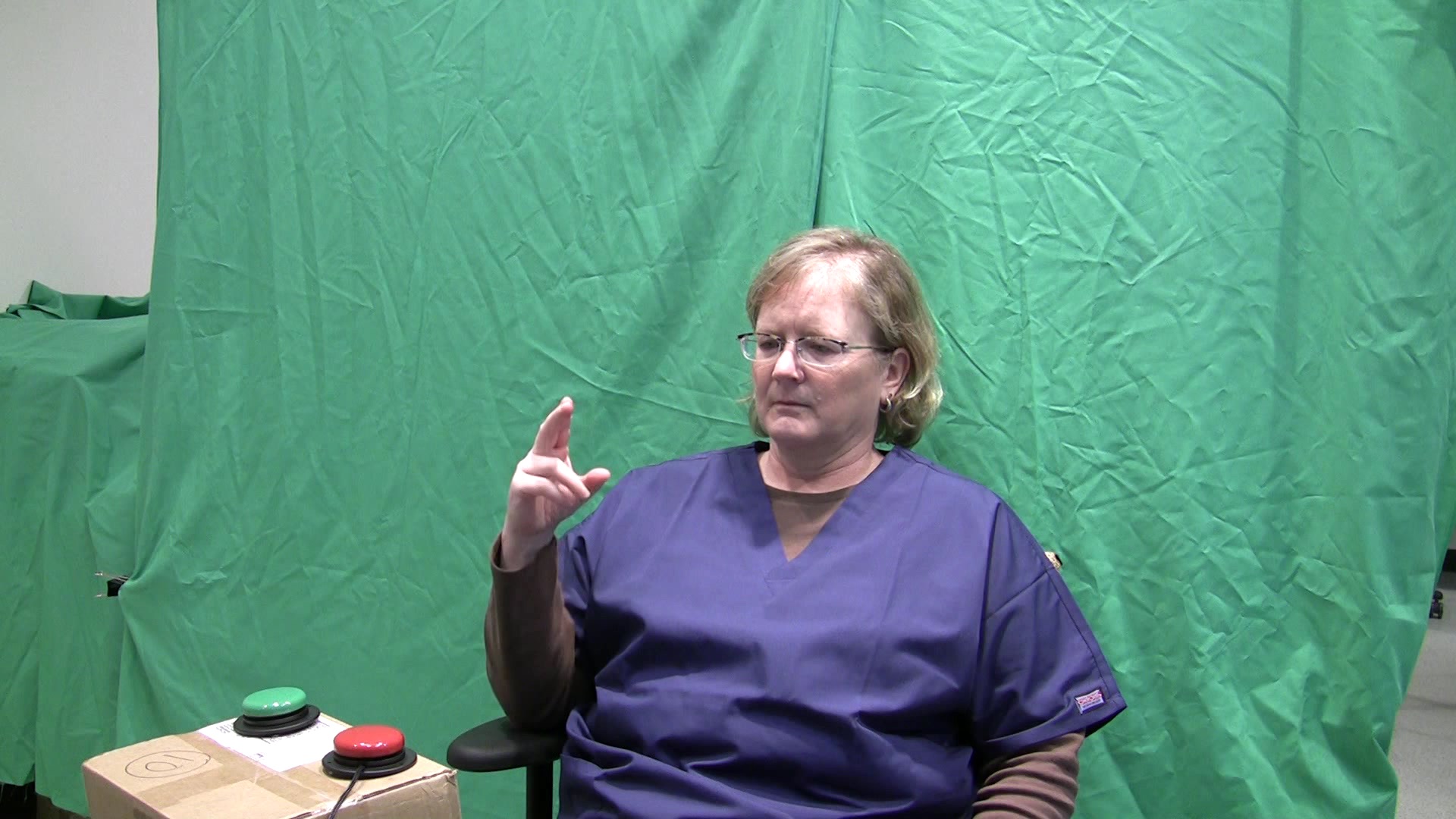}&
    \includegraphics[width=.48\linewidth]{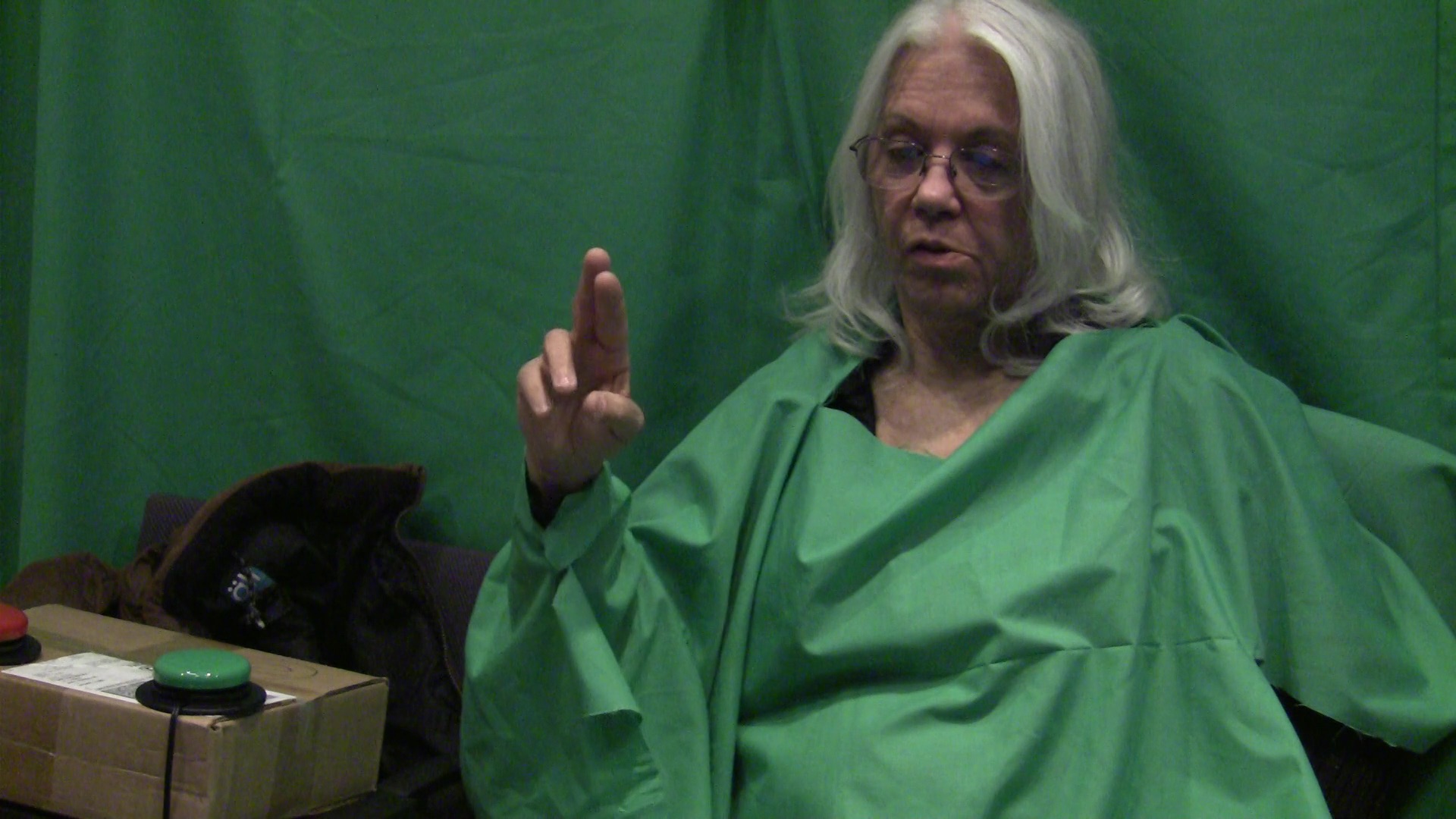}
  \end{tabular}
  \caption{Example video frames from the four signers.}
  \label{fig:frames}
\end{figure}

We use video recordings of four native ASL signers\footnote{This section includes material previously published in~\cite{kim2013}.}. 
The data were recorded at 60 frames per second in a studio environment. 
Each signer signed a list of 300 words as they appeared on a computer screen in front of the signer. 
There were two non-overlapping lists of 300 words (one for signers 1 and 2, the other for signers 3 and 4. For the complete lists, see Table \ref{t:wordlist1} and \ref{t:wordlist2}). 
Each word was spelled twice, yielding 600 word instances signed by each signer. 
The lists contained English and foreign words, including proper names and common English nouns. The total number of frames in four signers' data is $\sim$ 350k frames.
The recording settings, including differences in environment and
camera placement across recording sessions, are illustrated in Figure~\ref{fig:frames}.

The signers sat at a chair, and the screen showing the word to be signed was in front of them. Each word was repeated twice, and the signers were asked to press the green button if they correctly signed the word, otherwise the red button. If the green button was pressed, the screen would move to the next (the same word if not repeated yet, or the next word in the word list). The start and end of each word were indicated by pressing a button, allowing automatic partition of the recording into a separate video for every word. We recorded six sessions with conversational and fluent speed. To synchronize the video and word signed, every video was verified and manually labeled by multiple annotators with the times and letter identities of the peaks of articulation (see Sec.~\ref{sec:peaks}). The peak annotations are used for the training portion of the data in each experiment to segment a word into letters (the boundary between consecutive letters is defined as the midpoint between their peaks). For more details about data and annotation steps, please refer to \cite{keane_NELS}. \\

\subsubsection{Hand localization and segmentation}
For every signer, we trained a model for hand detection similar to that used in~\cite{kim2012,Liwicki-Everingham-09}. 
Using manually annotated hand regions, marked as polygonal regions of interest (ROI) in 30 frames, we fit a mixture of Gaussians $P_{hand}$ to the color of the
hand pixels in L*a*b color space. 
Using the same 30 frames, we also built a single-Gaussian color model $P_{bg}^x$ for every pixel $x$ in the image excluding pixel values in or near marked hand ROIs. 
Then, given a test frame, we label each pixel as hand or background based on an
odds ratio:  Given the color triplet $\mathbf{c}_x=[l_x,a_x,b_x]$ at pixel $x$, we assign it to hand if
\begin{equation}
  \label{eq:handodds}
  P_{hand}(\mathbf{c}_x)\pi_{hand}\,>\,
  P_{bg}^x(\mathbf{c}_x)(1-\pi_{hand}),
\end{equation}
where the prior $\pi_{hand}$ for hand size is estimated from the same 30
training frames.

Since this simple model produces rather noisy output, we next clean
it up by a sequence of filtering steps.  We suppress pixels that fall
within regions detected as faces by the Viola-Jones face detector~\cite{violajones},
since these tend to be false positives.  We also suppress pixels that
passed the log-odds test but have a low estimated value of $P_{hand}$.
These tend to correspond to movements in the scene, e.g., a signer
changing position and thus revealing previously occluded portions of the background; for such pixels the value of $P_{bg}$ may be low,
but so is $P_{hand}$. Finally, we suppress pixels outside of a
(generous) spatial region where the
signing is expected to occur. The largest surviving connected
component of the resulting binary map is treated as a mask that
defines the detected hand region. Some examples of resulting hand
regions are shown in
Figures~\ref{f:recog3} ( and more in Figure ~\ref{fig:recog1}, Figure ~\ref{f:recog2} and Figure ~\ref{f:recog4}). Note that while this procedure currently requires
manual annotation for a small number of frames in our offline
recognition setting, it could be fully
automated in a realistic interactive setting, by asking the subject
to place his/her hand in a few defined locations for calibration.\\

\subsubsection{Handshape descriptors} 
For most experiments, we use histograms of oriented
gradients (HOG~\cite{HOG}) as the visual descriptor
(feature vector) for a given hand region\footnote{In initial experiments we use SIFT descriptors, described in Section \ref{sec:pre-exp}. 
Their performances are similar but HoG feature is slightly better.}.
We first resize the tight bounding box of the hand region to a
canonical size of 128$\times$128 pixels, and then compute HOG features
on a spatial pyramid of regions, 4$\times$4,
8$\times$8, and 16$\times$16 grids,  with eight orientation bins per
grid cell, resulting in 2688-dimensional descriptors.  Pixels outside of the
hand mask are ignored in this computation. For HMM-based recognizer, to speed
up computation, these descriptors were projected to at most 200
principal dimensions; the 
exact dimensionality in each experiment was tuned on a development set. For DNN frame classifiers, we found that finer grids did not improve much with increasing complexities, so we use 128-dimensional descriptors.\\

\section{An initial experiment:  tandem models and shallow classifiers}
\label{sec:pre-exp}
Before proceeding to our main experiments, we first analyze the linguistic features we proposed with a preliminary experiment by using two native ASL signers' data\footnote{This section includes material previously published in~\cite{kim2012}.}. We use neural networks for frame classifiers, and tandem HMM models for a recognizer. And we compare to Gaussian Mixture Model HMM as a baseline. For image descriptors, we use SIFT (scale invariant
feature transform) feature for this experiments. SIFT features is a commonly used type
of image appearance features, and we extract SIFT features from each image, based on local
histograms of oriented image gradients \cite{lowe}.

First we train single layer neural networks (NN), which predict letter or phonological features. The inputs to the NNs are the SIFT features concatenated over a window of several frames around each frame.  The NNs are implemented with Quicknet~\cite{quicknet}. We consider several choices for the NN output functions:  posteriors (softmax), log posteriors, and linear outputs.  We obtain NN training labels for each frame either from the manually labeled apogees or from forced alignments produced by the baseline HMM-based recognizer.   To derive a letter label for each frame given only the apogees, we assume that there is a letter boundary in the middle of each segment between consecutive apogees.  NNs have one hidden layer with 1000 hidden nodes.

The task is recognition of one fingerspelled word at a time, delimited by the signer's button presses.  We use a speaker-dependent 10-fold setup:  We divide each speaker's data ($\sim 600$ words, $\sim 3000$ letters) randomly into ten subsets.  In each fold, eight of the subsets (80\% of the data) are used for training both NNs and HMMs, one (10\%) for tuning NN and HMM parameters, and one (10\%) as a final test set.  We implement the HMMs with HTK~\cite{htk} and language models with SRILM~\cite{srilm}.  
We train smoothed backoff bigram letter language models using lexicons of various sizes, consisting of the most frequent words in the ARPA CSR-III text, which includes English words and names~\cite{csr-III}.  

The tuning parameters, and their most frequently chosen values, are the SIFT pyramid depth (1+2), PCA dimensionality (80), window size (13), NN output function (linear), whether or not image features are appended to NN outputs (yes), number of states per letter HMM (3), number of silence states (9), number of Gaussians per state (8), and language model lexicon size (5k-word).  We tune independently in each fold, i.e., we perform ten independent experiments,
and report the average test performance over the folds.  Recognition performance is measured via the letter error rate, the Levenshtein distance between the hypothesized letter sequence and the reference sequence as a percentage of the reference sequence length. \\

\begin{figure}[h!]
\begin{center}
	\includegraphics[width=1\linewidth]{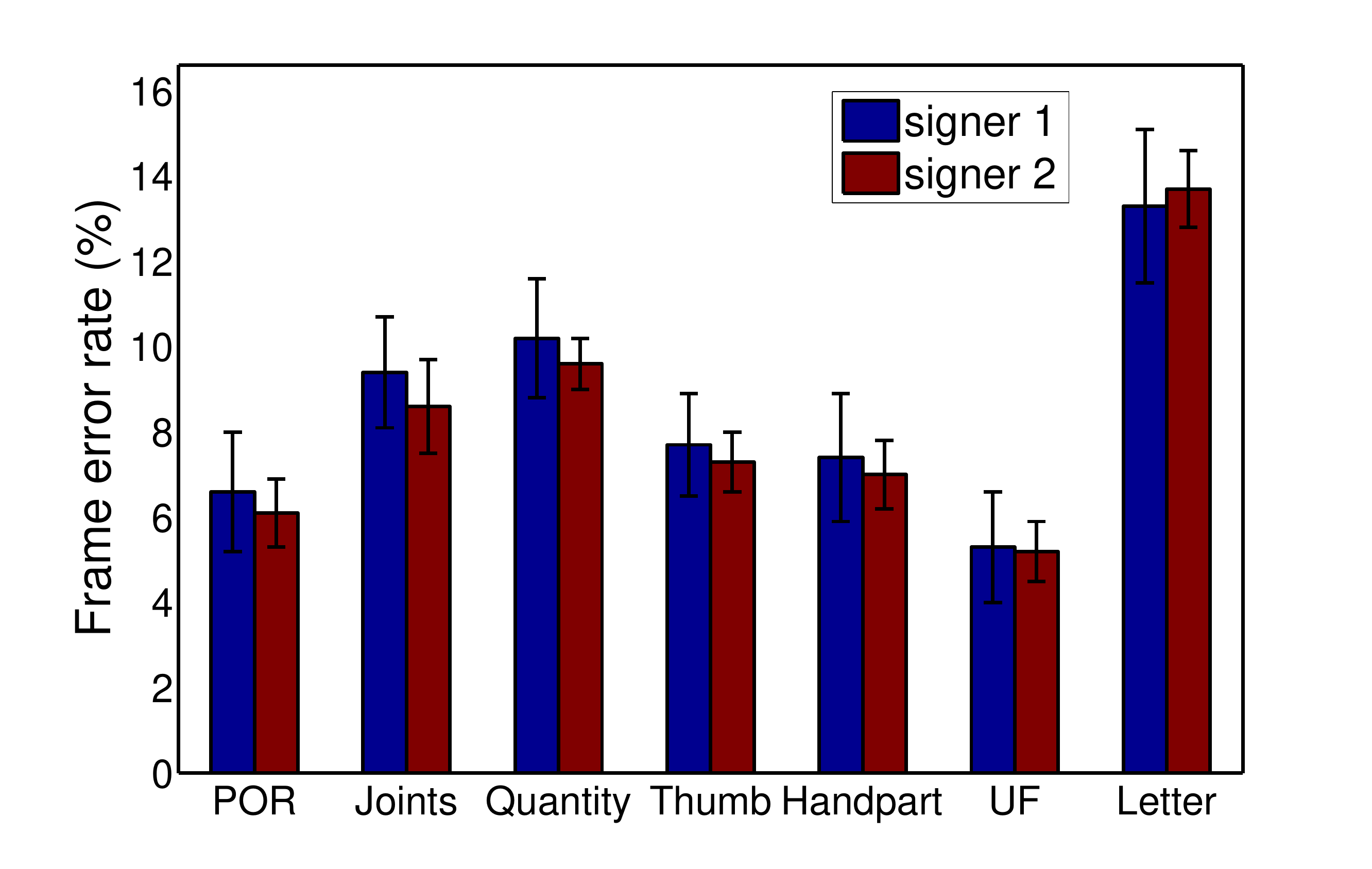}
\end{center}
\caption[Frame error rates of letter and feature NNs]{Frame error rates of letter and feature NNs, averaged over ten folds, and their standard deviations.  
Chance performance (100 - frequency of the most likely class) is $\sim$ 25\% to $\sim$ 55\%, depending on the classifier.}
\label{f:MLP}
\end{figure}

\noindent{\bf Results} Figure~\ref{f:MLP} gives the frame error rates of the NNs, measured with respect to frame labels produced from the manual apogee labels as described above.  All of the classifiers perform much better than chance.

\begin{figure}[h!]
	\hspace{-2.5em}\includegraphics[width=1.2\linewidth]{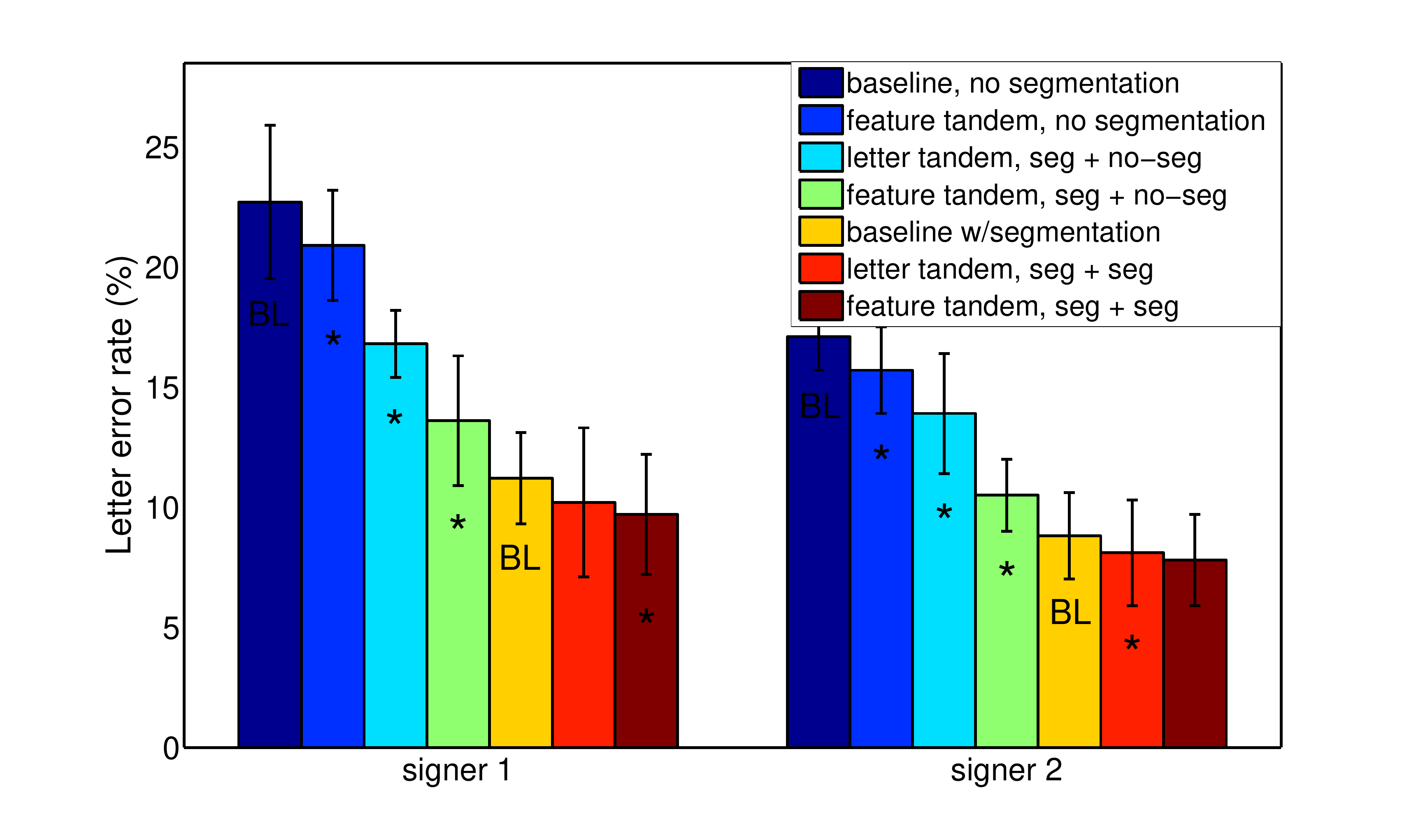}
\caption[Letter error rates and standard deviations on two signers]{Letter error rates and standard deviations on two signers.  For tandem systems, ``seg + no-seg'' indicates that segmentation based on manual apogee labels was used for NN training but not for HMM training; ``seg + seg'' = the segmentation was used for both NN and HMM training; ``no segmentation'' = forced alignment using the HMM baseline was used to generate NN training labels.  An asterisk (`*') indicates statistically significant improvement over the corresponding baseline (`BL') using the same training labels for the HMMs, according to a MAPSSWE test~\cite{pal} at $p < 0.05$.}
\label{f:main}
\end{figure}

Figure~\ref{f:main} shows letter sequence recognition results:  a comparison of the two tandem models and the baseline HMM-based system, using different types of training labels.  For the baseline system, there are two choices for training:  either (1) the manual apogee labels are used to generate a segmentation into letters, and each letter HMM is trained only on the corresponding segments; or (2) the manual labels are ignored and the HMMs are trained without any segmentation, using Expectation-Maximization on sequences corresponding to entire words.  For the tandem systems, we consider three choices:  (1) the manual labels are used to generate a segmentation, and both the NNs and HMMs are trained using this labeled segmentation; (2) the segmentation is used for NN training, but HMMs are trained without it; or (3) the segmentation is not used at all, and the labels for NN training are derived via forced alignment using the baseline (segmentation-free) HMM.  The reason for this comparison is to determine to what extent the (rather time-intensive) manual labeling is helpful.

The tandem systems consistently improve over the corresponding baselines, with the phonological feature-based tandem models outperforming the letter-based tandem models.  Using the manual labels in training makes a large difference to all of the recognizers' error rates, with the forced alignment-based labels producing poorer performance (iterating the forced alignment procedure does not help).  However, in all cases the tandem-based models improve over the baselines and the feature-based models improve over the letter-based tandem models.  

\section{DNN frame classification performance}
\label{sec:dnn-exp}
\begin{figure}\centering
\centering
\vspace{-.25in}
\hspace{-.0in}\includegraphics[width=0.9\linewidth]{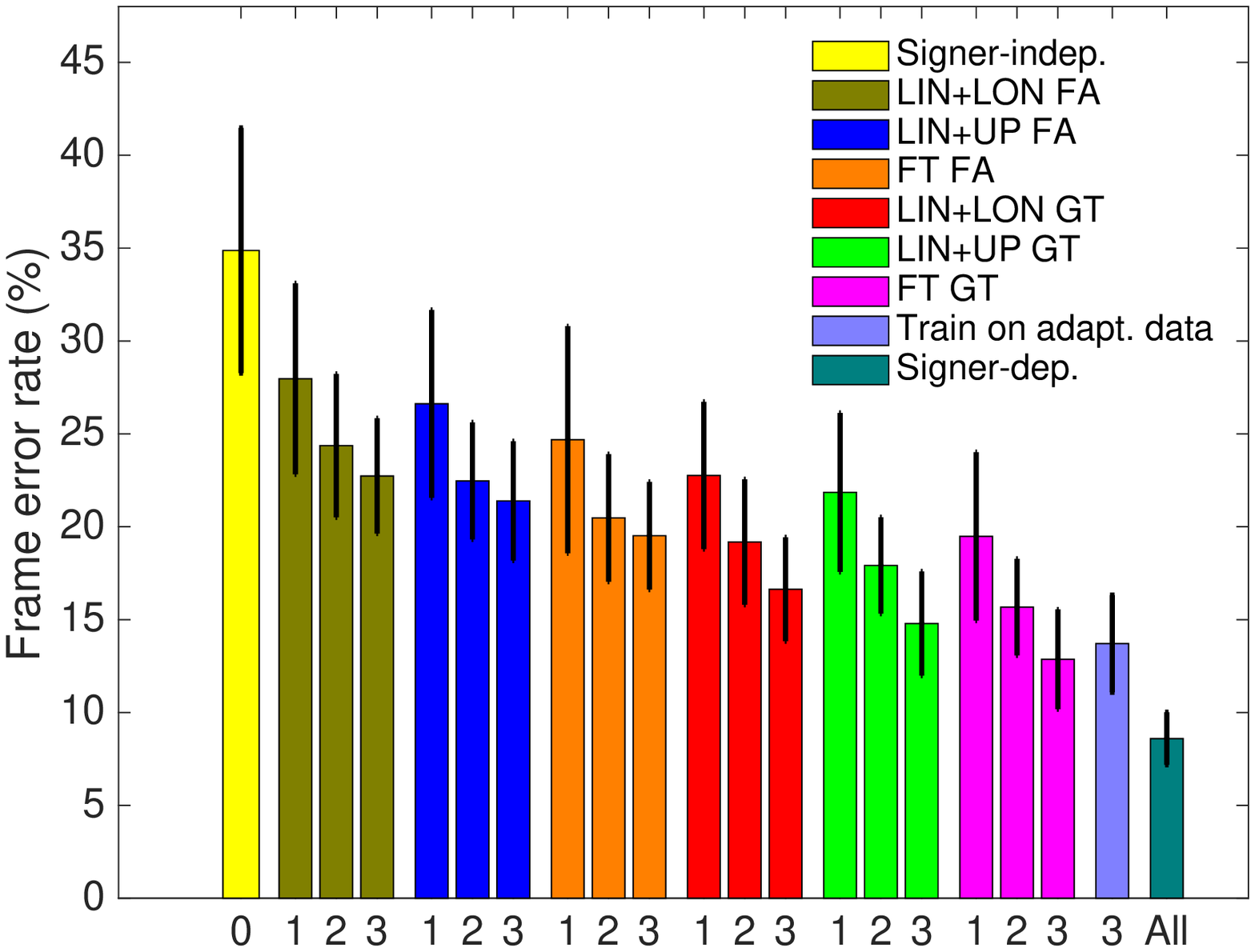} \\
\vspace{-.05in}
\hspace{-.0in}\includegraphics[width=0.9\linewidth]{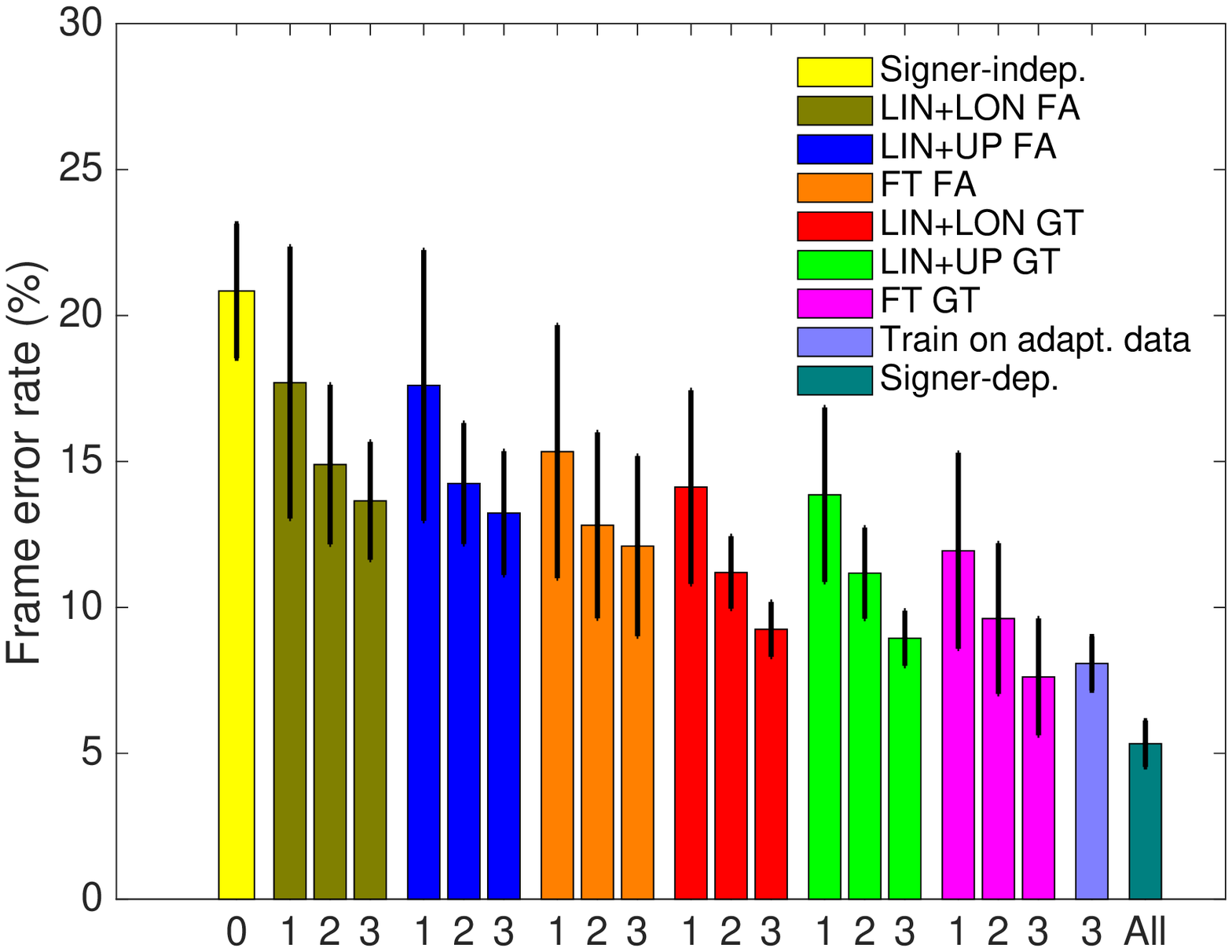}
\vspace{-.15in}
\caption[Frame errors with DNN classifiers, with various settings]{Frame errors with DNN classifiers, with various settings. (Top) letter classifiers (Bottom) Phnological feature classifiers. The horizontal axis labels indicate the amount of adaptation data (0, 1, 2, 3 = none, 5\%, 10\%, 20\% of the test signer's data, corresponding to no adaptation (signer-independent), $\sim29$, $\sim58$, and $\sim115$ words).  GT = ground truth labels; FA = forced alignment labels; FT = fine-tuning. We also added trained DNN on only 20\% of the test signer's data. Signer-dependent DNN uses 80\% of the test signer's data.}
\label{fig:adapt_fer}
\end{figure}

Since all of our fingerspelling recognition models use DNN frame classifiers as a building block,
we first examine the performance of the frame classifiers\footnote{This section includes material previously published in~\cite{kim2016}.}.

The DNNs are trained for seven tasks (letter classification and
classification of each of the six phonological features). For training DNNs, we split training data, which is used for training recognizer, into training data and validation data for DNN, consisting of 90\% and 10\% of the original training data,respectively. The input is
the $128$-dimensional HOG features concatenated over a $21$-frame window.
The DNNs have
 three hidden layers, each with $3000$ ReLUs~\cite{Zeiler_13a}.
 Network learning is done with cross-entropy training with a weight
 decay penalty of $10^{-5}$, via stochastic gradient descent (SGD)
 over $100$-sample minibatches for up to $30$ epochs, with
 dropout~\cite{Srivas_14a} at a rate of $0.5$ at each hidden layer,
 fixed momentum of $0.95$, and initial learning rate of $0.01$, which
 is halved when held-out error rate stops improving.  We pick the best-performing epoch on held-out data.  The network
 structure and hyperparameters were tuned on held-out (signer-independent) data in initial experiments.  

We consider the signer-dependent setting (where the DNN is trained on
data from the test signer), signer-independent setting (where the DNN
is trained on data from all except the test signer), and
signer-adapted setting (where the signer-independent DNNs are adapted
using adaptation data from the test signer).
 For LIN+UP and LIN+LON in adapting DNNs (Section \ref{sec:dnn}), we adapt by running SGD over minibatches of $100$ samples with a fixed momentum of $0.9$ for up to $20$ epochs, with initial learning rate of $0.02$ (which is halved when error rate stops improving on the adaptation data).  For fine-tuning, we use the same SGD procedure as for the signer-independent DNNs.  We pick the epoch with the lowest error rate on the adaptation data.

The frame error rates for all settings are given in
Fig.~\ref{fig:adapt_fer}.  For the signer-adapted case, we consider DNN
adaptation with different types and amounts of supervision.  The types
of supervision include fully labeled adaptation data (``GT'', for
``ground truth'', in the figure), where the peak locations for all
letters are manually annotated; as well as adaptation data labeled
only with the letter sequence but not the timing information.  In the
latter case, we use the baseline tandem system to generate forced
alignments (``FA'' in the figure).  We consider amounts of adaptation
data from 5\% to 20\% of the test signer's full data.

These results show that among the adaptation methods, LIN+UP slightly outperforms LIN+LON, and fine-tuning outperforms both
LIN+UP and LIN+LON.  Without adaptation, we also consider speed normalization and it provides consistent but very small improvements. For letter sequence recognition experiments in
the next section, we adapt via fine-tuning using $20\%$ of the test signer's data.

\begin{figure*}[t]
\centering
\begin{tabular}{@{}c@{\hspace{0.02\linewidth}}c@{\hspace{0.01\linewidth}}c@{\hspace{0.01\linewidth}}c@{}}
No Adapt. & \caja{c}{c}{LIN+UP (Forced-Align.)} & \caja{c}{c}{LIN+UP (Ground Truth)} \\
\includegraphics[width=0.34\linewidth]{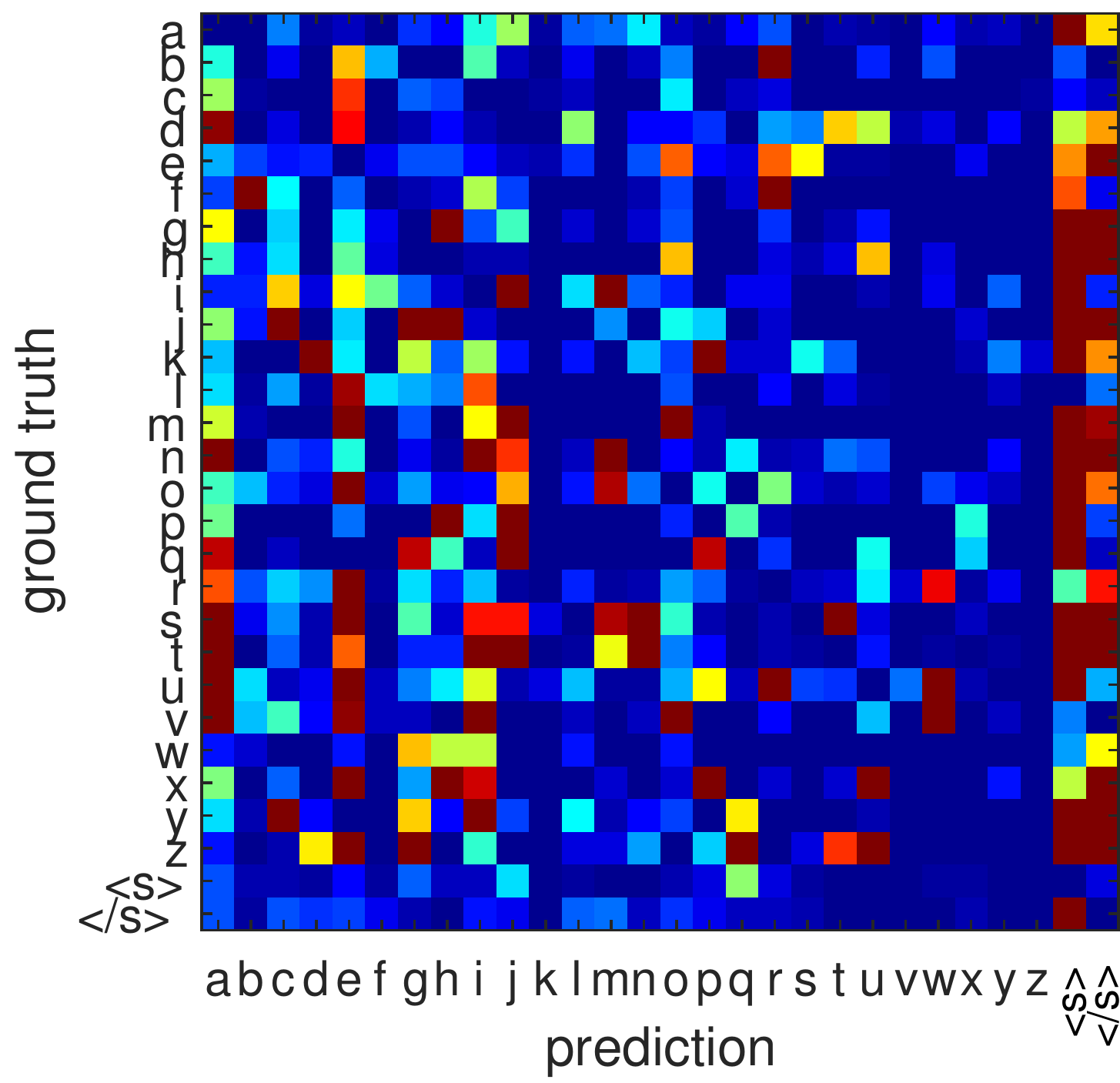} & 
\includegraphics[width=0.28\linewidth]{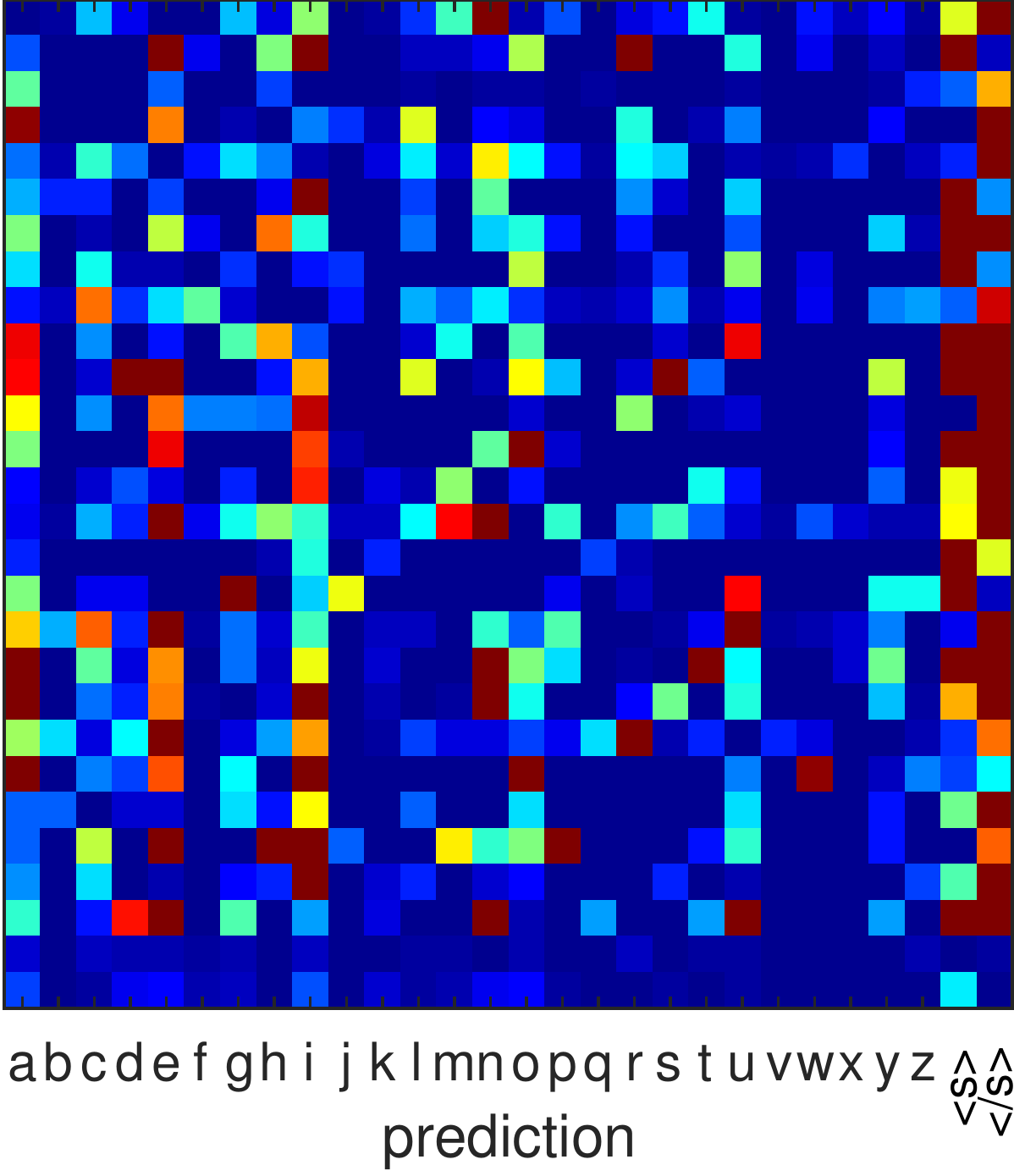} & 
\includegraphics[width=0.34\linewidth]{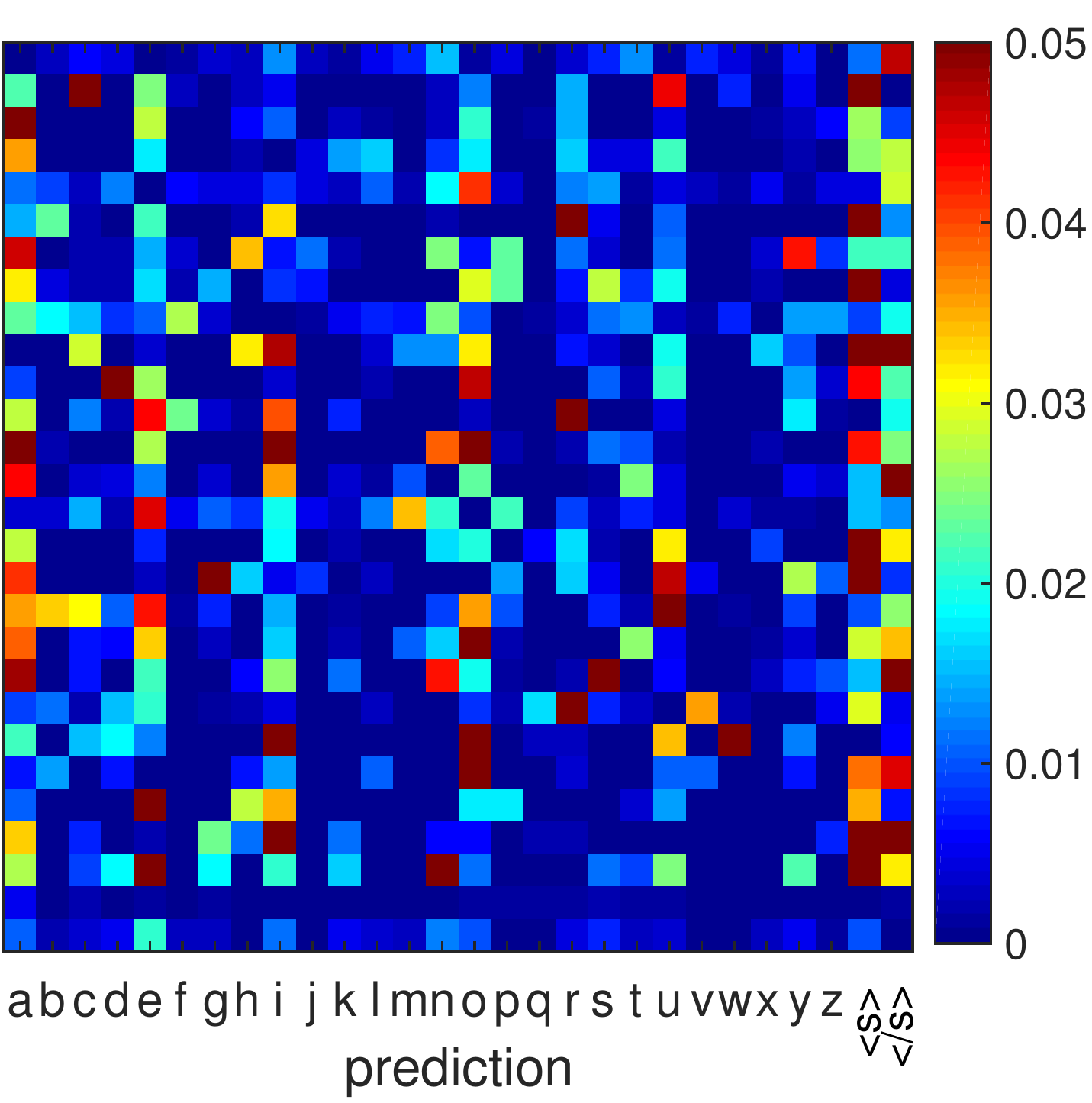} \\
\end{tabular}
\caption[Confusion matrices of DNN classifiers for one test signer (Signer 1)]{Confusion matrices of DNN classifiers for one test signer (Signer 1).  $20\%$ of the test signer's data (115 words) was used for adaptation, and a disjoint $70\%$ was used to compute confusion matrices.  
Each matrix cell is the empirical probability of the predicted class (column) given the ground-truth class (row).  The diagonal has been zeroed out for clarity.}
\label{fig:confmat}
\end{figure*}

In Fig.~\ref{fig:confmat}, we further analyze the DNNs via confusion matrices. One of the main effects is the large number of incorrect predictions of the non-signing classes ($<$s$>$, $<$/s$>$).  We observe the same effect with the phonological feature classifiers.  This may be due to the previously mentioned fact that non-linguistic gestures are variable and easy to confuse with signing when given a new signer's image frames.  The confusion matrices show that, as the DNNs are adapted, this is the main type of error that is corrected.

\section{Letter recognition experiments}
\label{sec:LER-exp}

\subsection{Signer-dependent recognition}
Our first continuous letter recognition experiments are signer-dependent; that is, we train and test on the same signer, for each of four signers. 
For each signer, we use a 10-fold setup:  In each fold, 80\% of the data is used as a training set, 10\% as a development set for tuning parameters, and the remaining 10\% as a final test set.    
We independently tune the parameters in each fold, but to make it comparable to adaptation experiments later, we use 8 out of 10 folds to compute the final test results and report the average letter error rate (LER) over those 8 folds.
For language models, we train letter bigram language models from large online dictionaries of varying sizes that include both English words and names\karenedit{~\cite{csr}}. 
We use HTK~\cite{htk} to implement the tandem HMM-based recognizers and SRILM~\cite{srilm} to train the language models. 
The HMM parameters (number of Gaussians per state, size of language model vocabulary, transition penalty and language model weight), as well as the dimensionality of the HOG descriptor input and HOG depth, were tuned to minimize development set letter error rates for the tandem HMM system.  The front-end and language model hyperparameters were kept fixed for the SCRFs (in this sense the SCRFs are slightly disadvantaged).  
Additional parameters tuned for the SCRF rescoring models included the N-best list sizes, type of feature functions, choice of language models, and L1 and L2 regularization parameters.  Finally, for the first-pass SCRF, we tuned step size, maximum length of segmentations, and number of training epochs. 

Table~\ref{t:LER} (last row) shows the signer-dependent letter recognition results. SCRF rescoring improves over the tandem HMM, and the first-pass SCRF outperforms the others.  

Note that in our experimental setup, there is some overlap of word
types between training and test data.  This is a realistic setup,
since in real applications some of the test words will have been
previously seen and some will be new.  However, for comparison, we
have also conducted the same experiments while keeping the training,
development, and test vocabularies disjoint; in this modified setup,
letter error rates increase by about 2-3\% overall, but the SCRFs
still outperform the tandem HMM model.

\begin{table*}[ht!]
\centering
\resizebox{\linewidth}{!}{
\begin{tabular}{|l||c|c|c|c|c||c|c|c|c|c||c|c|c|c|c|}
\hline
        & \multicolumn{5}{c||}{Tandem HMM} & \multicolumn{5}{c||}{Rescoring SCRF} & \multicolumn{5}{c|}{1st-pass SCRF} \\ \hline
Signer  & 1 & 2 &  3 &  4 & {\bf Mean} &  1 &  2 &  3 &  4 & {\bf Mean} & 1 &  2 &  3 &  4 & {\bf Mean} \\ \hline\hline
Signer-independent  & 54.1&	54.7&	62.6&	57.5&	{\bf 57.2}&	52.6&	51.2&	61.1&	56.3&	{\bf 55.3}&	55.3&	53.3&	72.5&	61.4&	{\bf 60.6} \\ \hline
Forced align. & 30.2&    38.5&   39.6&   36.1&   {\bf 33.6}&     39.5&   36.0&   38.2 &   34.5&   {\bf 32.0}&    24.4&   24.9&   36.5&   35.5&   {\bf 30.3} \\ \hline
Ground truth  & 22.0&    13.0&   31.6&   21.4&   {\bf 22.0}&     22.4&   13.5&   29.5&   21.4&   {\bf 21.7}&     15.2&   10.6&   24.9&   18.4&   {\bf 17.3} \\ \hline
Signer-dependent &     13.8&    7.1&   26.1&   11.5 & {\bf 14.6}& 10.2 &   7.0&   19.1&    10.0 &{\bf 11.5} & 8.1&    7.7&    9.3&   10.1&  {\bf 8.8} \\ \hline 
\end{tabular}
}
\caption{Letter error rate (\%) on four test signers.}
\label{t:LER}
\end{table*}



\subsection{Signer-independent recognition}
\label{sec:sigindep_exp}

In the signer-independent setting, we would like to recognize
fingerspelled letter sequences from a new signer, given a model
trained only on data from other signers.  For each of the four test
signers, we train models on the remaining three signers, and report
the performance for each test signer and averaged over the four test
signers.  For direct comparison with the signer-dependent experiments,
each test signer's performance is itself an average over the 8 test
folds for that signer.

As shown in the first line of Table~\ref{t:LER}, the signer-independent
performance of the three types of recognizers is quite poor, with the
rescoring SCRF somewhat outperforming the tandem HMM and first-pass
SCRF.  The poor performance is perhaps to be expected with such a
small number of training signers.

\subsection{Signer-adapted recognition}
\label{sec:sigadap_exp}

The remainder of Table~\ref{t:LER} (second and third rows) gives the connected letter
recognition performance obtained with the three types of models
using DNNs adapted via fine-tuning, using different types of
adaptation data (ground-truth, GT, vs.~forced-aligned, FA).  For all
models, we do not retrain the models with the adapted DNNs, but tune
hyperparameters of the recognizer on 10\% of the test signer's data.  The tuned models
are evaluated on an unseen 10\% of the test signer's remaining data;
finally, we repeat this for eight choices of tuning and test sets,
covering the 80\% of the test signer's data that we do not use for
adaptation, and report the mean letter error rate over the test sets.

As shown in Table~\ref{t:LER}, adaptation allowed the performance jump
to up to 30.3\% letter error rate with forced-alignment adaptation
labels and up to 17.3\% error rate with ground-truth adaptation labels.
All of the adapted models improve similarly, but interestingly, the
first-pass SCRF is slightly worse than the others before adaptation
and better (by 4.4\% absolute) after ground-truth adaptation.  One
hypothesis is that the first-pass SCRF is more dependent on the DNN
performance, while the tandem model uses the original image features
and the rescoring SCRF uses the tandem model hypotheses and scores.
Once the DNNs are adapted, however, the first-pass SCRF outperforms
the other models. Figure \ref{fig:recog1},\ref{f:recog2},\ref{f:recog3},\ref{f:recog4} illustrate the recognition task with several models.

Additionally, since DNN-HMM hybrid models are recently popular for speech recognition task (e.g., \cite{dahl2012context}), we also conduct an initial experiment with them.  We use Kaldi toolkit \cite{Povey_ASRU2011} for implementation. But their performance for our recognition task was quite poor, for both signer-dependent and signer-independent experiments, so they were not pursued in depth for the reported experiments. We conjecture that such models may need more data to perform well, which is not the case for our task.



\section{Extensions and analysis}

We next analyze our results and consider potential extensions for
improving the models.

\subsection{Analysis: Could we do better by training entirely on adaptation data?}
\label{sec:scratch}

In this section we consider alternatives to the adaptation setting we have chosen --
adapting the DNNs while using sequence models (HMMs/SCRFs) trained
only on signer-independent data.  We fix the model to a first-pass
SCRF and the adaptation data to 20\% of the test signer's data
annotated with ground-truth peak labels.  In this setting, we consider
two alternative ways of using the adaptation data:  (1) using the adaptation data from the
test signer to train both the DNNs and sequence model
from scratch, ignoring the signer-independent training set; and (2)
training the DNNs from scratch on the adaptation data, but using the
SCRF trained on the training signers. We compare these options with
our best results using the signer-independent SCRF and DNNs fine-tuned
on the adaptation data.  The results are shown in Table
\ref{t:LER_gt}. Frame error rates of DNNs trained from scratch using 20\% of the test signer's data is given in Figure \ref{fig:adapt_fer}.

We find that ignoring the signer-independent training set and training
both DNNs and SCRFs from scratch on the test signer
(option (1) above) works remarkably well, better than the
signer-independent models and even better than adaptation via
forced alignment (see Table~\ref{t:LER}).
However, training the
SCRF on the training signers but DNNs from scratch on the adaptation
data (option (2) above) improves performance further.  However,
neither of these outperforms our previous best approach of
signer-independent SCRFs plus DNNs fine-tuned on the adaptation data. Figure \ref{fig:adapt_fer} shows that DNNs trained from scratch is slightly worse than fine-tuned DNNs, but letter recognition has some gap in Table \ref{t:LER_gt}. It may be due to fact that small improvement of DNNs can effect letter recognition performance more.

\begin{table*}[ht!]
\centering
\resizebox{\linewidth}{!}{
\begin{tabular}{|c|c|c|c|c|c||c|c|c|c|c||c|c|c|c|c|}
\hline
& \multicolumn{5}{|c||}{SCRF + DNNs trained from scratch} & \multicolumn{5}{c||}{Sig.-indep. SCRF, DNNs trained from scratch} & \multicolumn{5}{c|}{Sig.-indep. model w/ fine-tuned DNN} \\ \hline
Signer & 1 & 2 & 3 & 4 & {\bf Mean} & 1 & 2 & 3 & 4 & {\bf Mean} & 1 & 2 & 3 & 4 & {\bf Mean} \\ \hline\hline
Error rate & 23.2 &  18.2 &  28.9 &  30.1 & {\bf 25.1} & 18.4 &  12.6 &  27.0 &  20.7 & {\bf 19.7}  & 15.2&   10.6&   24.9&   18.4&   {\bf 17.3} \\ \hline
\end{tabular}
}
\caption[Letter error rates (\%) for different settings of SCRF and DNN
  training in the signer-adapted case]{Letter error rates (\%) for different settings of SCRF and DNN
  training in the signer-adapted case.  Details are given in Section~\ref{sec:scratch}.}
\label{t:LER_gt}
\end{table*}

\subsection{Analysis:  Letter vs.~feature DNNs}

\begin{figure*}[t]
\centering
\begin{tabular}{c}
\includegraphics[width=0.9\linewidth]{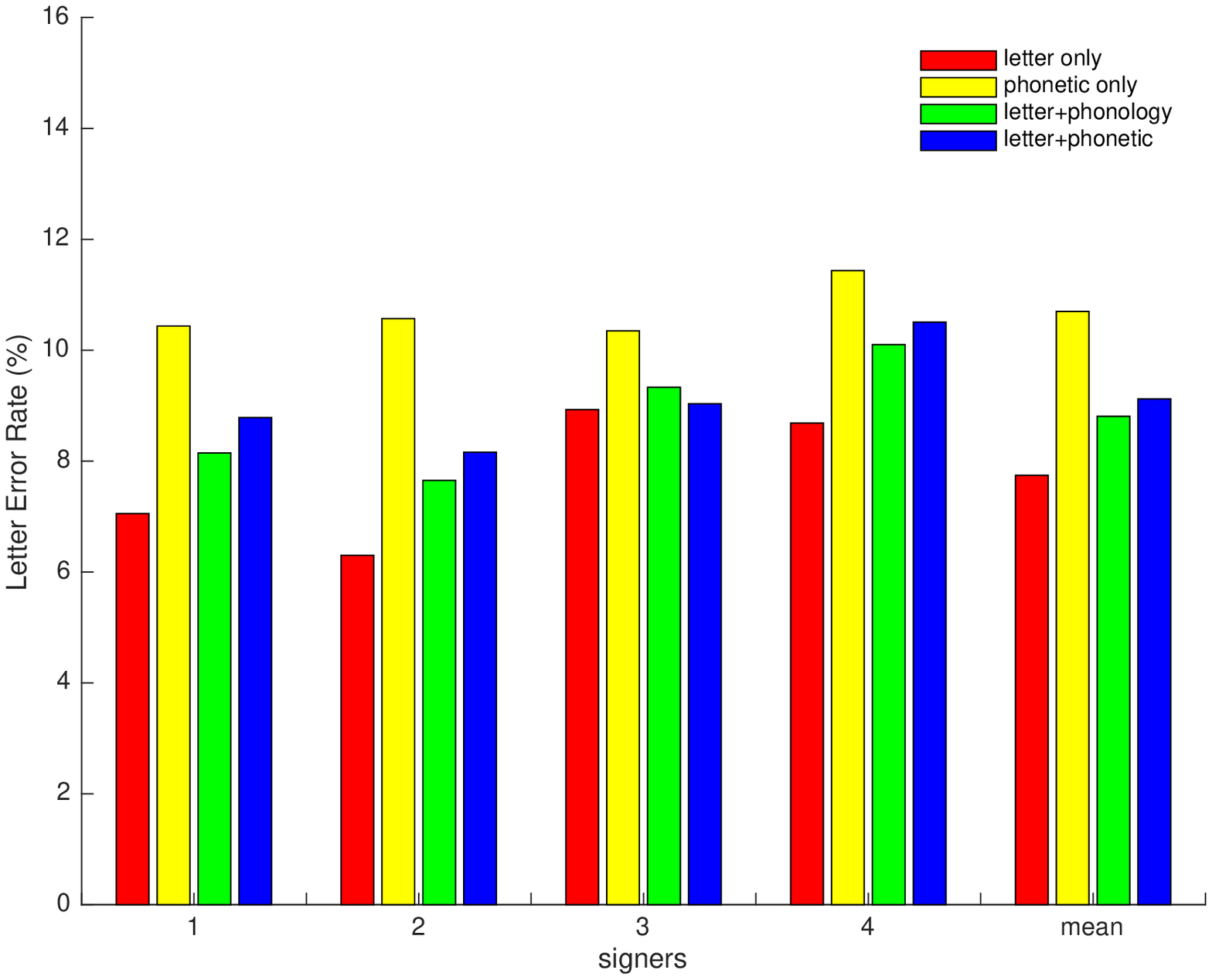} \\
\includegraphics[width=0.9\linewidth]{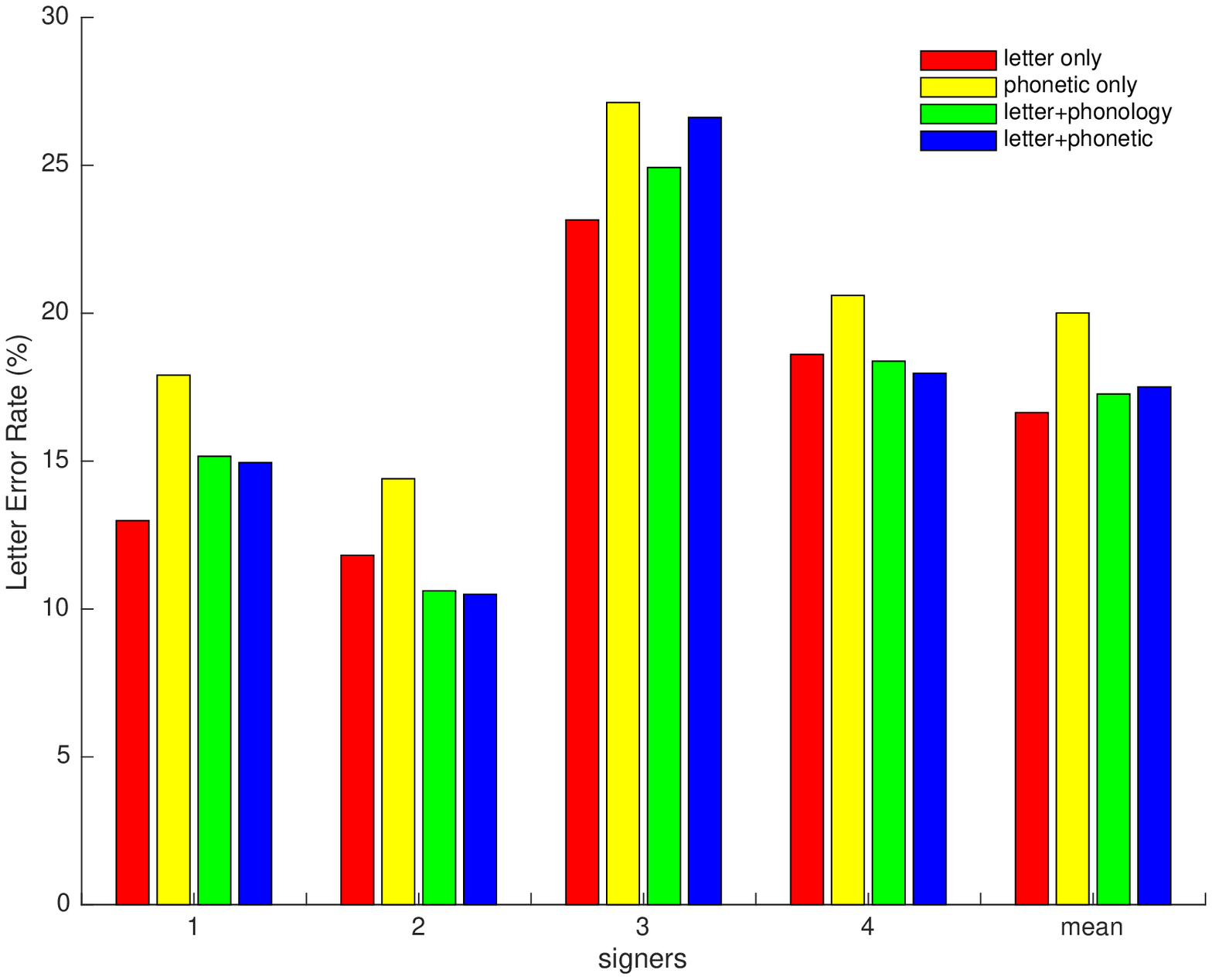} \\
\end{tabular}
\caption[Comparisons with different DNN classifiers]{Comparisons with different DNN classifiers; (Top) Signer-dependent recognition and (Bottom) Signer-independent recognition with frame annotations and adaptation. We compare: letter only, phonetic features only, letter + phonological feature \cite{bren} and letter + phonetic feature \cite{Keane2014diss}. Using only letter outperforms others.} 
\label{fig:ling_fe}
\end{figure*}

We next compare the letter DNN classifiers and the phonological
feature DNN classifiers in the context of the first-pass SCRF
recognizers.  We also consider an alternative sub-letter feature set,
in particular a set of phonetic features introduced by Keane \cite{Keane2014diss}, whose feature values are listed in Table \ref{t:phonetic}. We use the first-pass SCRF
with either only letter classifiers, only phonetic feature
classifiers, letter + phonological feature classifiers, and letter +
phonetic feature classifiers.  We do not consider the case of
phonological features alone, because they are not discriminative for
some letters. Figure \ref{fig:ling_fe} shows the letter recognition results
for the signer-dependent and signer-adapted settings.  

We find that using letter classifiers alone outperforms the other
options in the speaker-dependent setting, achieving 7.6\% letter error rate.  For signer-adapted
recognition, phonological or phonetic features are helpful in addition
to letters for two of the signers (signers 2 and 4) but not for the
other two (signers 1,3); on average, using letter classifiers alone is
best in both cases, achieving 16.6\% error rate on average.  In contrast, in an initial experiment in Section \ref{sec:pre-exp},  we found that phonological features 
  outperform letters in the tandem HMM.  However, those experiments
  used a tandem model with neural networks with
  a single hidden layer; we conjecture that with more layers, we are
  able to do a 
  better job at the more complicated task of letter classification.

\begin{table*}[ht!]
\centering
\resizebox{\linewidth}{!}{
\begin{tabular}{ccccccccccccccc}
\hline
letter&\multicolumn{2}{|c|}{index} & \multicolumn{2}{c|}{middle} & \multicolumn{2}{c|}{ring}& \multicolumn{2}{c|}{pinky}&spread&\multicolumn{4}{|c|}{thumb}&palm \\ \hline

& MCP &PIP& MCP& PIP& MCP& PIP& MCP& PIP& y& z& PIP& touch \\ \hline \hline
a& 90& 90& 90& 90& 90& 90& 90& 90& 0& 0& 90& 180& i& for\\
b& 180& 180& 180& 180& 180& 180& 180& 180& 0& -45& 90& 180& r& for\\ 
c& 180& 90& 180& 90& 180& 90& 180& 90& 0& 0& 0& 135& -& for\\
d& 180& 180& 90& 135& 90& 90& 90& 90& 0& 0& 45& 180& m& for\\
e& 135& 90& 135& 90& 135& 90& 135& 90& 0& -45& 0& 90& r& for\\
f& 90& 135& 180& 180& 180& 180& 180& 180& 1& 0& 45& 180& i& for\\
g& 180& 180& 90& 90& 90& 90& 90& 90& 0& 0& 90& 180& m& in\\
h& 180& 180& 180& 180& 90& 90& 90& 90& 0& -45& 90& 180& r& in\\
i& 90& 90& 90& 90& 90& 90& 180& 180& 0& -45& 90& 180& r& for\\
j& 90& 90& 90& 90& 90& 90& 180& 180& 0& -45& 90& 180& r& dwn\\
k& 180& 180& 90& 180& 90& 90& 90& 90& 0& 0& 90& 180& m& for\\
l& 180& 180& 90& 90& 90& 90& 90& 90& 0& 90& 0& 180& -& for\\
m& 90& 135& 90& 135& 90& 135& 90& 90& 0& -45& 90& 180& p& for\\
n& 90& 135& 90& 135& 90& 90& 90& 90& 0& -45& 90& 180& r& for\\
o& 135& 135& 135& 135& 135& 135& 135& 135& 0& -45& 0& 180& m/i& for\\
p& 180& 180& 90& 180& 90& 90& 90& 90& 0& 0& 90& 180& m& dwn\\
q& 180& 180& 90& 90& 90& 90& 90& 90& 0& 0& 90& 180& m& dwn\\
r& 180& 180& 180& 180& 90& 90& 90& 90& -1& -45& 0& 180& r& for\\
s& 90& 90& 90& 90& 90& 90& 90& 90& 0& -45& 45& 180& r& for\\
t& 90& 135& 90& 90& 90& 90& 90& 90& 0& -45& 90& 180& m& for\\
u& 180& 180& 180& 180& 90& 90& 90& 90& 0& -45& 90& 180& r& for\\
v& 180& 180& 180& 180& 90& 90& 90& 90& 1& -45& 90& 180& r& for\\
w& 180& 180& 180& 180& 180& 180& 90& 90& 1& -45& 90& 180& p& for\\
x& 180& 135& 90& 90& 90& 90& 90& 90& 0& -45& 45& 180& m& for\\
y& 90& 90& 90& 90& 90& 90& 180& 180& 1& 90& 0& 180& -& for\\
z& 180& 180& 90& 90& 90& 90& 90& 90& 0& 0& 45& 180& m& for\\
zz& 180& 180& 180& 180& 90& 90& 90& 90& 1& 0& 45& 180& m& for\\ \hline 
\end{tabular}
}
\vspace{-.1in}
\caption[Phonetic features]{Phonetic features \cite{Keane2014diss}. The numerical values refer to joint angles in each finger.}
\label{t:phonetic}
\end{table*}

\subsection{Analysis:  DNNs vs.~CNNs}
\label{sec:cnn-exp}

\begin{figure}\centering
\centering
\vspace{-.25in}
\hspace{-.0in}\includegraphics[width=0.9\linewidth]{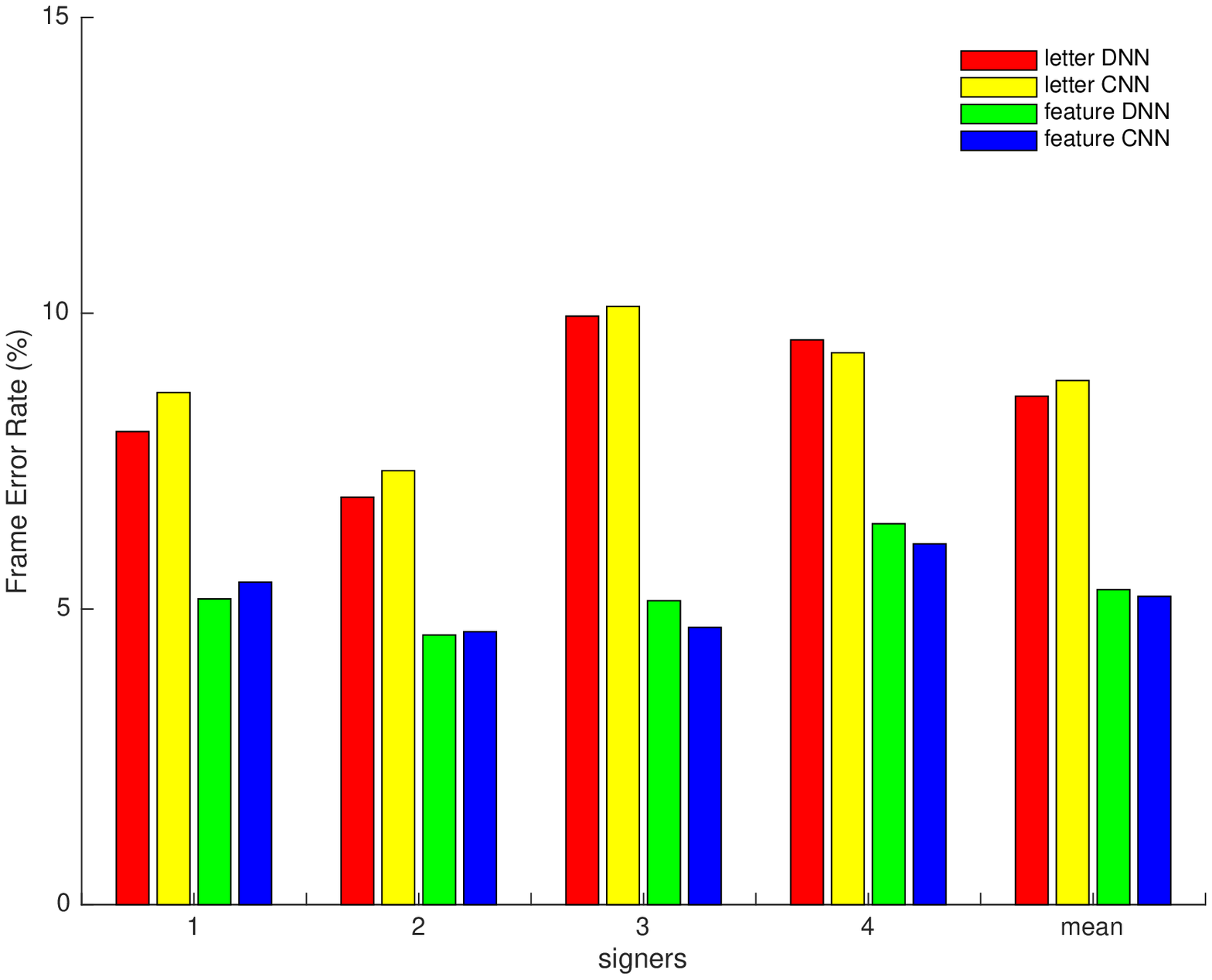} \\
\vspace{-.05in}
\caption[Comparisons between DNN and CNN frame classifiers]{Comparisons between DNN and CNN frame classifiers, for signer-dependent experiments over four signers.}
\label{fig:cnn_fer}
\end{figure}

We also conduct experiments with convolutional neural networks (CNNs) as frame classifiers. In all experiments so far, we have used HoG features as our image descriptor. However, using CNNs with raw image pixels without any hand-crafted image descriptor has recently become popular and has shown improved performance for image recognition tasks (e.g., \cite{krizhevsky2012imagenet, simonyan2014very}). Thus, to see the effectiveness of CNNs for our frame classifiers, we compare the feature classification performance of letter and phonological features between DNNs and CNNs. We use a signder-dependent setting and the same 8-fold setup. Inputs to CNNs are grayscale $64 \times 64 \times T$ pixels. $T$ is the number of input frames used in the input window, as in our DNNs. For the structure of CNNs, we use 32 kernels of $3 \times 3$ filters with stride 1 for the first and second convolutional layer. Then we add a max pooling layer with $2 \times 2$ pixel window, with stride 2. For the third and fourth convolutional layer, we use 64 kernels of $3 \times 3$ filters with stride 1. Again, we add a max pooling layer with $2 \times 2$ pixel window, with stride 2. For all convolutional layers, ReLUs are used for the nonlinearity. Finally, two fully-connected layers with 2000 ReLUs are added, followed by a softmax output layer. We use dropout to prevent overfitting, 0.25 probability for all convolutional layers and 0.5 for fully-connected layers. Training is done by stochastic gradient descent with learning rate 0.01, learning rate decay 1e-6, and momentum 0.9. Mini-batch size is 100. $T$ was tuned and set to 21. For CNN implementation, we use Keras \cite{chollet2015} with Theano \cite{2016arXiv160502688short}. Figure \ref{fig:cnn_fer} shows the results for signer-dependent setting. We find that the performances of the DNN and CNN are comparable. Specifically, for letter classification, DNNs are slightly better, and for phonological feature classification, CNNs are slightly better, but the gaps are very small. Our best letter recognizer only depends on the outputs of feature classifiers, so we may assume that the performance of letter recognition using CNNs would be comparable to that using DNNs.

\subsection{Improving performance in the force-aligned adaptation case}

We next attempt to improve the performance of adaptation in the
absence of ground-truth (manually annotated) peak labels.  Using only
the letter label sequence for the adaptation data, we use the
signer-independent tandem recognizer to get force-aligned frame
labels.  We then adapt (fine-tune) the DNNs using the force-aligned
adaptation data (as before in the FA case).  We then
re-align the test signer's adaptation data with the adapted
recognizer. Finally, we adapt the DNNs again with the re-aligned data.
Throughout this experiment, we do not change the recognizer but only
update the DNNs.  We use first-pass SCRF with letter classifiers for these experiments. Using this iterative realignment approach, we are able to further improve
the recognition error rate in the FA case by about 1.3\%, as shown in Table \ref{t:LER_fa}.

\begin{table*}[ht!]
\centering
\resizebox{\linewidth}{!}{
\begin{tabular}{|c|c|c|c|c||c|c|c|c|c|}
\hline
\multicolumn{5}{|c||}{Fine-tuning with FA} &
\multicolumn{5}{c|}{Fine-tuning with FA + realignment} \\ \hline
Signer 1 & Signer 2 & Signer 3 & Signer 4 & {\bf Mean} & Signer 1 & Signer 2 & Signer 3 & Signer 4 & {\bf Mean} \\ \hline\hline
22.7 & 26.0 & 33.4&  34.8&   {\bf 29.2}&  21.9& 25.3&  30.5&  34.0& {\bf 27.9} \\ \hline
\end{tabular}
}
\caption{Letter error rates (\%) with iterated forced-alignment (FA) adaptation.}
\label{t:LER_fa}
\end{table*}

\subsection{Improving performance with segmental cascades}

Finally, we consider whether we can improve upon the performance of
our best models, the first-pass SCRFs, by rescoring their results in a second pass with
more powerful features.  We follow the discriminative segmental
cascades (DSC) approach of \cite{tang2015}, where a simpler first-pass SCRF is used for lattice
generation and a second SCRF, with more computationally demanding
features, is used for rescoring.  

For these experiments we start with the most successful first-pass
SCRF in the above experiments, which uses letter DNNs and is adapted
with 20\% of the test signer's data with ground-truth peak labels.  For the second-pass SCRF, we use the
first-pass score as a feature, and add to it two more complex
features:  a segmental DNN, which takes as input an entire
hypothesized segment and produces posterior probabilities for all of
the letter classes; and the ``peak detection'' feature described in
Section~\ref{sec:method}. For training segmental DNNs, we use ground truth segmentation from peak annotations as training data. Because they have variable lengths,  we use the means of each a third of the segment as $\textrm{div$_s$}$ in Section~\ref{sec:re_scrf}. We use the same structure and learning strategy as the DNN frame classifiers. We use the same bigram language model as Tandem HMM and rescoring SCRF models.

As shown in Table~\ref{t:DSC}, for signer-independent setting, it 
slightly improves the average letter error rate over four signers, from
16.6\% in the first pass to 16.2\% in the second pass.  This
improvement, while small, is statistically significant at the p=0.05
level. For the statistical significance test, we use MAPSSWE test~\cite{pal}. These results combine the most successful ideas from our
experiments and form our final best result for signer-independent experiment. But for signer-dependent setting, this approach achieves the comparable performance and does not improve the recognition in the second pass.

\begin{table*}[ht!]
\centering
\resizebox{\linewidth}{!}{
\begin{tabular}{|c|c|c|c|c|c||c|c|c|c|c|}
\hline
&\multicolumn{5}{|c||}{Signer-dependent} &
\multicolumn{5}{c|}{Signer-adaptation} \\ \hline
&Signer 1 & Signer 2 & Signer 3 & Signer 4 & {\bf Mean} & Signer 1 & Signer 2 & Signer 3 & Signer 4 & {\bf Mean} \\ \hline\hline
1st-pass& 7.1&   6.7&    7.7&    8.7&   {\bf 7.6}&    13.0  & 11.8 &  23.2 &  18.6& {\bf 16.6} \\ \hline
2nd-pass& 7.2& 6.5& 8.1&8.6&    {\bf 7.6}&  13.0& 11.2&  21.7&  18.8& {\bf 16.2} \\ \hline
\end{tabular}
}
\caption{Letter error rates (\%) with second pass cascade.}
\label{t:DSC}
\end{table*}


\subsection{Analysis: decomposition of errors}

\begin{table*}[ht!]
\centering
\resizebox{\linewidth}{!}{
\begin{tabular}{|l||c|c|c||c|c|c||c|c|c|}
\hline
        & \multicolumn{3}{c||}{Tandem HMM} & \multicolumn{3}{c||}{Rescoring SCRF} & \multicolumn{3}{c|}{1st-pass SCRF} \\ \hline
Signer  & D & S &  I & D & S &  I  & D & S &  I \\ \hline\hline
S1&13.5&	5.7&	2.8&	13.5&	7.2&	1.8&	8.3&	3.5&	1.8	\\ \hline
S2&6.6&	2.8&	3.7&	6.2&	3.7&	3.5&	4.7&	2.5&	4.3	\\ \hline
S3&18&	11.3&	2.4&	19&	9.2&	1.4&	12.4&	7.7&	1.7	\\ \hline
S4&10.6&	7.3&	3.5&	9.6&	7.8&	3.9&	8.8&	7.1&	2.8	\\ \hline
{\bf Mean}&{\bf 12.2}&{\bf 	6.8}&{\bf 	3.1}&{\bf 	12.1}&{\bf 	7}&{\bf 	2.7}&{\bf 	8.6}&{\bf 	5.2}&{\bf 	2.6}	\\ \hline
\end{tabular}
}
\caption[Decomposition of letter error rate with adaptation]{Decomposition of letter error rate with adaptation (\%); relative numbers of D=Deletion,S=substitution,I=Insertion of each signer and mean of them, with respect to each recognizer.}
\label{t:decomp_adapt}
\end{table*}

We also decompose the letter error rate for further analysis. Here we consider signer-adaptation experiments. We compute the number of substitution errors (S), deletion errors (D) and insertion errors (I) to match hypothesis produced by models to ground truth letter sequence.
 
\begin{figure*}[!th]
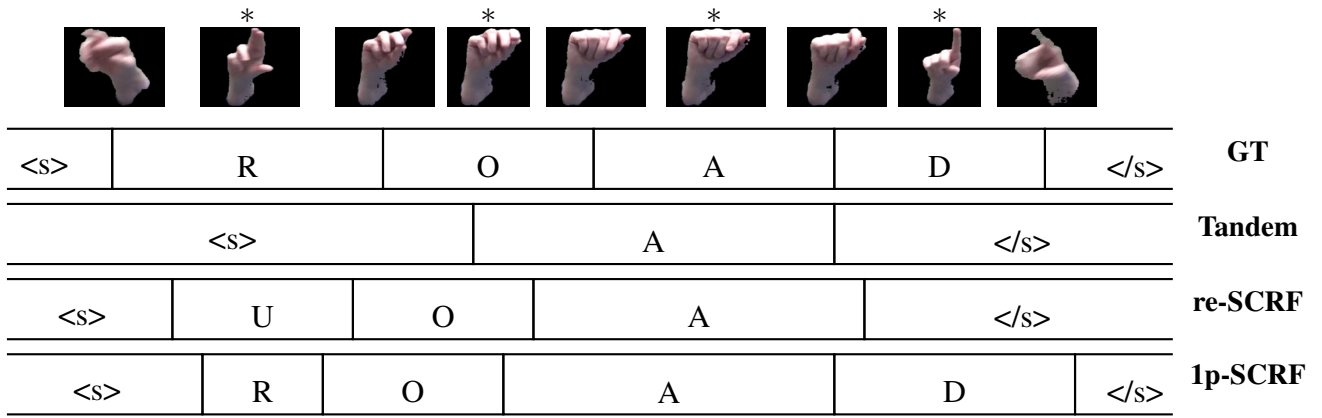

\begin{tikzpicture}
  \foreach \sta/\sto/\letter/\imnum in { 6.1/7.0/R/2, 7.0/7.7/O/4, 7.7/8.5/A/6, 8.5/9.2/D/8} {
    \draw [black,thick] ($(4*\sta,0)$) rectangle ($(4*\sto,.8)$);
    \node at ($(4*\sta,0.3)!.5!(4*\sto,.3)$) {\letter};
    \node at ($(4*\sta,1.6)!.5!(4*\sto,1.6)$)
    {\includegraphics[width=1.7cm]{ex/ex_rita\imnum}};
    \node at ($(4*\sta,2.3)!.5!(4*\sto,2.3)$) {$\ast$};
  }
  \foreach \pos/\imnum in {6.1/1, 7.0/3, 7.7/5, 8.5/7, 9.2/9}
  \node at ($(4*\pos,1.6)$) {\includegraphics[width=1.7cm]{ex/ex_rita\imnum}};
  \draw [black,thick] (23,0)--(4*6.1,0)--(4*6.1,.8)--(23,.8);
  \node at (23.5,.3) {<s>};
  \draw [black,thick] (38.5,0)--(9.2*4,0)--(9.2*4,.8)--(38.5,.8);
  \node at (38,.3) {</s>};
  \node at (39.5,.5) {\small \bf GT};
  
  \foreach \sta/\sto/\letter in { 7.3/8.5/A} {
    \draw [black,thick] ($(4*\sta,-1)$) rectangle ($(4*\sto,-.2)$);
    \node at ($(4*\sta,-.7)!.5!(4*\sto,-.7)$) {\letter};
  }
    \draw [black,thick] (23,-1)--(4*7.3,-1)--(4*7.3,-.2)--(23,-.2);
    \node at (26,-.7) {<s>};
    \draw [black,thick] (38.5,-1)--(4*8.5,-1)--(4*8.5,-.2)--(38.5,-.2);
    \node at (36.5,-.7) {</s>};
    \node at (39.5,-.5) {\small \bf Tandem};

    \foreach \sta/\sto/\letter in { 6.3/6.9/U, 6.9/7.5/O, 7.5/8.6/A} {
    \draw [black,thick] ($(4*\sta,-2)$) rectangle ($(4*\sto,-1.2)$);
    \node at ($(4*\sta,-1.7)!.5!(4*\sto,-1.7)$) {\letter};
  }
    \draw [black,thick] (23,-2)--(4*6.3,-2)--(4*6.3,-1.2)--(23,-1.2);
    \node at (24,-1.7) {<s>};
    \draw [black,thick] (38.5,-2)--(4*8.6,-2)--(4*8.6,-1.2)--(38.5,-1.2);
    \node at (36.5,-1.7) {</s>};
    \node at (39.5,-1.5) {\small \bf re-SCRF};
    
    \foreach \sta/\sto/\letter in { 6.4/6.8/R, 6.8/7.4/O, 7.4/8.5/A, 8.5/9.3/D} {
    \draw [black,thick] ($(4*\sta,-3)$) rectangle ($(4*\sto,-2.2)$);
    \node at ($(4*\sta,-2.7)!.5!(4*\sto,-2.7)$) {\letter};
  }
    \draw [black,thick] (23,-3)--(4*6.4,-3)--(4*6.4,-2.2)--(23,-2.2);
    \node at (24.2,-2.7) {<s>};
    \draw [black,thick] (38.5,-3)--(4*9.3,-3)--(4*9.3,-2.2)--(38.5,-2.2);
    \node at (38,-2.7) {</s>};
    \node at (39.5,-2.5) {\small \bf 1p-SCRF};  

\end{tikzpicture}
\caption[Illustration of improved deletion and substitution]{From top to bottom: frames from the word ROAD, with asterisks denoting the peak frame
  for each letter and ``$<$s$>$'' and ``$<$/s$>$'' denoting periods before the first letter
  and after the last letter; ground-truth segmentation based on peak
annotations (GT); segmentation produced by the tandem HMM (Tandem); segmentation produced by the rescoring SCRF (re-SCRF); and segmentation produced by the first-pass SCRF (1p-SCRF). From Signer 3.}
\label{f:recog3}
\end{figure*}

The results are reported in Table \ref{t:decomp_adapt}, which shows that first-pass SCRF model mostly improves deletion and substitution errors over other models.  Figure \ref{f:recog3} show a such example; tandem HMM have many deletions and rescoring SCRF has a substitution. Those are corrected with first-pass SCRF model. More details with additional analysis for signer-dependent and signer-independent settings are in Appendix \ref{sec:decomp}.

\chapter{Conclusion}
This thesis tackles the problem of unconstrained fingerspelled letter
sequence recognition in ASL, where the letter sequences are not
restricted to any closed vocabulary.  This problem is challenging due
to both the small amount of available training data and the
significant variation between signers.  
Our recognition
experiments have compared HMM-based and segmental models with features
based on DNN
classifiers, and have investigated a range of settings including
signer-dependent, signer-independent, and signer-adapted.  Our main findings are:

\begin{itemize}

  \item Signer-independent fingerspelling recognition, where the test
    signer is unseen in training, is quite
    challenging, at least with the small data sets available to date,
    with letter error rates around 60\%.
    On the other hand, even with a small amount of training data,
    signer-dependent recognition is quite successful, reaching letter
    error rates below 10\%; and adaptation allows us to bridge a
    large part of the gap between signer-independent and
    signer-dependent performance.

  \item Our best results are using two-pass discriminative segmental
    cascades with features based on frame-level DNN letter
    classifiers.  This approach achieves an average letter error rate
    of 7.6\% in the signer-dependent
    setting and 16.2\% LER in the signer-adapted setting.  The
    adapted models use signer-independent SCRFs and DNNs adapted by
    fine-tuning on the adaptation data.  The adapted 
    results are obtained using ~115 words of adaptation data with
    manual annotations of letter peaks.

  \item If less adaptation data is available, we can still get
    improvements from adaptation down to about 30 words of adaptation
    data.

  \item  In the absence of manual peak annotations for the adaptation data, we can 
    automatically align the adaptation data and still get a
    significant boost in performance over the signer-independent case,
    and we can iteratively improve the performance by re-aligning the
    data with the adapted models.  Our best adapted models using
    automatically aligned adaptation data achieve 27.3\% letter error rate.

  \item The main types of errors that are addressed by adapting the
    DNN classifiers are confusions between signing and non-signing segments.
\end{itemize}

The future directions we are interested in are as follows:

\begin{itemize}
\item We would like to investigate new sequence recognition approaches. Specifically, we have used separate feature classifiers and recognizers for our models. It would be interesting to combine both and train the recognizer ``end-to-end'', as in, e.g., \cite{graves2014towards} for speech recognition. For feature classifier, using recurrent neural networks, such as long-short term memory networks, would also be interesting.
\item We are also interested in new adaptation methods, especially for the case where no manual annotations are available for the adaptation data.
\item We still have relatively small amount of training data and would like to expand our data collection to a larger number of signers and to more realistic data collected ``in the wild'', such as online videos from Deaf social media.  Also, acquiring manual annotations is expensive, and we would like to reduce the dependence on manually aligned data.  As larger data sets are collected, they might alleviate the need for manual alignments, or it may still be necessary to develop better ways of taking advantage of weakly labeled data.
\item In real data, fignerspelling is usually embedded within running ASL. Future work will consider jointly detecting and recognizing fingerspelling sequences.
\end{itemize}

\begin{singlespace}
\bibliography{reference,description,icassp16b,csl2016}
\bibliographystyle{plain}
\end{singlespace}

\appendix
\chapter{Examples of hypotheses produced by models}

\begin{figure*}[!th]
\begin{tikzpicture}
  \foreach \sta/\sto/\letter/\imnum in { 6.0/6.6/V/2, 6.6/7.2/E/4, 7.2/7.8/N/6, 7.8/8.4/I/8, 8.4/9.0/C/10, 9.0/9.4/E/12} {
    \draw [black,thick] ($(4*\sta,0)$) rectangle ($(4*\sto,.8)$);
    \node at ($(4*\sta,0.3)!.5!(4*\sto,.3)$) {\letter};
    \node at ($(4*\sta,1.6)!.5!(4*\sto,1.6)$)
    {\includegraphics[width=1.1cm]{ex/ex_andy\imnum}};
    \node at ($(4*\sta,2.3)!.5!(4*\sto,2.3)$) {$\ast$};
  }
  \foreach \pos/\imnum in {6.0/1, 6.6/3, 7.2/5, 7.8/7, 8.4/9, 9.0/11, 9.4/13}
  \node at ($(4*\pos,1.6)$) {\includegraphics[width=1.1cm]{ex/ex_andy\imnum}};
  \draw [black,thick] (23,0)--(4*6.0,0)--(4*6.0,.8)--(23,.8);
  \node at (23.5,.3) {<s>};
  \draw [black,thick] (38.5,0)--(9.4*4,0)--(9.4*4,.8)--(38.5,.8);
  \node at (38,.4) {</s>};
  \node at (39.5,.5) {\small \bf GT};
  
  \foreach \sta/\sto/\letter in { 6.1/6.8/V, 6.8/7.2/E, 7.2/8.0/N, 8.0/8.6/I, 8.6/9.2/C} {
    \draw [black,thick] ($(4*\sta,-1)$) rectangle ($(4*\sto,-.2)$);
    \node at ($(4*\sta,-.7)!.5!(4*\sto,-.7)$) {\letter};
  }
    \draw [black,thick] (23,-1)--(4*6.1,-1)--(4*6.1,-.2)--(23,-.2);
    \node at (23.5,-.7) {<s>};
    \draw [black,thick] (38.5,-1)--(4*9.2,-1)--(4*9.2,-.2)--(38.5,-.2);
    \node at (38,-.7) {</s>};
    \node at (39.5,-.5) {\small \bf Tandem};

    \foreach \sta/\sto/\letter in { 6.2/6.8/V, 6.8/7.2/E, 7.2/8.1/N, 8.1/8.7/J, 8.7/9.3/E} {
    \draw [black,thick] ($(4*\sta,-2)$) rectangle ($(4*\sto,-1.2)$);
    \node at ($(4*\sta,-1.7)!.5!(4*\sto,-1.7)$) {\letter};
  }
    \draw [black,thick] (23,-2)--(4*6.2,-2)--(4*6.2,-1.2)--(23,-1.2);
    \node at (23.5,-1.7) {<s>};
    \draw [black,thick] (38.5,-2)--(4*9.3,-2)--(4*9.3,-1.2)--(38.5,-1.2);
    \node at (38,-1.7) {</s>};
    \node at (39.5,-1.5) {\small \bf re-SCRF};
    
    \foreach \sta/\sto/\letter in { 6.0/6.6/V/2, 6.6/7.2/E/4, 7.2/8.0/N/6, 8.0/8.5/I/8, 8.5/9.0/C/10, 9.0/9.5/E/12} {
    \draw [black,thick] ($(4*\sta,-3)$) rectangle ($(4*\sto,-2.2)$);
    \node at ($(4*\sta,-2.7)!.5!(4*\sto,-2.7)$) {\letter};
  }
    \draw [black,thick] (23,-3)--(4*6.0,-3)--(4*6.0,-2.2)--(23,-2.2);
    \node at (23.5,-2.7) {<s>};
    \draw [black,thick] (38.5,-3)--(4*9.5,-3)--(4*9.5,-2.2)--(38.5,-2.2);
    \node at (38.4,-2.7) {</s>};
    \node at (39.5,-2.5) {\small \bf 1p-SCRF};  

\end{tikzpicture}
\caption[Examples of hypothesis produced by models]{From top to bottom: frames from the word VENICE, with asterisks denoting the peak frame
  for each letter and ``$<$s$>$'' and ``$<$/s$>$'' denoting periods before the first letter
  and after the last letter; ground-truth segmentation based on peak
annotations (GT); segmentation produced by the tandem HMM (Tandem); segmentation produced by the rescoring SCRF (re-SCRF); and segmentation produced by the first-pass SCRF (1p-SCRF). From Signer 1.}
\label{fig:recog1}
\end{figure*}

\begin{figure*}[!th]
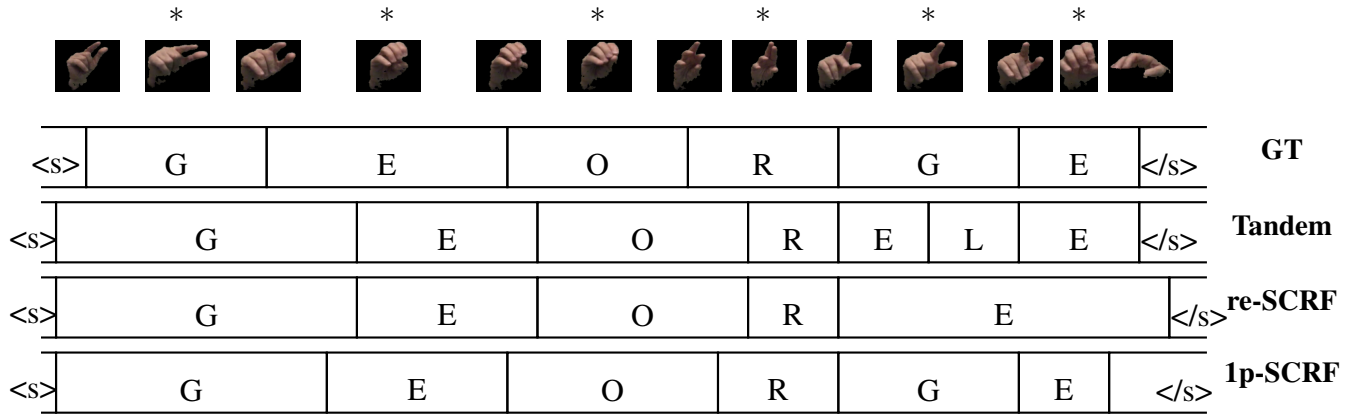

\begin{tikzpicture}
  \foreach \sta/\sto/\letter/\imnum in { 5.9/6.5/G/2, 6.5/7.3/E/4, 7.3/7.9/O/6, 7.9/8.4/R/8, 8.4/9.0/G/10, 9.0/9.4/E/12} {
    \draw [black,thick] ($(4*\sta,0)$) rectangle ($(4*\sto,.8)$);
    \node at ($(4*\sta,0.3)!.5!(4*\sto,.3)$) {\letter};
    \node at ($(4*\sta,1.6)!.5!(4*\sto,1.6)$)
    {\includegraphics[width=1.1cm]{ex/ex_drucie\imnum}};
    \node at ($(4*\sta,2.3)!.5!(4*\sto,2.3)$) {$\ast$};
  }
  \foreach \pos/\imnum in {5.9/1, 6.5/3, 7.3/5, 7.9/7, 8.4/9, 9.0/11, 9.4/13}
  \node at ($(4*\pos,1.6)$) {\includegraphics[width=1.1cm]{ex/ex_drucie\imnum}};
  \draw [black,thick] (23,0)--(4*5.9,0)--(4*5.9,.8)--(23,.8);
  \node at (23.2,.3) {<s>};
  \draw [black,thick] (38.5,0)--(9.4*4,0)--(9.4*4,.8)--(38.5,.8);
  \node at (38,.3) {</s>};
  \node at (39.5,.5) {\small \bf GT};
  
  \foreach \sta/\sto/\letter in { 5.8/6.8/G, 6.8/7.4/E, 7.4/8.1/O, 8.1/8.4/R, 8.4/8.7/E, 8.7/9.0/L, 9.0/9.4/E} {
    \draw [black,thick] ($(4*\sta,-1)$) rectangle ($(4*\sto,-.2)$);
    \node at ($(4*\sta,-.7)!.5!(4*\sto,-.7)$) {\letter};
  }
    \draw [black,thick] (23,-1)--(4*5.8,-1)--(4*5.8,-.2)--(23,-.2);
    \node at (22.9,-.7) {<s>};
    \draw [black,thick] (38.5,-1)--(4*9.4,-1)--(4*9.4,-.2)--(38.5,-.2);
    \node at (38,-.7) {</s>};
    \node at (39.5,-.5) {\small \bf Tandem};

    \foreach \sta/\sto/\letter in { 5.8/6.8/G, 6.8/7.4/E, 7.4/8.1/O, 8.1/8.4/R, 8.4/9.5/E} {
    \draw [black,thick] ($(4*\sta,-2)$) rectangle ($(4*\sto,-1.2)$);
    \node at ($(4*\sta,-1.7)!.5!(4*\sto,-1.7)$) {\letter};
  }
    \draw [black,thick] (23,-2)--(4*5.8,-2)--(4*5.8,-1.2)--(23,-1.2);
    \node at (22.9,-1.7) {<s>};
    \draw [black,thick] (38.5,-2)--(4*9.5,-2)--(4*9.5,-1.2)--(38.5,-1.2);
    \node at (38.4,-1.7) {</s>};
    \node at (39.5,-1.5) {\small \bf re-SCRF};
    
    \foreach \sta/\sto/\letter in { 5.8/6.7/G, 6.7/7.3/E, 7.3/8.0/O, 8.0/8.4/R, 8.4/9.0/G, 9.0/9.3/E} {
    \draw [black,thick] ($(4*\sta,-3)$) rectangle ($(4*\sto,-2.2)$);
    \node at ($(4*\sta,-2.7)!.5!(4*\sto,-2.7)$) {\letter};
  }
    \draw [black,thick] (23,-3)--(4*5.8,-3)--(4*5.8,-2.2)--(23,-2.2);
    \node at (22.9,-2.7) {<s>};
    \draw [black,thick] (38.5,-3)--(4*9.3,-3)--(4*9.3,-2.2)--(38.5,-2.2);
    \node at (38.2,-2.7) {</s>};
    \node at (39.5,-2.5) {\small \bf 1p-SCRF};  

\end{tikzpicture}
\caption{Similar to Figure~\ref{fig:recog1} for the word GEORGE, from Signer 2.}
\label{f:recog2}
\end{figure*}

\begin{figure*}[!th]
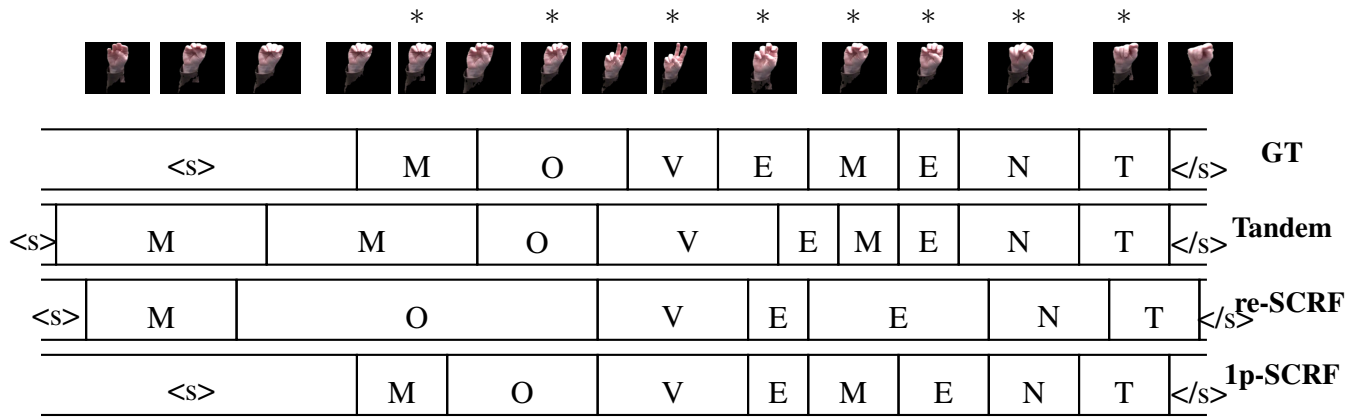

\begin{tikzpicture}
  \foreach \sta/\sto/\letter/\imnum in { 6.8/7.2/M/2, 7.2/7.7/O/4, 7.7/8.0/V/6, 8.0/8.3/E/8,8.3/8.6/M/10, 8.6/8.8/E/12, 8.8/9.2/N/14, 9.2/9.5/T/16} {
    \draw [black,thick] ($(4*\sta,0)$) rectangle ($(4*\sto,.8)$);
    \node at ($(4*\sta,0.3)!.5!(4*\sto,.3)$) {\letter};
    \node at ($(4*\sta,1.6)!.5!(4*\sto,1.6)$)
    {\includegraphics[width=1.1cm]{ex/ex_robin\imnum}};
    \node at ($(4*\sta,2.3)!.5!(4*\sto,2.3)$) {$\ast$};
  }
  \foreach \pos/\imnum in {6.8/1, 7.2/3, 7.65/5, 9.6/17} 
  \node at ($(4*\pos,1.6)$) {\includegraphics[width=1.1cm]{ex/ex_robin\imnum}};
  \draw [black,thick] (23,0)--(4*6.8,0)--(4*6.8,.8)--(23,.8);
  \node at (25,.3) {<s>};
  \draw [black,thick] (38.5,0)--(9.5*4,0)--(9.5*4,.8)--(38.5,.8);
  \node at (38.4,.3) {</s>};
  \node at (39.5,.5) {\small \bf GT};
  
    \foreach \pos/\imnum in {6.5/3, 6.25/2, 6.0/1} 
  \node at ($(4*\pos,1.6)$) {\includegraphics[width=1.1cm]{ex/ex_robin\imnum_pre}};
  
  \foreach \sta/\sto/\letter in { 5.8/6.5/M, 6.5/7.2/M, 7.2/7.6/O, 7.6/8.2/V/8,8.2/8.4/E, 8.4/8.6/M, 8.6/8.8/E, 8.8/9.2/N, 9.2/9.5/T} {
    \draw [black,thick] ($(4*\sta,-1)$) rectangle ($(4*\sto,-.2)$);
    \node at ($(4*\sta,-.7)!.5!(4*\sto,-.7)$) {\letter};
  }
    \draw [black,thick] (23,-1)--(4*5.8,-1)--(4*5.8,-.2)--(23,-.2);
    \node at (22.9,-.7) {<s>};
    \draw [black,thick] (38.5,-1)--(4*9.5,-1)--(4*9.5,-.2)--(38.5,-.2);
    \node at (38.4,-.7) {</s>};
    \node at (39.5,-.5) {\small \bf Tandem};

    \foreach \sta/\sto/\letter in { 5.9/6.4/M, 6.4/7.6/O, 7.6/8.1/V, 8.1/8.3/E,8.3/8.9/E, 8.9/9.3/N, 9.3/9.6/T} {
    \draw [black,thick] ($(4*\sta,-2)$) rectangle ($(4*\sto,-1.2)$);
    \node at ($(4*\sta,-1.7)!.5!(4*\sto,-1.7)$) {\letter};
  }
    \draw [black,thick] (23,-2)--(4*5.9,-2)--(4*5.9,-1.2)--(23,-1.2);
    \node at (23.2,-1.7) {<s>};
    \draw [black,thick] (38.5,-2)--(4*9.6,-2)--(4*9.6,-1.2)--(38.5,-1.2);
    \node at (38.75,-1.7) {</s>};
    \node at (39.6,-1.5) {\small \bf re-SCRF};
    
    \foreach \sta/\sto/\letter in { 6.8/7.1/M, 7.1/7.6/O, 7.6/8.1/V,8.1/8.3/E, 8.3/8.6/M, 8.6/8.9/E, 8.9/9.2/N, 9.2/9.5/T} {
    \draw [black,thick] ($(4*\sta,-3)$) rectangle ($(4*\sto,-2.2)$);
    \node at ($(4*\sta,-2.7)!.5!(4*\sto,-2.7)$) {\letter};
  }
    \draw [black,thick] (23,-3)--(4*6.8,-3)--(4*6.8,-2.2)--(23,-2.2);
    \node at (25,-2.7) {<s>};
    \draw [black,thick] (38.5,-3)--(4*9.5,-3)--(4*9.5,-2.2)--(38.5,-2.2);
    \node at (38.4,-2.7) {</s>};
    \node at (39.5,-2.5) {\small \bf 1p-SCRF};  

\end{tikzpicture}
\caption{Similar to Figure~\ref{fig:recog1} for the word MOVEMENT, from Signer 4.}
\label{f:recog4}
\end{figure*}

Figures~\ref{fig:recog1}, \ref{f:recog2}, \ref{f:recog3} and \ref{f:recog4} illustrate the recognition
task and hypothesis produced by models. In each of these figures we show the ground truth segments. The peak
frames are shown on top of each letter's segment; the hand region
segmentation masks were obtained automatically using the probabilistic
model described in Section~\ref{sec:data}. We also show intermediate
frames, obtained at midpoints between peaks, as well as frames before
the first peak and after the last peak. Below the ground truth
segmentations 
are the segmentations obtained with the tandem HMM, rescoring SCRF, and at
the bottom are segmentations obtained with the first-pass SCRF. 

\chapter{Example images of phonological feature values}
\label{sec:ex_ph_features}

\begin{table*}
\centering
\resizebox{\linewidth}{!}{
\begin{tabular}{|c||cccccc|}
\hline
{\bf Feature}        & \multicolumn{6}{c|}{{\bf Examle images of each value}} \\ \hline
&&&&&& \\
SF point of reference & \includegraphics[width=0.2\linewidth]{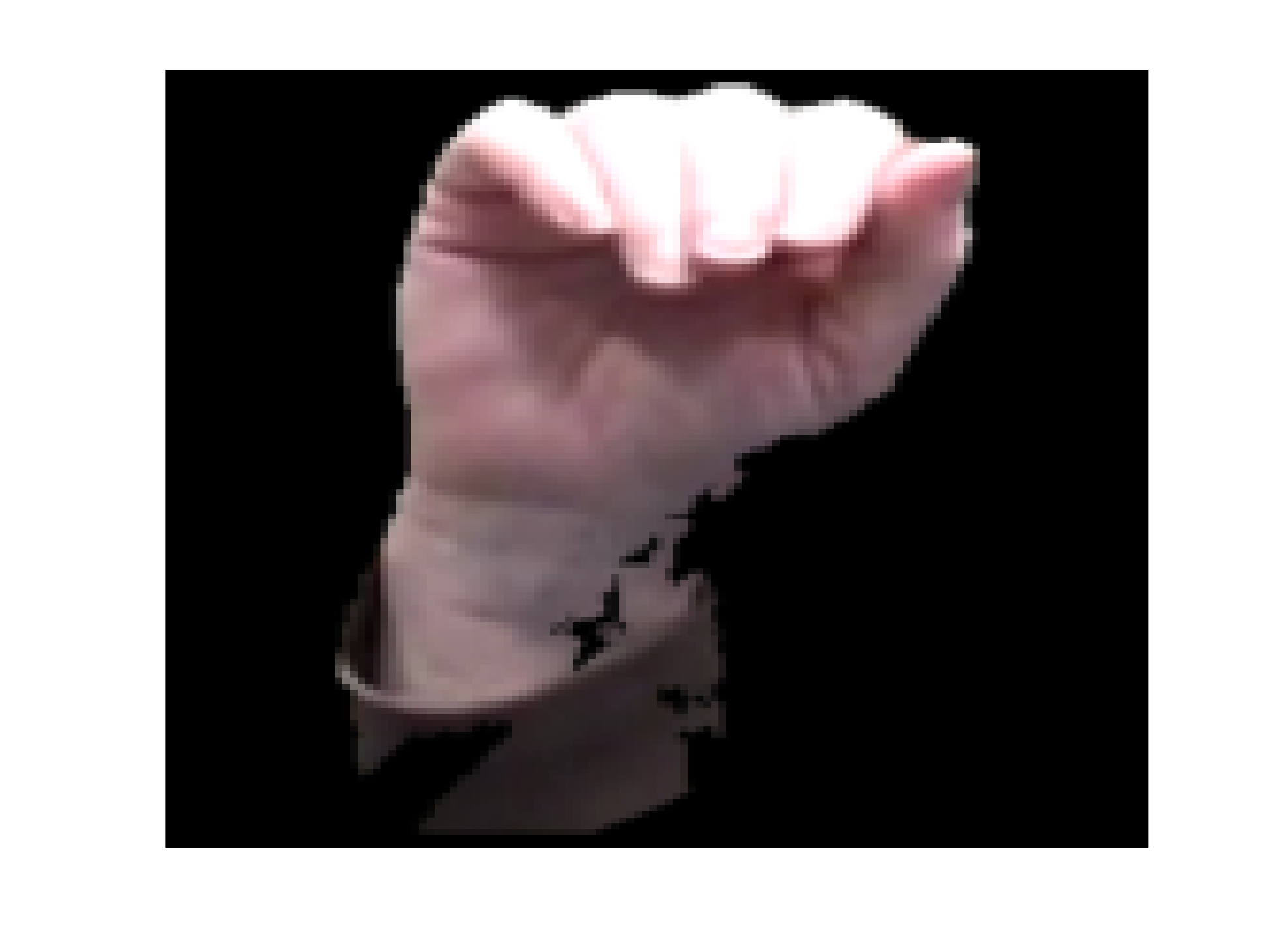} & \includegraphics[width=0.2\linewidth]{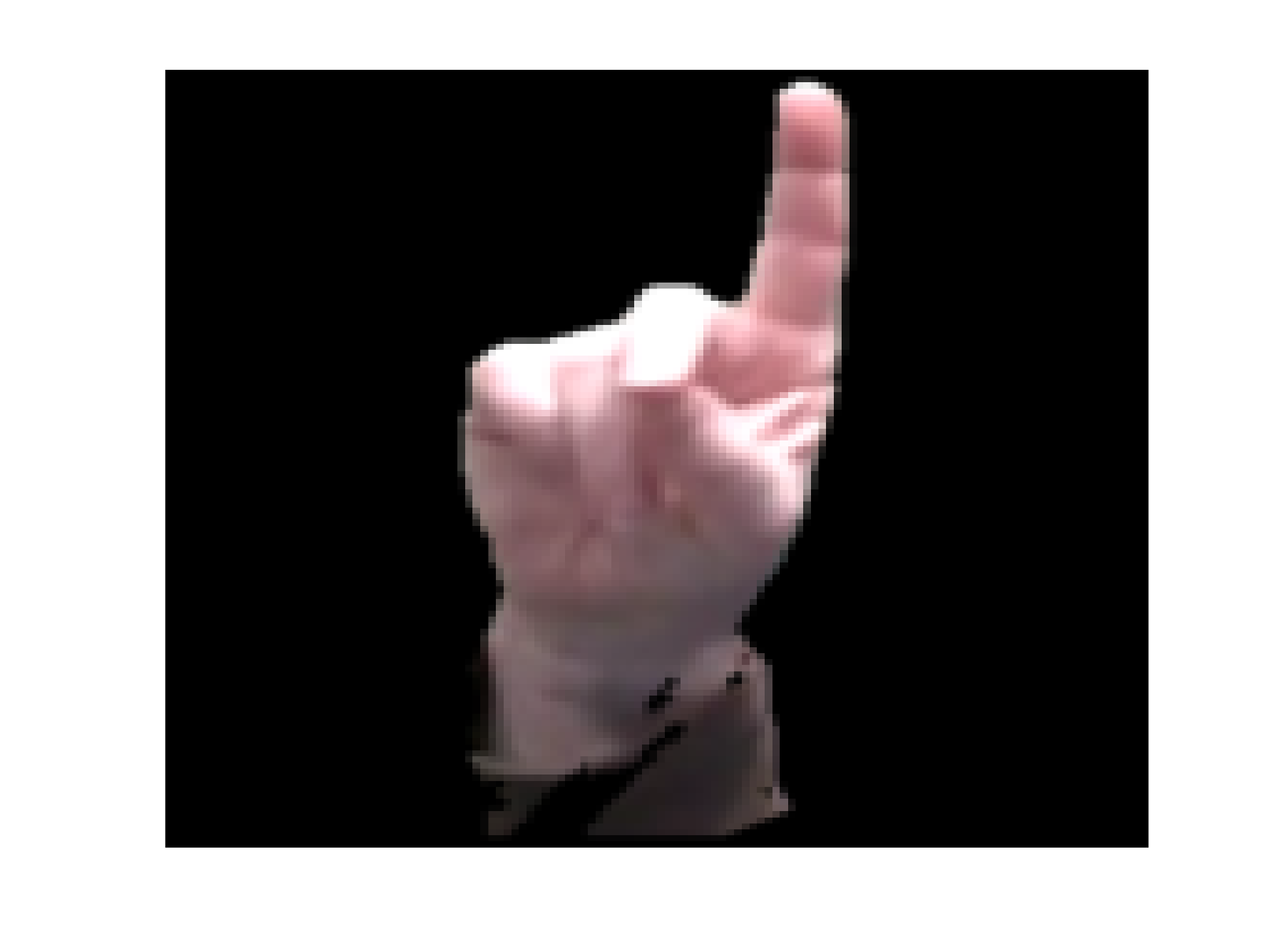} & \includegraphics[width=0.2\linewidth]{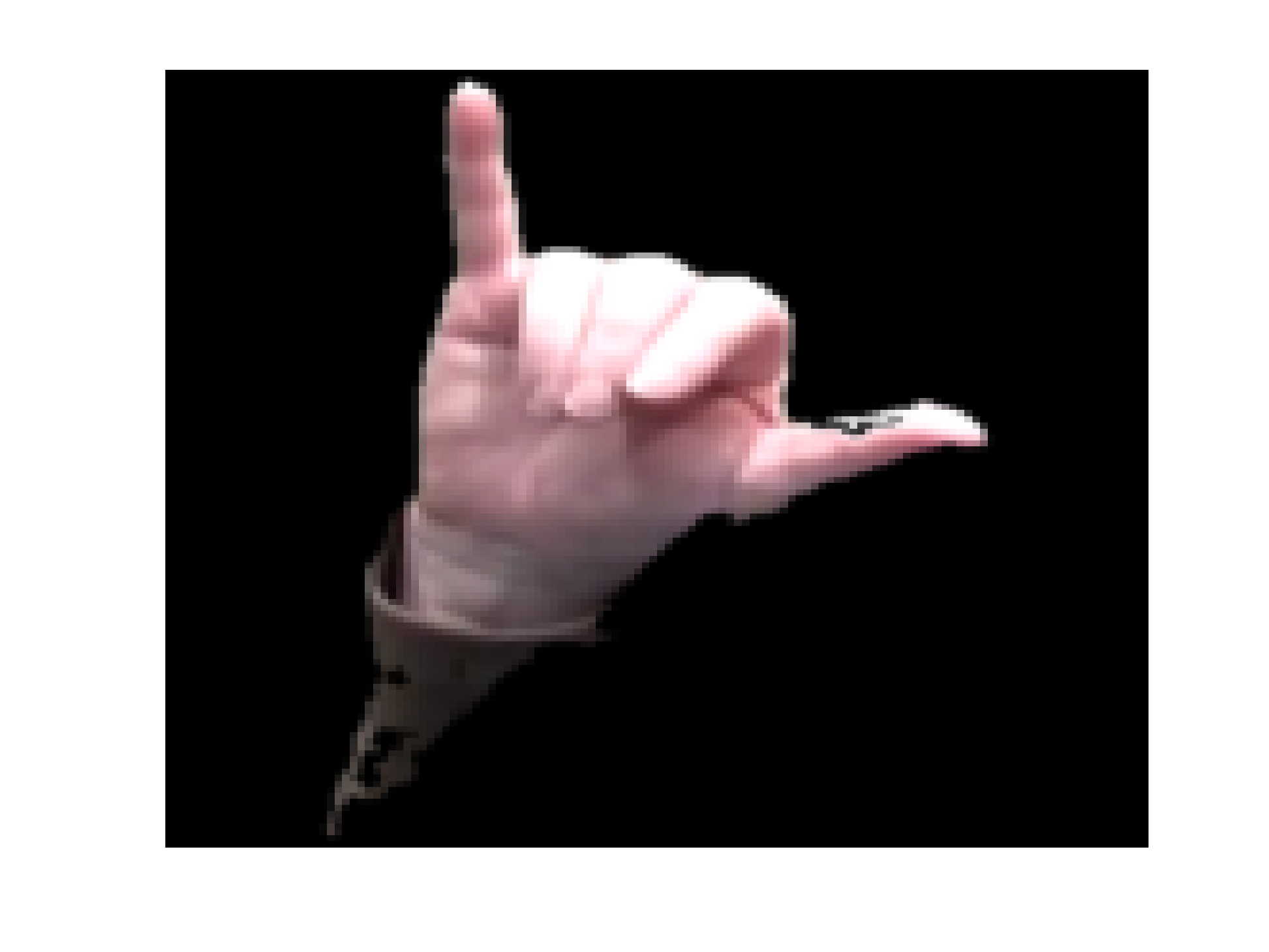} & & & \\  
& radial ~(A) & ulnar ~(D) & radial/ulnar ~(Y) & & & \\ \hline 
&&&&&& \\ 
SF joints & \includegraphics[width=0.2\linewidth]{ex/al_d.pdf} & \includegraphics[width=0.2\linewidth]{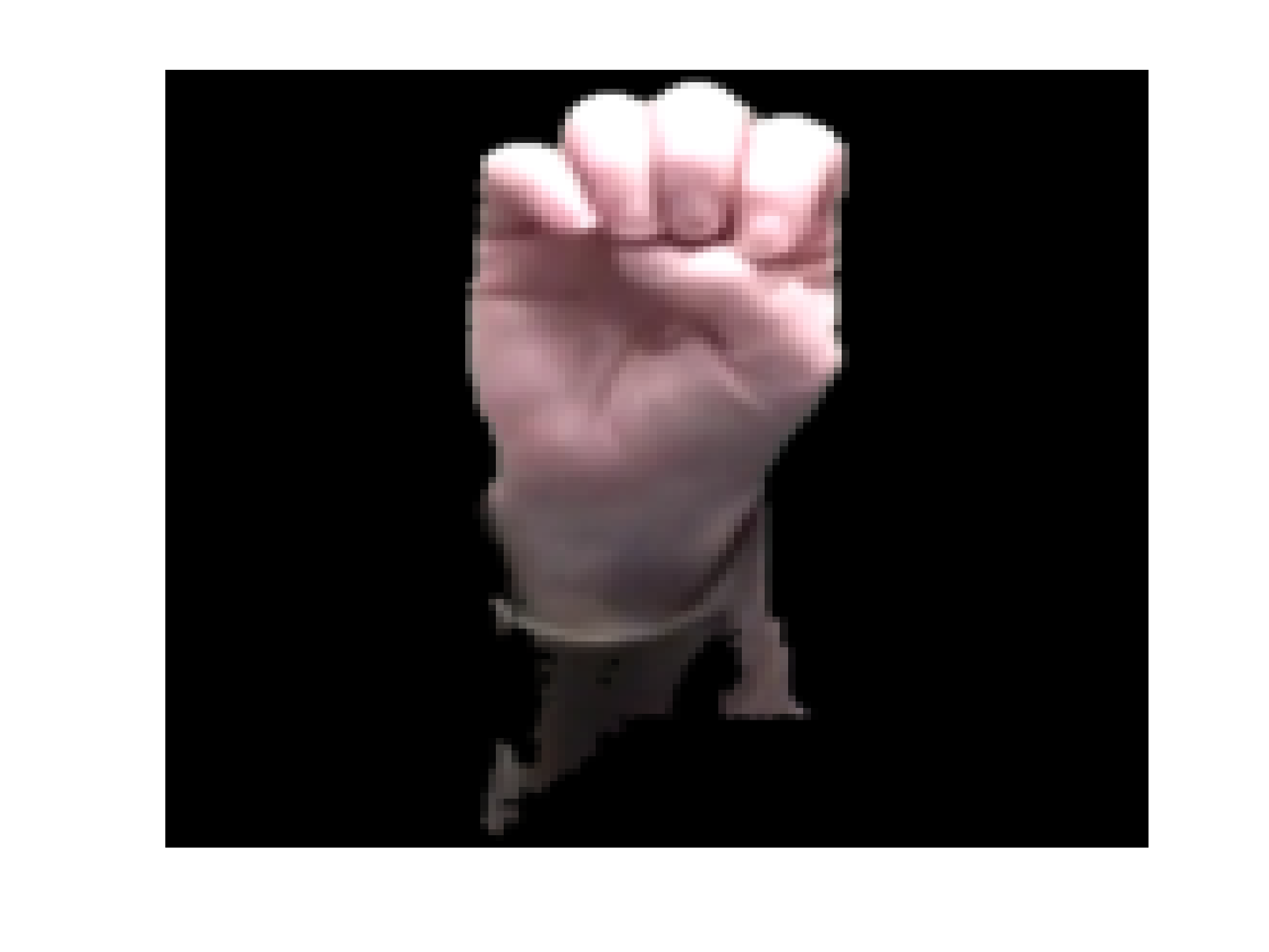} & \includegraphics[width=0.2\linewidth]{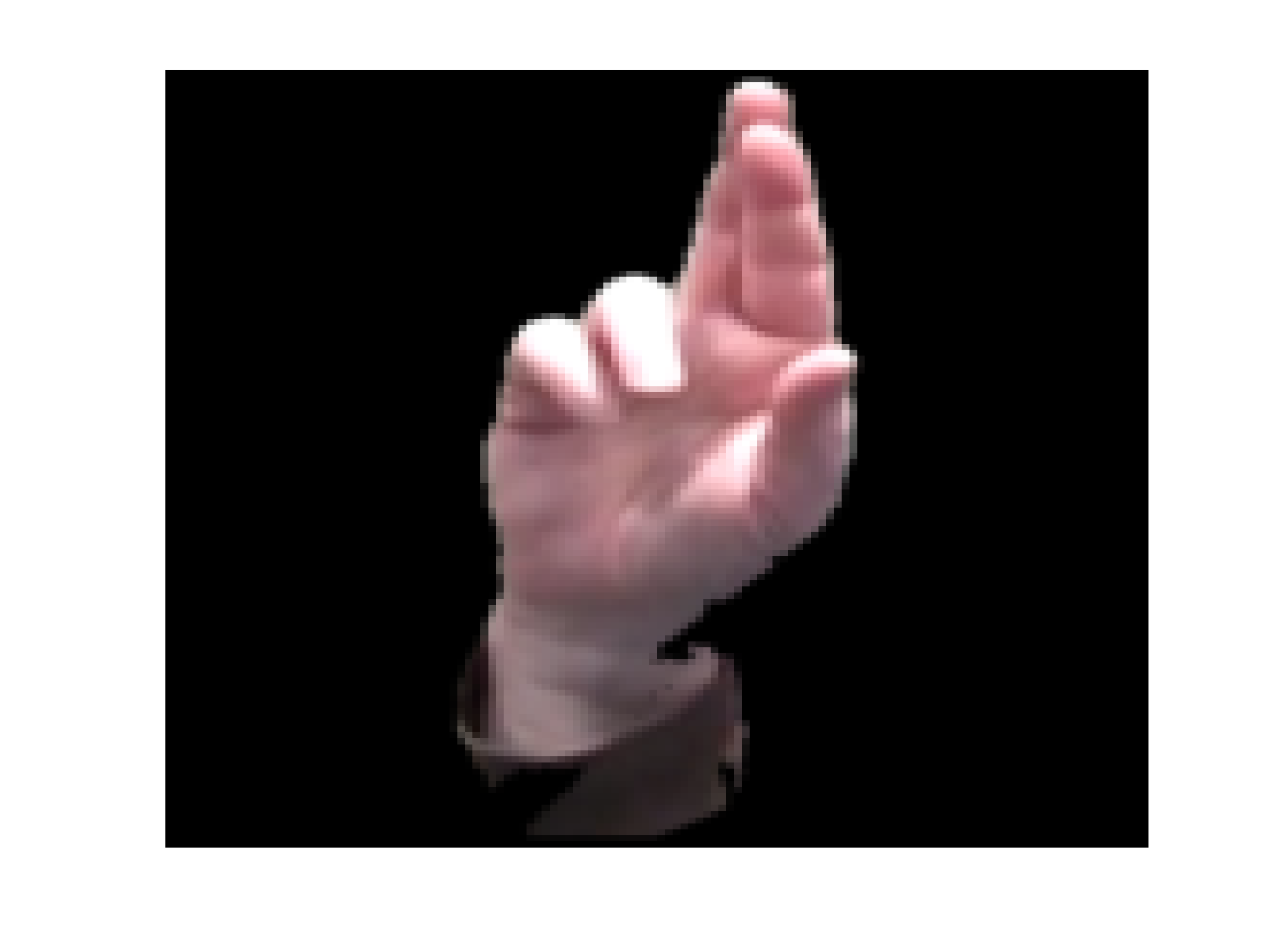} &\includegraphics[width=0.2\linewidth]{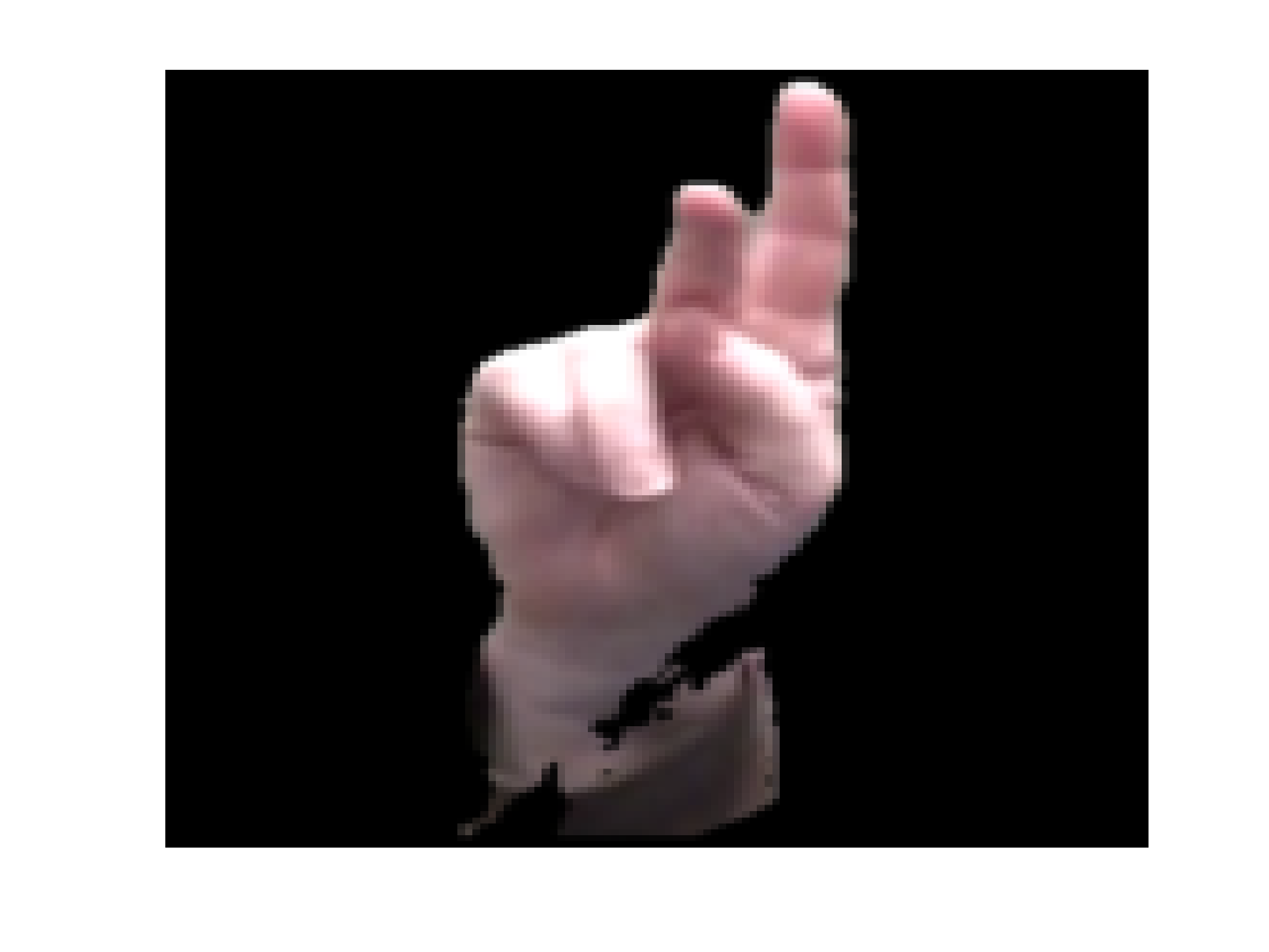} &\includegraphics[width=0.2\linewidth]{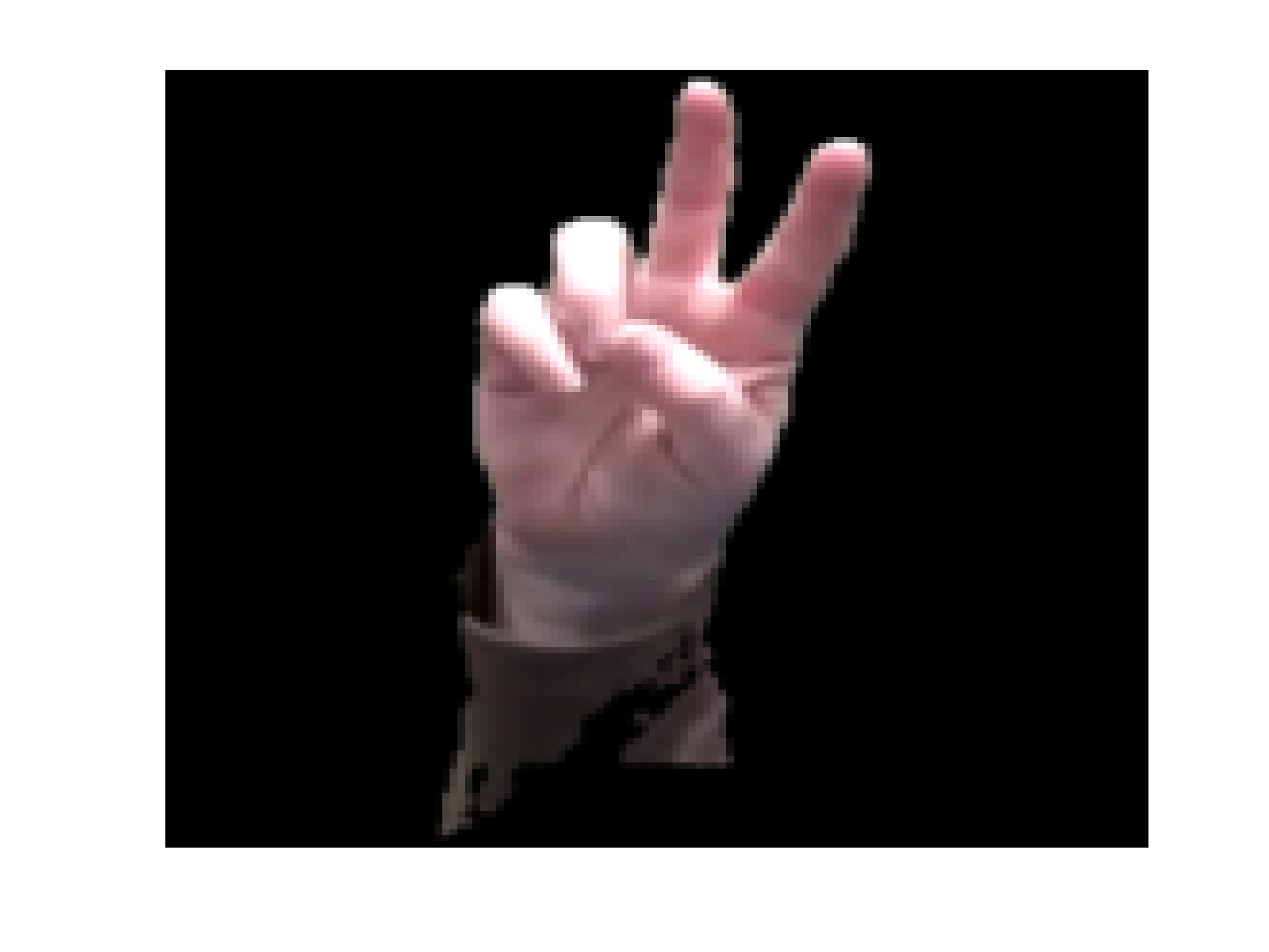} &\includegraphics[width=0.2\linewidth]{ex/al_a.pdf} \\ 
 & flexed:base ~(D) & flexed:nonbase ~(E) & flexed:base \& nonbase ~(R) & stacked ~(K)& crossed ~(V)& spread ~(A) \\ \hline
&&&&&& \\
SF quantity & \includegraphics[width=0.2\linewidth]{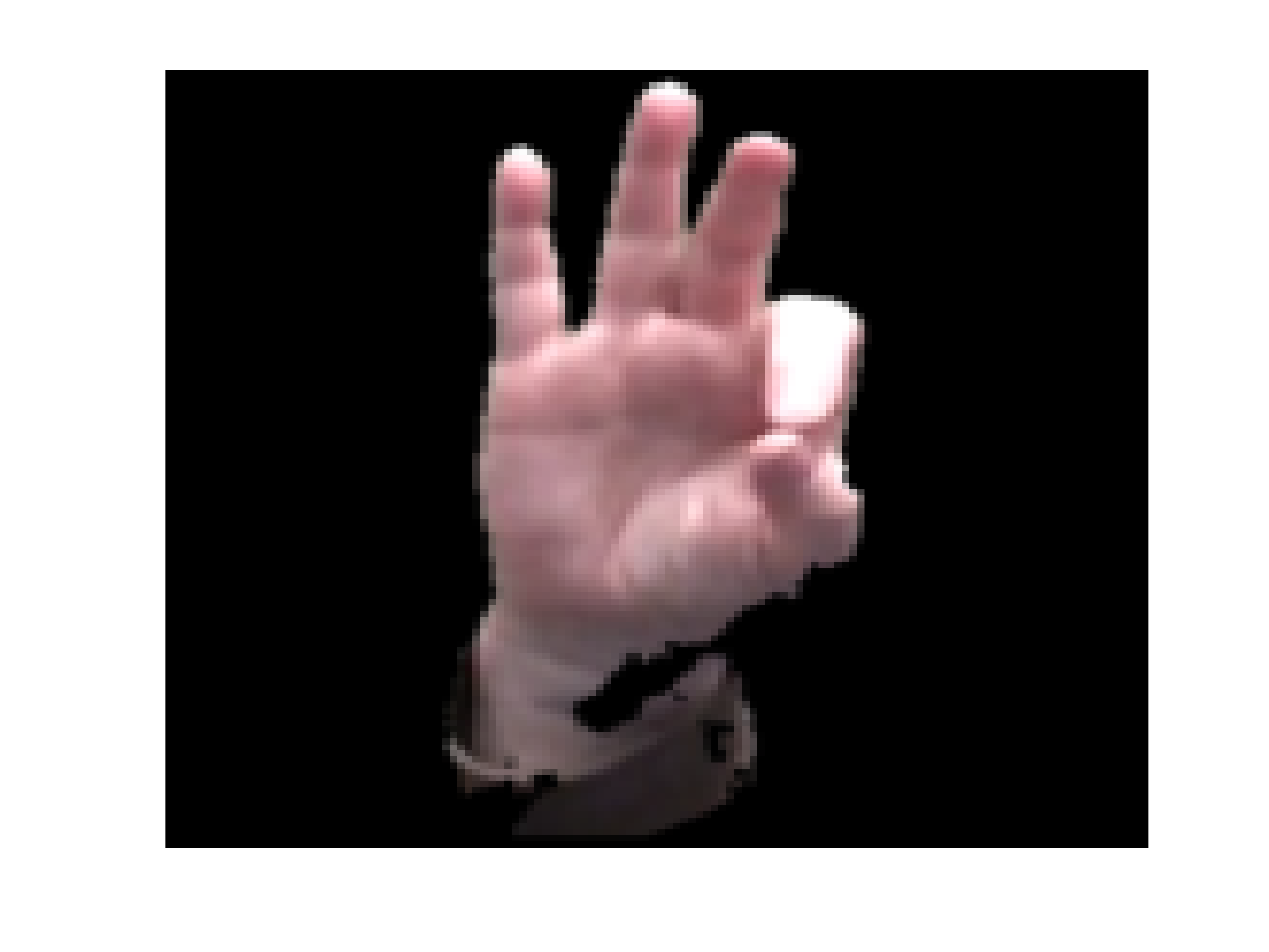} & \includegraphics[width=0.2\linewidth]{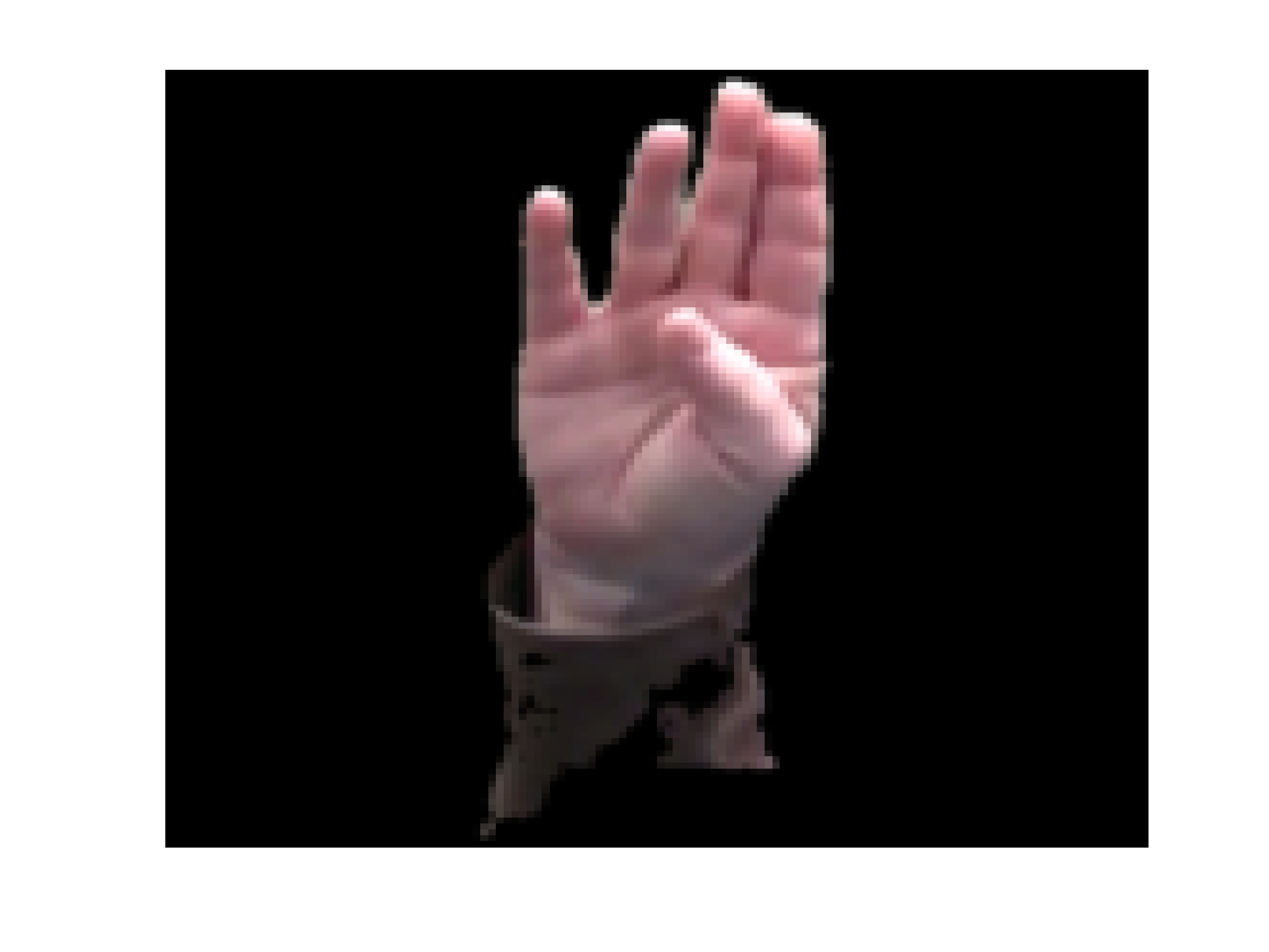} & \includegraphics[width=0.2\linewidth]{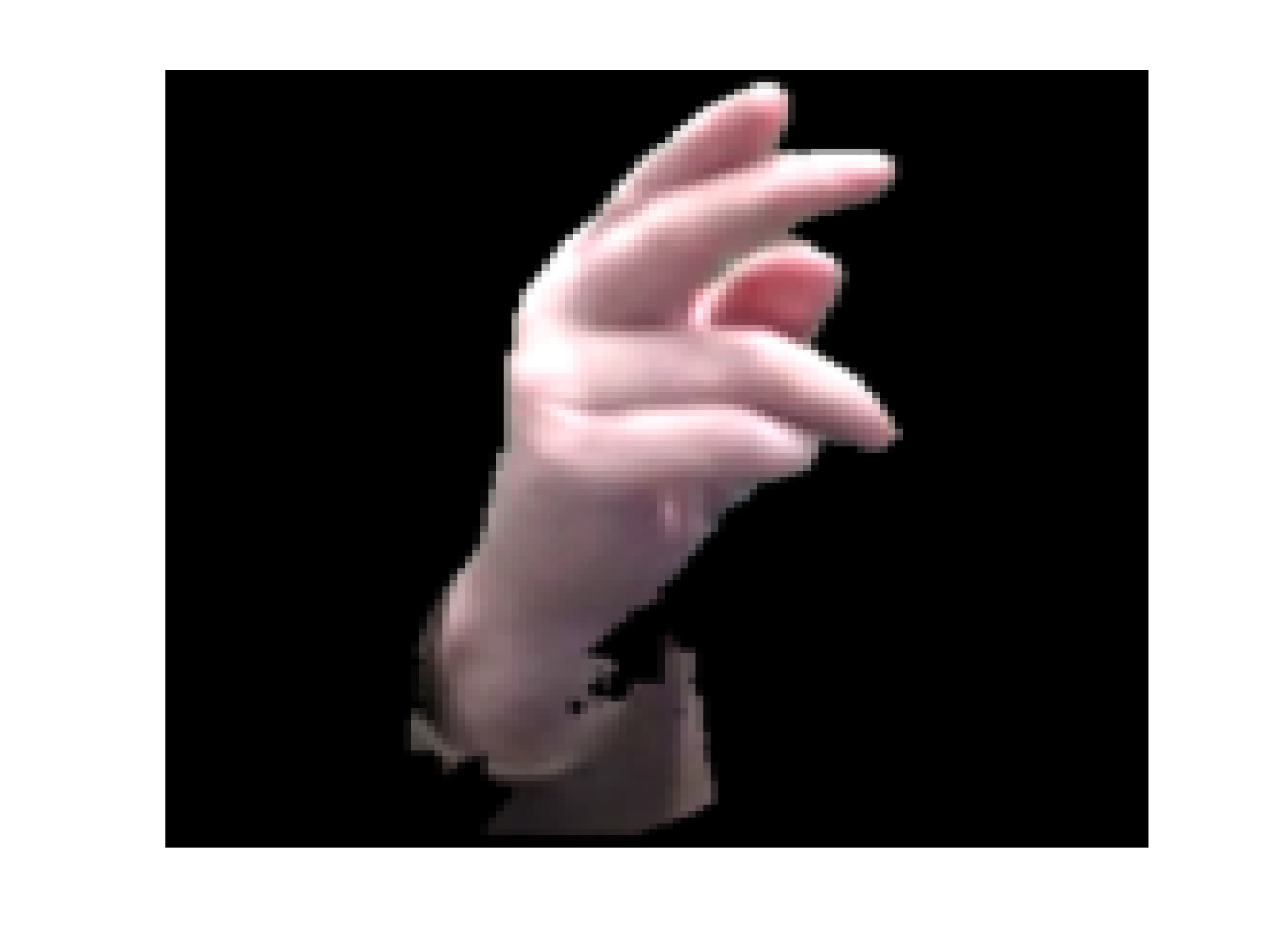} & \includegraphics[width=0.2\linewidth]{ex/al_d.pdf}& & \\
& all ~(F)& one ~(B) & one $>$ all ~(H) & all $>$ one ~(D) & & \\ \hline
&&&&&& \\
SF thumb & \includegraphics[width=0.2\linewidth]{ex/al_a.pdf} & \includegraphics[width=0.2\linewidth]{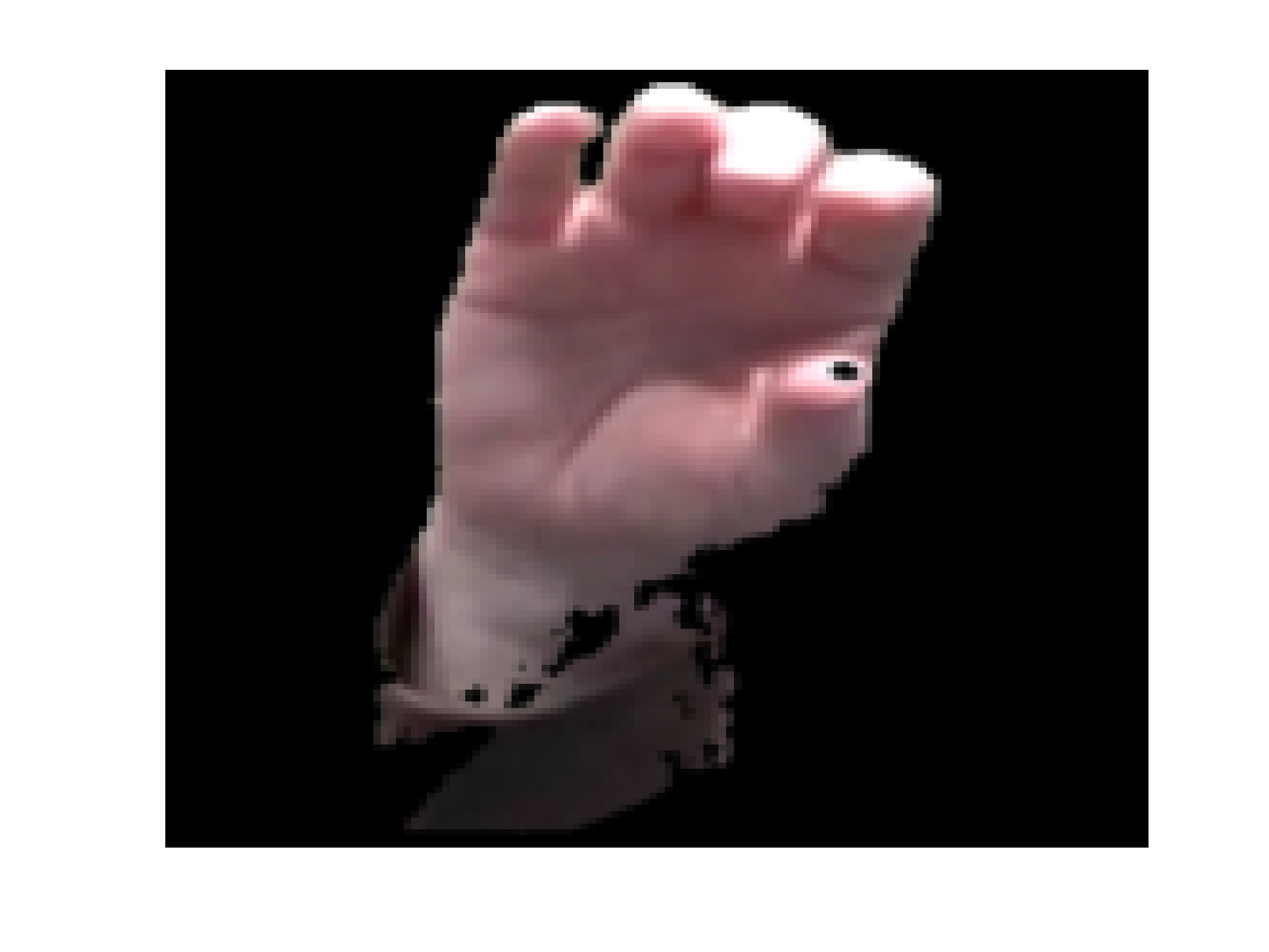} &&&& \\
& unopposed ~(A)& opposed ~(C)&&&& \\ \hline
&&&&&& \\
SF handpart & \includegraphics[width=0.2\linewidth]{ex/al_a.pdf} & \includegraphics[width=0.2\linewidth]{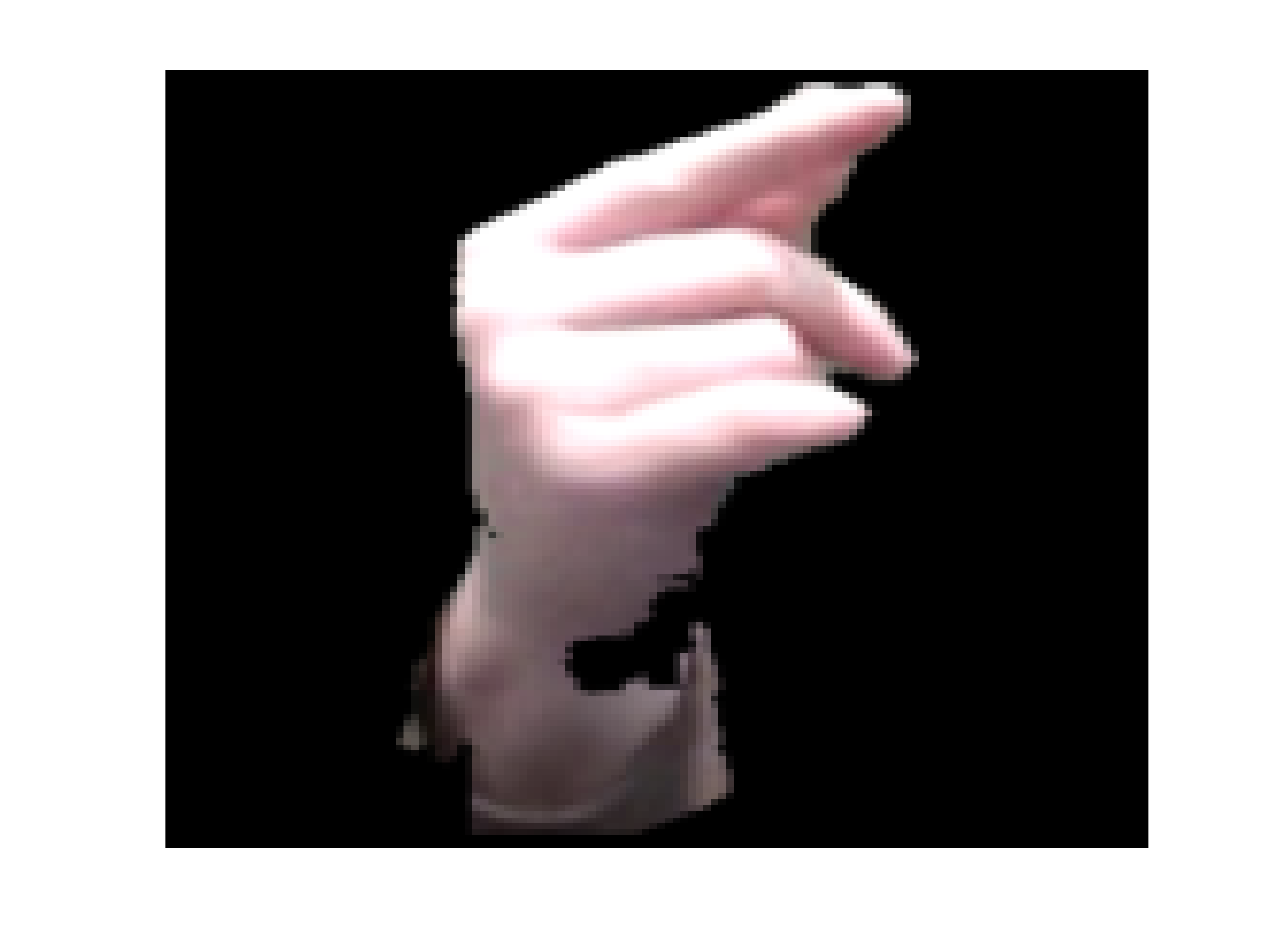} & \includegraphics[width=0.2\linewidth]{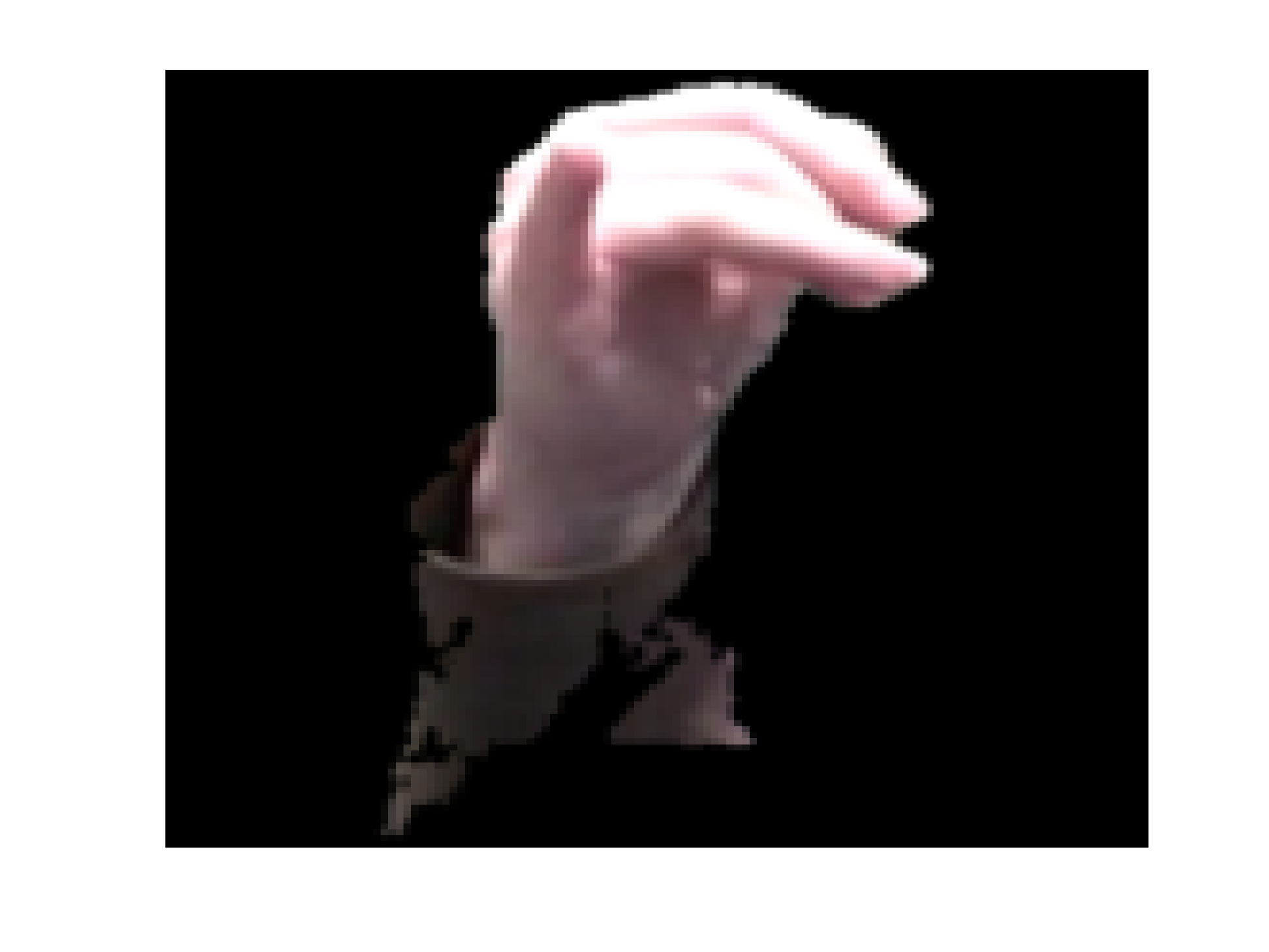} & & & \\
& base ~(A)& palm ~(G)& ulnar ~(J)&&& \\ \hline
&&&&&& \\
Unselected Fingers & \includegraphics[width=0.2\linewidth]{ex/al_d.pdf} & \includegraphics[width=0.2\linewidth]{ex/al_a.pdf} &&&& \\ 
&open ~(D)& closed ~(A)&&&& \\ \hline

\end{tabular}
}
\caption[Example images of each phonological feature values]{Example images of each phonological feature values with letter labels. For details, see Table \ref{t:features}.}
\label{f:ex_ph_features}
\end{table*}

\clearpage
\newpage

\chapter{Analysis: decomposition of errors}
\label{sec:decomp}


For further analysis of recognition approaches, we decompose the letter error rate. After having recognized hypothesis from models, optimal string match using dynamic programming match each of the recognized and reference label sequences. With the optimal alignment, the number of substitution errors (S), deletion errors (D) and insertion errors (I) to match from reference label sequence to recognized hypothesis can be computed. Please note that letter error rate is computed as:
\[
\text{Letter Error Rate} = \frac{D+S+I}{N} * 100\text{\%}
\]
and we use implementation of HTK \cite{htk} for this analysis. The results are reported in Table \ref{t:decomp_dep} with signer-dependent experiment, Table \ref{t:decomp_indep} with signer-independent, and Table \ref{t:decomp_adapt2} with adaptation. With signer-independent setting (Table \ref{t:decomp_indep}), 1st-pass model seems to produce more deletions, and less substitution and insertion. Note that 1st-pass model has worse letter error rates than other models (Table \ref{t:LER}). But, with adaptation and more accurate DNN frame classifiers (Table \ref{t:decomp_adapt2}) , first-pass SCRF model improves deletion and substitution errors over other models.   Also, for signer-dependent recognition, Table \ref{t:decomp_dep} shows that rescoring SCRF improves over tandem HMM, and first-pass SCRF improves over other models, mainly for deletion error.

\begin{table*}[ht!]
\centering
\resizebox{\linewidth}{!}{
\begin{tabular}{|l||c|c|c||c|c|c||c|c|c|}
\hline
        & \multicolumn{3}{c||}{Tandem HMM} & \multicolumn{3}{c||}{Rescoring SCRF} & \multicolumn{3}{c|}{1st-pass SCRF} \\ \hline
Signer  & D & S &  I & D & S &  I  & D & S &  I \\ \hline\hline
S1&26.4&	22.2&	5.5&	28.5&20.4&3.7&	39.5&13.8&1.7	\\ \hline
S2&26.4&18.9&9.4	&	22.9&	20.1&	8.2&24.6	&20.9&8.1		\\ \hline
S3&	37.1&20.4&5.1&39.4&18.0&3.8& 59.3  & 11.5  &  2.4\\ \hline
S4&	33.6&18.8&5.1&33.2&18.3&4.9& 38.9&   19.7&    4.1\\ \hline
{\bf Mean}&{\bf 30.9}&{\bf 20.1}&{\bf 6.3}&{\bf 31.0}&{\bf 19.2}&{\bf 5.2}&	{\bf 40.6 } & {\bf 16.5}&   {\bf 4.1}\\ \hline
\end{tabular}
}

\caption[Decomposition of letter error rate with signer-independent experiment]{Decomposition of letter error rate with signer-independent experiment (\%); relative numbers of D=Deletion,S=substitution,I=Insertion of each signer and mean of them, with respect to each recognizer.}
\label{t:decomp_indep}
\end{table*}

\begin{table*}[ht!]
\centering
\resizebox{\linewidth}{!}{
\begin{tabular}{|l||c|c|c||c|c|c||c|c|c|}
\hline
        & \multicolumn{3}{c||}{Tandem HMM} & \multicolumn{3}{c||}{Rescoring SCRF} & \multicolumn{3}{c|}{1st-pass SCRF} \\ \hline
Signer  & D & S &  I & D & S &  I  & D & S &  I \\ \hline\hline
S1&13.5&	5.7&	2.8&	13.5&	7.2&	1.8&	8.3&	3.5&	1.8	\\ \hline
S2&6.6&	2.8&	3.7&	6.2&	3.7&	3.5&	4.7&	2.5&	4.3	\\ \hline
S3&18&	11.3&	2.4&	19&	9.2&	1.4&	12.4&	7.7&	1.7	\\ \hline
S4&10.6&	7.3&	3.5&	9.6&	7.8&	3.9&	8.8&	7.1&	2.8	\\ \hline
{\bf Mean}&{\bf 12.2}&{\bf 	6.8}&{\bf 	3.1}&{\bf 	12.1}&{\bf 	7}&{\bf 	2.7}&{\bf 	8.6}&{\bf 	5.2}&{\bf 	2.6}	\\ \hline
\end{tabular}
}
\caption[Decomposition of letter error rate with adaptation]{Decomposition of letter error rate with adaptation (\%); relative numbers of D=Deletion,S=substitution,I=Insertion of each signer and mean of them, with respect to each recognizer.}
\label{t:decomp_adapt2}
\end{table*}

\begin{table*}[ht!]
\centering
\resizebox{\linewidth}{!}{
\begin{tabular}{|l||c|c|c||c|c|c||c|c|c|}
\hline
        & \multicolumn{3}{c||}{Tandem HMM} & \multicolumn{3}{c||}{Rescoring SCRF} & \multicolumn{3}{c|}{1st-pass SCRF} \\ \hline
Signer  & D & S &  I & D & S &  I  & D & S &  I \\ \hline\hline
S1&   10.0&    2.3  &  1.9 & 6.2&    2.9&    1.4&   4.6 & 1.2 & 1.3	\\ \hline
S2&        3.8&    1.7&    1.6& 3.8&    1.6&    1.6 & 4.4 & 0.7 & 1.6 \\ \hline
S3&14.2 &   6.6&    3.4&  7.9 &   5.1&    4.4	& 3.7&3.1&1.0 \\ \hline
S4& 5.4&    3.7 &   2.3 &  4.9&    2.8 &   2.1&	4.5&2.3&1.8 \\ \hline
{\bf Mean}&{\bf 8.4}&{\bf 3.6}&{\bf 2.3}&{\bf 5.7}&{\bf 3.1}&{\bf 2.4}&{\bf 	4.3}  &{\bf  1.8}&{\bf  1.4}\\ \hline
\end{tabular}
}
        
        

\caption[Decomposition of letter error rate with signer-dependent experiment]{Decomposition of letter error rate with signer-dependent experiment (\%); relative numbers of D=Deletion,S=substitution,I=Insertion of each signer and mean of them, with respect to each recognizer.}
\label{t:decomp_dep}
\end{table*}

\clearpage
\newpage

\chapter{Word lists recorded and used in experiments}
\label{sec:app_wordlist}

\begin{table*}
\centering
\resizebox{\linewidth}{!}{
\begin{tabular}{|llllll|}
\hline 
ABERDEEN & DANY & HEADLIGHT & MAANANTAI & POWAZNIE & TANZANIA \\ 
AFGHANISTAN & DEBBIE & HEI & MARY & PREKLADATEL & TAXI \\ 
AFRICA & DECK & HERB & MATERIAL & PROCVICOVAT & TIFFANY \\ 
AHOJ & DEKUJI & HIMALAYA & MATT & PRZEPRASZAM & TOBIAS \\ 
ALAN & DINOSAUR & HLAD & MAURITANIA & PUHU & TOBY \\ 
ALCAPULCO & DLACZEGO & HOGYAN & MEDITERRANEAN & QUANTITY & TOISTEKAN \\ 
ALEXANDER & DNIA & HOL & MENNYIBE & QUARRY & TOKYO \\ 
AMY & DOGFIGHT & HUOMENTA & MEXICO & QUARTER & TOM \\ 
ANGELICA & DON & HUONE & MIA & QUEEN & TUHAT \\ 
ANN & EARTHQUAKE & HYVAA & MILUJI & QUENTIN & TULIP \\ 
ANTEEKSI & EGYENESEN & IGEK & MINA & QUESTION & TURQUOISE \\ 
APPETIZERS & ELNEZEST & IGEN & MISSA & QUICKSAND & TWIZZLERS \\ 
APRAXIA & ELSALVADOR & INFORMACJA & MISSISSIPPI & QUILT & USTA \\ 
AQUARIUM & ENRIQUE & INGLEWOOD & MITTEN & QUINCY & UTCA \\ 
ASPHYXIATION & EQUAL & INK & MOC & QUIZ & VACUM \\ 
ATAXIA & EVERGLADES & INSTRUMENT & MONGOLIA & QUOTATION & VAN \\ 
ATLANTIC & EXCEL & ITT & MOSCOW & RADO & VCERA \\ 
AXEL & EXECUTIVE & IZZY & MUSTANG & RAKASTAN & VENEZUELA \\ 
AXIS & EXPECTATION & JACQUELINE & NAOMI & RANGEROVER & VENICE \\ 
AXON & EXPERT & JADE & NAPERVILE & RANO & VIISI \\ 
BASIL & EXPO & JASON & NAVSTIVIL & RENDORSEG & VITEJ \\ 
BASS & EXXON & JAWBREAKER & NEIGHBORHOOD & REST & VIV \\ 
BEA & FAMILY & JEGY & NELJA & RIDDLE & VOITTE \\ 
BEEF & FANBELT & JEWELRY & NENI & RITA & WAFFLE \\ 
BEIJING & FANNY & JIMMY & NEROZUMIM & RUSS & WEED \\ 
BELYEG & FELESEG & JOE & NIC & SAM & WIL \\ 
BIL & FELIX & JOHN & NIGDY & SANFRANCISCO & WILIAM \\ 
BLAHOPREJI & FELKELNI & JOSH & NOGI & SARA & WINDSHIELD \\ 
BOO & FERFI & JSOU & NOTEBOOK & SAUCE & WING \\ 
BOTSWANA & FERN & JUICE & NOWYCH & SCOTLAND & WLOSY \\ 
BOX & FINDINGS & JUNA & OAKPARK & SEED & XAVIER \\ 
CABIN & FINN & KAHDEKSAN & OLE & SEQUEL & XENON \\ 
CADILLAC & FIR & KAKSI & ONKO & SILK & XENOPHOBIA \\ 
CAMEROON & FIREWIRE & KATE & OPRAVDU & SINA & XEROX \\ 
CAMILA & FIU & KDE & OVAL & SIUSIU & XMEN \\ 
CAMPFIRE & FLEA & KELY & OWEN & SKOKIE & XYLOPHONE \\ 
CARIBBEAN & FLOSSMOOR & KERUL & OXEN & SOFTSERVE & YARD \\ 
CARL & FLOUR & KOLIK & OXYGEN & SPICE & YELOWSTONE \\ 
CARP & FRANCESCA & KORHAZ & PALJONKO & SPOTYKAC & YKSI \\ 
CHRIS & FRANKLIN & KOSZI & PALYADVAR & SPRUCE & YOSEMITE \\ 
CHWILKE & FRED & KTO & PAM & SQUARE & ZACK \\ 
CIE & FURNITURE & KUUSI & PARAGUAY & SQUIREL & ZDROWIE \\ 
CLAW & GARY & LAMB & PENZVALTAS & STAFF & ZEBRA \\ 
CLEVELAND & GAYLE & LENTOKENTTA & PIEC & STOOL & ZGODA \\ 
CLIFF & GEORGE & LEO & POCALUJMY & STRAWBERY & ZGUBILAM \\ 
CLIFFHANGER & GIORDANO & LEXUS & POJD & SUN & ZIZEN \\ 
COLUMBUS & GLUE & LIBYA & PONY & SURGOS & ZOBACZENIA \\ 
CSOKIFAGYIT & GRAPE & LIFE & POSPESTE & SZIA & ZOE \\ 
CZESC & GRAVITY & LIQUID & POTREBUJI & TALLAHASSEE & ZOPAKOVAT \\ 
DAJ & GREG & LUGGAGE & POWAZANIEM & TANCOLNI & ZYC \\  \hline

\end{tabular}
}

\caption[Word list 1]{The first word list we use.}
\label{t:wordlist1}
\end{table*}

\begin{table*}
\centering
\resizebox{\linewidth}{!}{
\begin{tabular}{|llllll|}
\hline 
ACCOUNT & COUNTRY & GAME & MACHINE & PLAY & STATE \\ 
ACT & COUPLE & GARDEN & MAN & POINT & STATES \\ 
ACTION & COURSE & GIRL & MARKET & POLICE & STORY \\ 
ACTIVITY & COURT & GLASS & MATER & POLICY & STREET \\ 
AGE & DAUGHTER & GOD & MATERIAL & POSITION & STUDENT \\ 
AIR & DAY & GOVERNMENT & MATTER & POUND & STUDY \\ 
AMOUNT & DAYS & GROUND & MEMBER & POWER & SUBJECT \\ 
ANIMAL & DEAL & GROUP & METHOD & PRESIDENT & SUMMER \\ 
ANSWER & DEATH & HAIR & METING & PRESSURE & SUN \\ 
AREA & DECISION & HALL & MILE & PRICE & SUPPORT \\ 
ARGUMENT & DEGREE & HAND & MIND & PROBLEM & SYSTEM \\ 
ARM & DEPARTMENT & HEAD & MINISTER & PROCESS & TABLE \\ 
ART & DEVELOPMENT & HEALTH & MINUTE & PRODUCTION & TAX \\ 
ATTENTION & DIFFERENCE & HEART & MISS & PROGRAMME & TEACHER \\ 
ATTITUDE & DOCTOR & HELP & MOMENT & PURPOSE & TERMS \\ 
AUTHORITY & DOOR & HISTORY & MONEY & QUALITY & THEORY \\ 
BABY & DOUBT & HOME & MONTH & QUESTION & THING \\ 
BACK & EARTH & HORSE & MORNING & RATE & THINGS \\ 
BANK & EDUCATION & HOSPITAL & MOTHER & REASON & THOUGHT \\ 
BED & EFECT & HOTEL & MOUTH & RELATIONSHIP & TIME \\ 
BENEFIT & EFORT & HOUR & MOVEMENT & REPORT & TOP \\ 
BIT & END & HOUSE & MRS & RESEARCH & TOWN \\ 
BLOOD & ENERGY & HUSBAND & MUSIC & REST & TRADE \\ 
BODY & EUROPE & IDEA & NAME & RESULT & TREE \\ 
BOOK & EVENING & INCOME & NATION & RIVER & TROUBLE \\ 
BOY & EVENT & INDUSTRY & NATURE & ROAD & TRUTH \\ 
BRITAIN & EVIDENCE & INFORMATION & NEED & ROLE & TYPE \\ 
BROTHER & EXAMPLE & INTEREST & NEWSPAPER & ROM & UNION \\ 
BUILDING & EXPERIENCE & ISSUE & NIGHT & RULE & UNIVERSITY \\ 
BUSINESS & EYE & JOB & NUMBER & SCHOOL & USE \\ 
CAR & FACE & KIND & OFICE & SEA & VALUE \\ 
CASE & FACT & KNOWLEDGE & OFICER & SECURITY & VIEW \\ 
CENT & FAMILY & LABOUR & OIL & SENSE & VILAGE \\ 
CENTRE & FATHER & LAND & ONCE & SERVICE & VOICE \\ 
CENTURY & FEAR & LANGUAGE & ONE & SEX & WALL \\ 
CHAIR & FEELING & LAW & ORDER & SHOP & WAR \\ 
CHANCE & FEW & LEADER & ORGANIZATION & SHOULDER & WATER \\ 
CHANGE & FIELD & LEAST & OTHERS & SIDE & WAY \\ 
CHAPTER & FIGURE & LEG & PAPER & SIGN & WEEK \\ 
CHILD & FINGER & LETTER & PARENT & SITUATION & WEST \\ 
CHURCH & FIRE & LEVEL & PART & SIZE & WHILE \\ 
CITY & FISH & LIFE & PARTY & SOCIETY & WIFE \\ 
CLASS & FLOOR & LIGHT & PEOPLE & SON & WILL \\ 
CLOTHES & FOD & LINE & PERIOD & SORT & WINDOW \\ 
CLUB & FOOT & LITTLE & PERSON & SOUND & WOMAN \\ 
COMMITTEE & FORCE & LONDON & PICTURE & SOURCE & WORD \\ 
COMMUNITY & FORM & LOOK & PIECE & SOUTH & WORK \\ 
COMPANY & FRIEND & LORD & PLACE & SPACE & WORKER \\ 
CONTROL & FRONT & LOT & PLAN & STAFF & WORLD \\ 
COST & FUTURE & LOVE & PLANT & STAGE & YEAR \\  \hline

\end{tabular}
}

\caption[Word list 2]{The second word list we use.}
\label{t:wordlist2}
\end{table*}

\clearpage
\newpage


\end{document}